\pdfoutput=1
\documentclass[10pt,twocolumn,letterpaper]{article}
\usepackage{cvpr}      
\usepackage{graphicx}
\usepackage{amsmath}
\usepackage{amssymb}
\usepackage{booktabs}
\usepackage[pagebackref,breaklinks,colorlinks]{hyperref}
\usepackage[capitalize]{cleveref}
\crefname{section}{Sec.}{Secs.}
\Crefname{section}{Section}{Sections}
\Crefname{table}{Table}{Tables}
\crefname{table}{Tab.}{Tabs.}

\usepackage[utf8]{inputenc} 
\usepackage[T1]{fontenc}    
\usepackage{hyperref}       
\usepackage{url}            
\usepackage{booktabs}       
\usepackage{amsfonts}       
\usepackage{nicefrac}       
\usepackage{microtype}      
\usepackage{xcolor}         
\usepackage{array}
\usepackage{verbatim}
\usepackage{wrapfig}
\usepackage{booktabs}
\usepackage{units}
\usepackage{mathtools}
\usepackage{multirow}
\usepackage{amsthm}
\usepackage{amssymb}
\usepackage[noend,ruled,vlined]{algorithm2e}
\usepackage{etoolbox}
\usepackage{enumitem}
\usepackage{caption}
\usepackage{subcaption}
\usepackage[utf8]{inputenc} 
\usepackage{mathtools}
\usepackage{bbm}

\usepackage{multicol}

\usepackage{pifont}

\DeclareMathOperator*{\argmax}{arg\,max}

\newcommand{\methodname}{ALFA-Mix}

\makeatletter
\patchcmd{\@algocf@start}
{-1.5em}
{0pt}
{}{}
\makeatother

\newcommand{\MYhref}[3][black]{\href{#2}{\color{#1}{#3}}}

\newcommand{\Expec}{\mathop{\mathbb{E}}}
\newcommand{\Data}{\mathcal{D}}
\newcommand{\lblData}{{\Data}^{l}}
\newcommand{\ulblData}{{\Data}^{u}}
\newcommand{\Bgt}{B}
\newcommand{\params}{\boldsymbol\theta}
\newcommand{\encparams}{\params_e} 
\newcommand{\clsparams}{\params_c} 

\newcommand{\inp}{\boldsymbol{x}}
\newcommand{\inpu}{\boldsymbol{x}^{u}}

\newcommand{\Lbl}{Y}
\newcommand{\lbl}{y}
\newcommand{\lblmaxarg}[1]{{y^*_{#1}}}
\newcommand{\lblmax}{{y^*}}
\newcommand{\softmax}{\text{softmax}}

\newcommand{\lblsize}{K}
\newcommand{\ltn}{\boldsymbol{z}}
\newcommand{\ltnu}{\boldsymbol{z}^{u}} 
\newcommand{\ltnl}{\boldsymbol{z}^{l}} 
\newcommand{\Ltn}{\boldsymbol{Z}}
\newcommand{\Ltnu}{\boldsymbol{Z}^u}
\newcommand{\Ltnl}{\boldsymbol{Z}^l}
\newcommand{\ltnC}{\tilde{\ltn}}

\newcommand{\intp}{\tilde{\ltn}_{\bnoise}}
\newcommand{\R}{\mathbb{R}}
\newcommand{\prt}{{\ltn^{\star}}} 
\newcommand{\Prt}{{\Ltn^{\star}}} 

\newcommand{\noise}{\alpha}
\newcommand{\bnoise}{\boldsymbol\noise}

\newcommand{\noiseMax}{\noise_{\text{max}}} 
\newcommand{\bnoiseOpt}{\bnoise^*} 
\newcommand{\noiseOptMax}{\epsilon} 
 
\newcommand{\Centroid}{\mathcal{C}}
\newcommand{\Uncertain}{\mathcal{I}}

\newcommand{\loss}{\ell}

\begin{document}

\title{Active Learning by Feature Mixing}

\author{Amin Parvaneh$^{1}$\qquad Ehsan Abbasnejad$^{1}$\qquad Damien Teney$^{1,2}$\qquad Reza Haffari$^{3}$,\\ 
\vspace{4mm}
Anton van den Hengel$^{1,4}$\qquad Javen Qinfeng Shi$^{1}$\\
$^{1}$Australian Institute for Machine Learning, University of Adelaide\\
$^{2}$Idiap Research Institute \qquad
$^{3}$Monash University \qquad
$^{4}$Amazon\\
{\tt\small {\{\MYhref{mailto:amin.parvaneh@adelaide.edu.au}{amin.parvaneh}, \MYhref{mailto:ehsan.abbasnejad@adelaide.edu.au}{ehsan.abbasnejad},\MYhref{mailto:javen.shi@adelaide.edu.au}{javen.shi}, \MYhref{mailto:anton.vandenhengel@adelaide.edu.au}{anton.vandenhengel}\}@adelaide.edu.au}}\\
{\tt\small {\MYhref{mailto:damien.teney@idiap.ch}{damien.teney@idiap.ch}}} \qquad
{\tt\small {\MYhref{mailto:gholamreza.haffari@monash.edu}{gholamreza.haffari@monash.edu}}}
}

\maketitle

\begin{abstract}
    The promise of active learning (AL) is to reduce labelling costs by selecting the most valuable examples to annotate from a pool of unlabelled data. Identifying these examples is especially challenging with high-dimensional data (\eg images, videos) and in low-data regimes. In this paper, we propose a novel method for batch AL called {\methodname}. We identify unlabelled instances with sufficiently-distinct features by seeking inconsistencies in predictions resulting from interventions on their representations. We construct interpolations between representations of labelled and unlabelled instances then examine the predicted labels. We show that inconsistencies in these predictions help discovering features that the model is unable to recognise in the unlabelled instances. We derive an efficient implementation based on a closed-form solution to the optimal interpolation causing changes in predictions. Our method outperforms all recent AL approaches in 30 different settings on 12 benchmarks of images, videos, and non-visual data. The improvements are especially significant in low-data regimes and on self-trained vision transformers, where {\methodname} outperforms the state-of-the-art in 59\% and 43\% of the experiments respectively
    \footnote{The code is available at \url{https://github.com/aminparvaneh/alpha_mix_active_learning}.}.
\end{abstract}

\vspace{-5mm}
\section{Introduction}\label{intro}
\vspace{-2mm}
	
\begin{figure}[t]
    \centering
    \includegraphics[width=\linewidth]{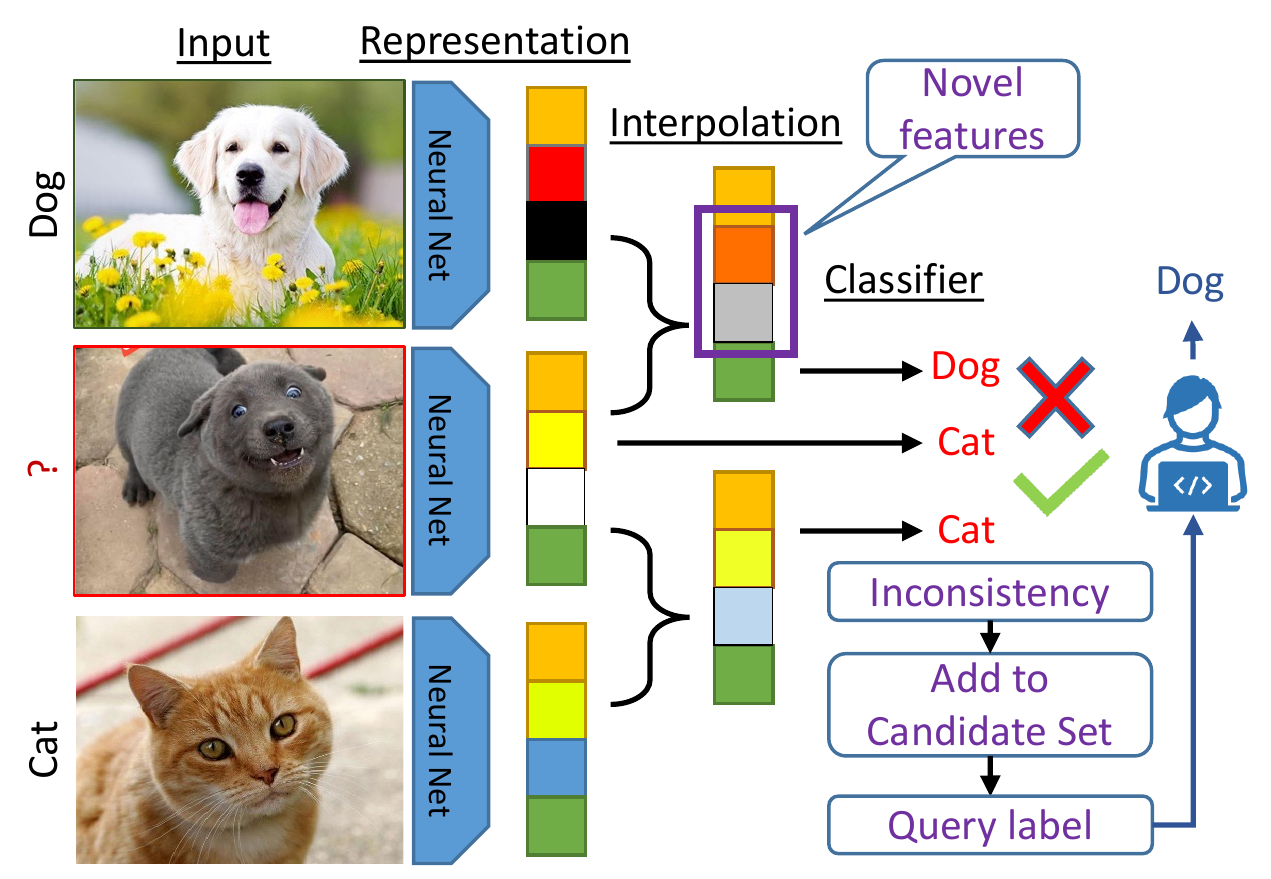}
    \vspace{-4mm}
    \caption{We propose to form linear combinations (\ie interpolations or mixing) of the features of an unlabelled instance (middle image) and of labelled ones (top and bottom images).
    The interpolated features are passed through the current classifier.
    We show that inconsistencies in the predicted labels indicate that the unlabelled instance may have novel features to learn from. 
    }
    \label{fig:approach}
    \vspace{-8mm}
\end{figure}

The success of machine learning applications depends on the quality and volume of the annotated datasets. High quality data annotations can be slow and expensive.  
Active learning (AL) aims to actively select the most valuable samples to be labelled in the training process iteratively, to boost the predictive performance.  
A popular setting called \emph{batch}~ AL~\cite{settles2009active} fixes a budget on the size of the batch of instances to be sent to an oracle for labelling.
The process is repeated over multiple rounds, allowing the model to be updated iteratively.
The core challenge is therefore to identify the most valuable instances to be included in this batch at each round, depending on the current model.

\begin{figure*}[t]
	\centering
	\begin{subfigure}{0.34\linewidth+2mm}
		\centering
		\includegraphics[width=\linewidth]{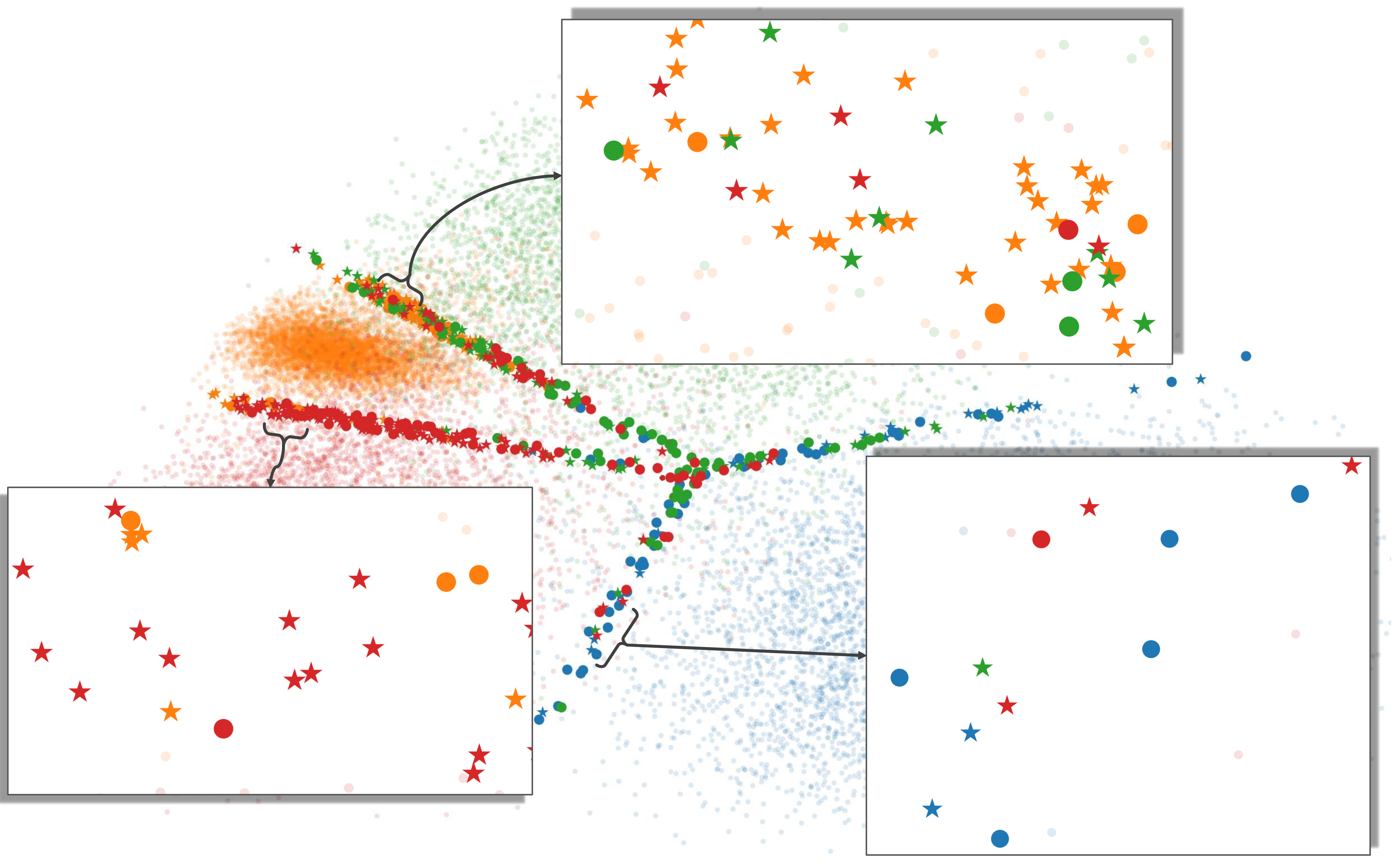}
		\vspace{-5mm}
		\caption{\textbf{\methodname} (ours)}
		\label{fig:plot_minimnist_ours}
	\end{subfigure}	
	\hspace{-2mm}	
	\begin{subfigure}{0.34\linewidth+0mm}
		\centering
		\includegraphics[width=\linewidth]{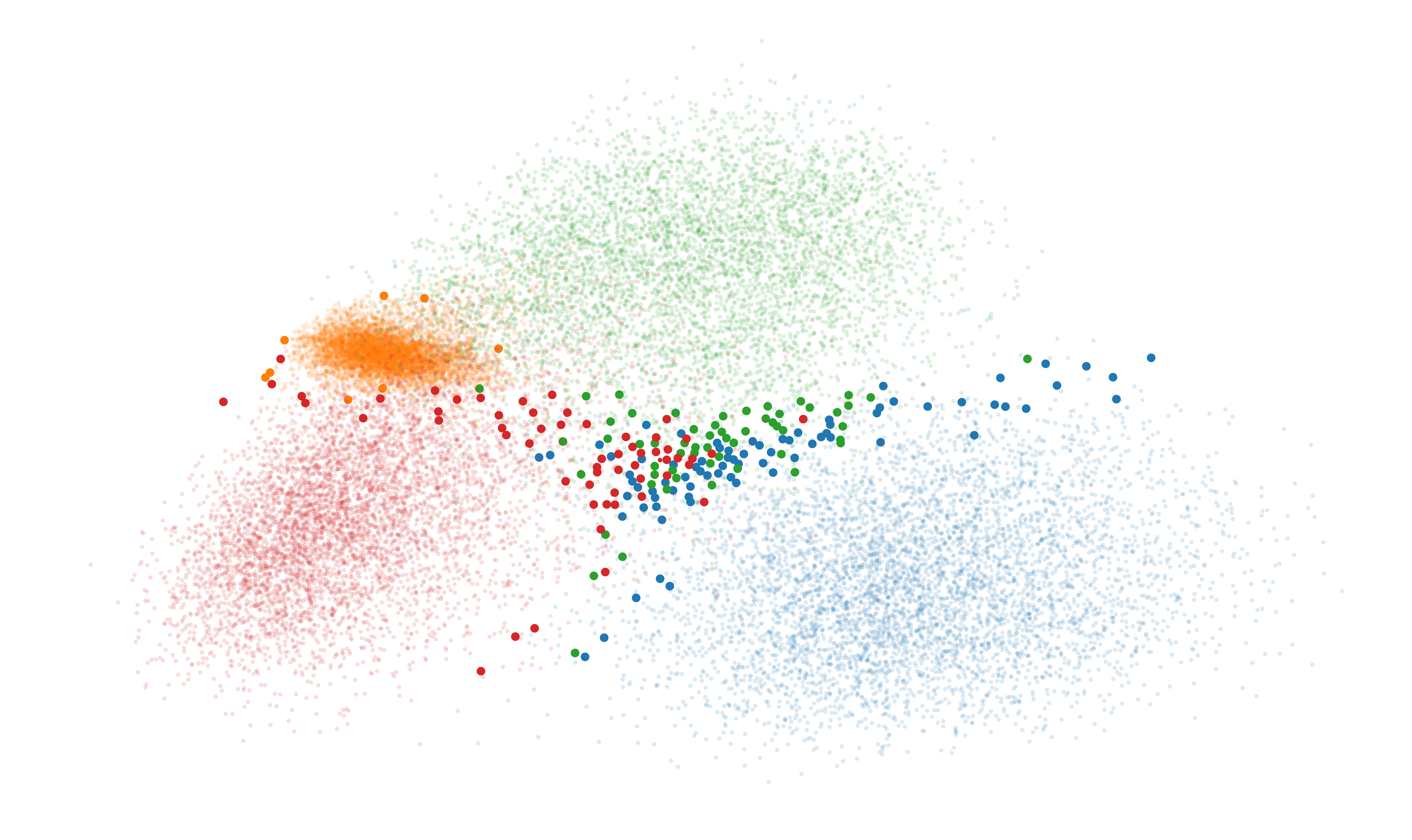}
		\vspace{-5mm}
		\caption{CDAL (ECCV 2020)~\cite{cdal_2020}}
		\label{fig:plot_minimnist_cdal}
	\end{subfigure}	
	\hspace{-10mm}
	\begin{subfigure}{0.34\linewidth+2mm}
		\centering
		\includegraphics[width=\linewidth]{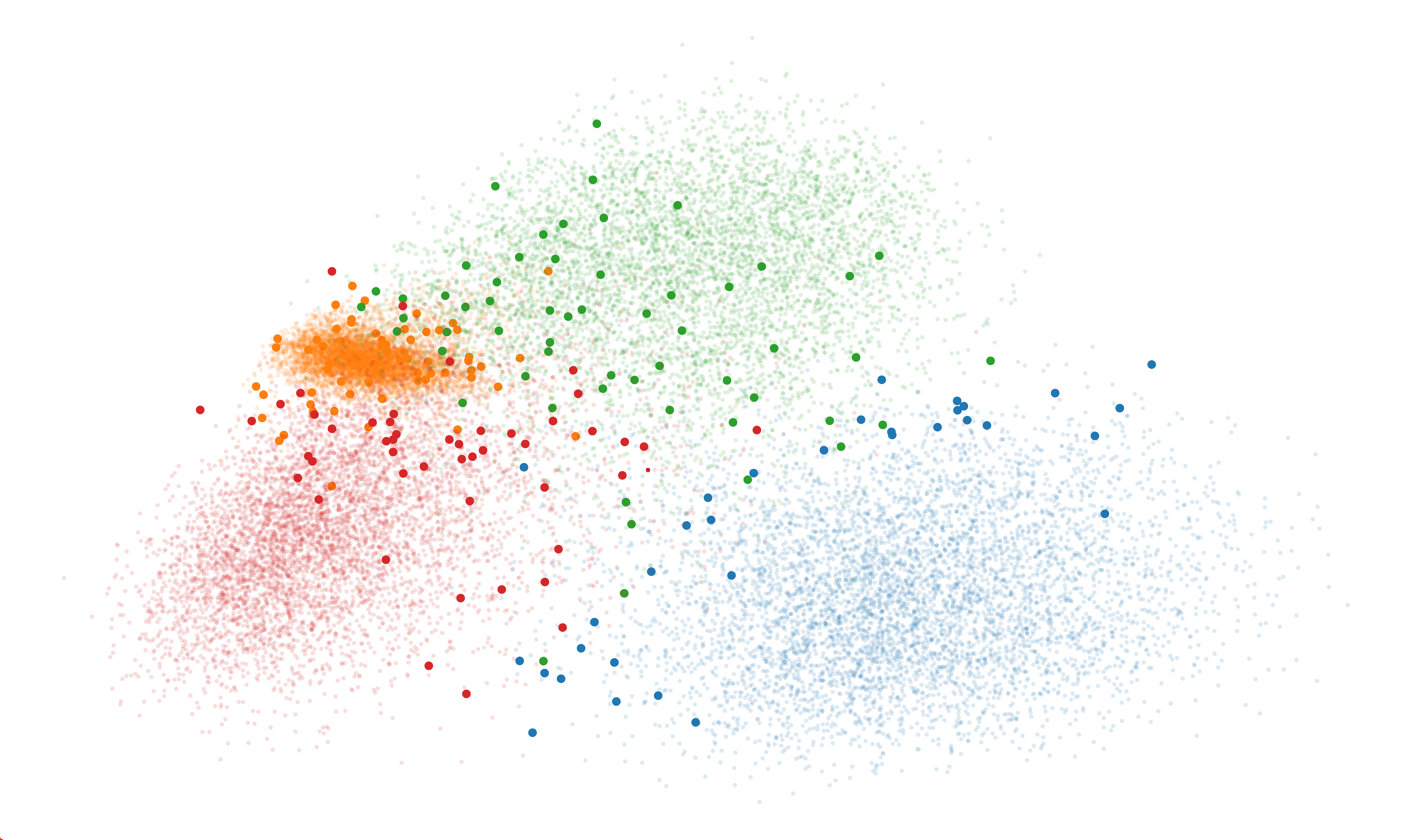}
		\vspace{-5mm}
		\caption{BADGE (ICLR 2020)~\cite{badge_iclr_2020}}
		\label{fig:plot_minimnist_badge}
	\end{subfigure}	
	\hspace{-10mm}
	\begin{subfigure}{0.34\linewidth+4mm}
		\centering
		\includegraphics[width=\linewidth]{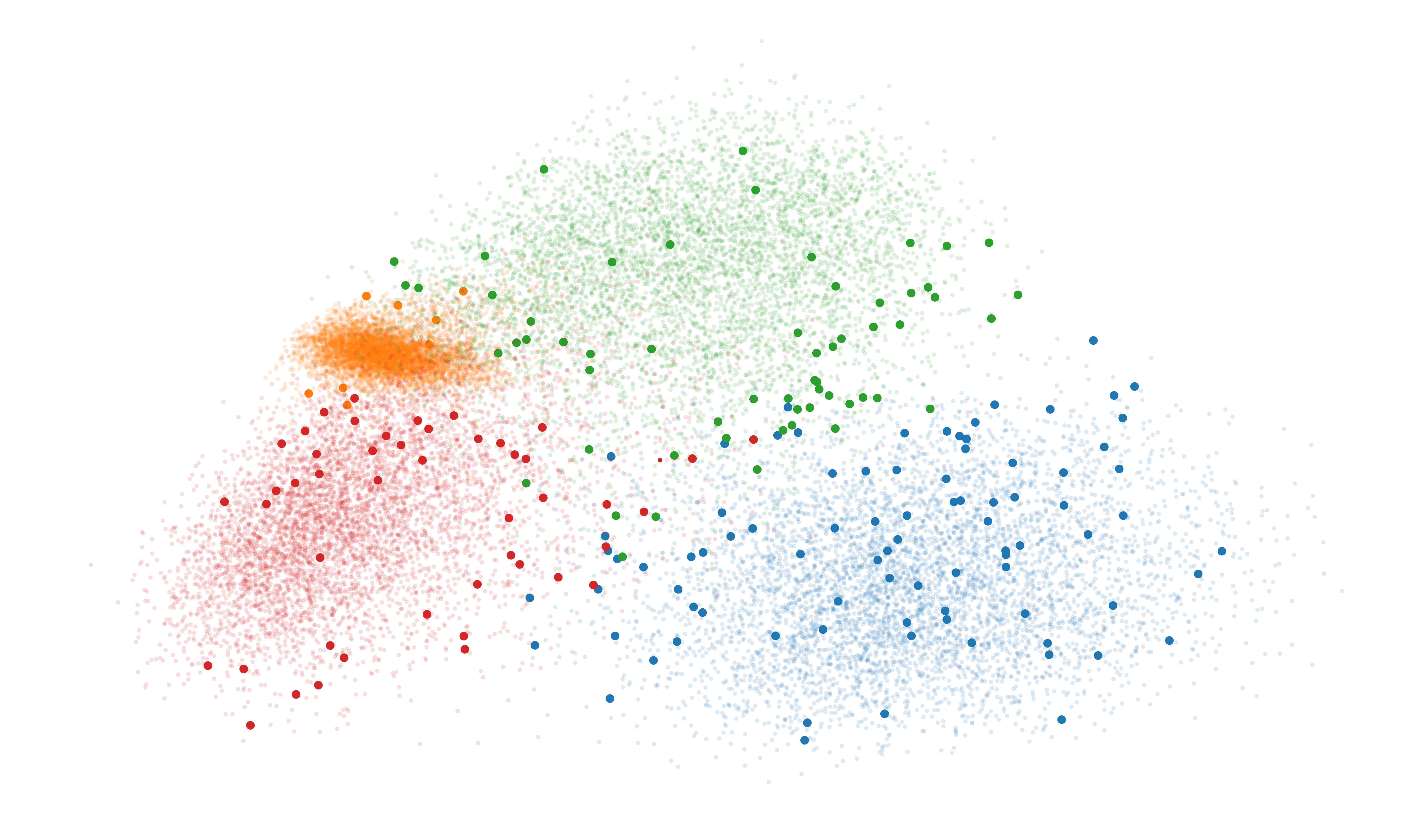}
		\vspace{-5mm}
		\caption{GCNAL (CVPR 2021)~\cite{gcn_2021_cvpr}}
		\label{fig:plot_minimnist_gcn}
	\end{subfigure}	
	\hspace{-10mm}
	\begin{subfigure}{0.34\linewidth+0mm}
		\centering
		\includegraphics[width=\linewidth]{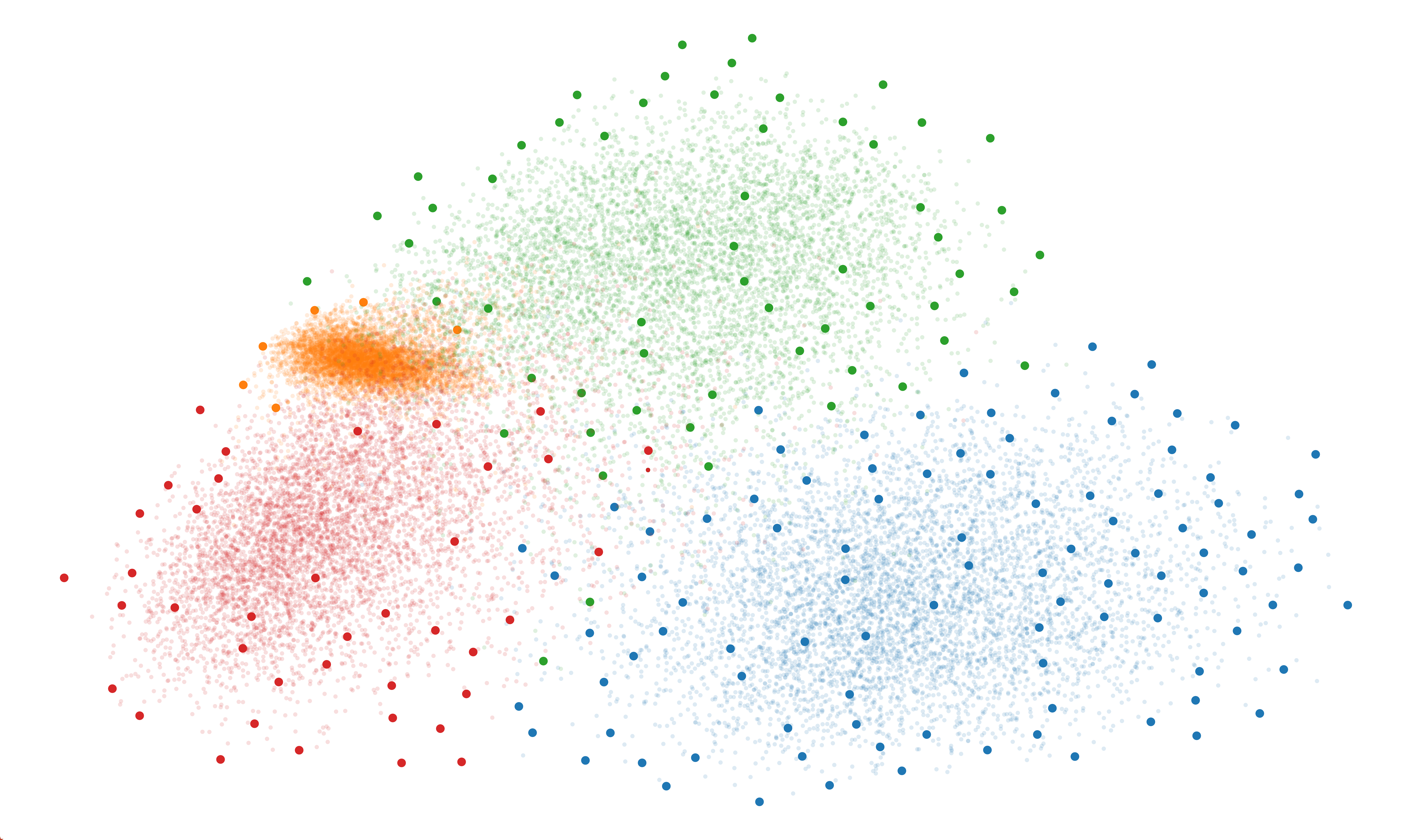}
		\vspace{-5mm}
		\caption{CoreSet (ICLR 2018)~\cite{coreset_iclr_2018}}
		\label{fig:plot_minimnist_coreset}
	\end{subfigure}
	\hspace{-10mm}
	\begin{subfigure}{0.34\linewidth+4mm}
		\centering
		\includegraphics[width=\linewidth]{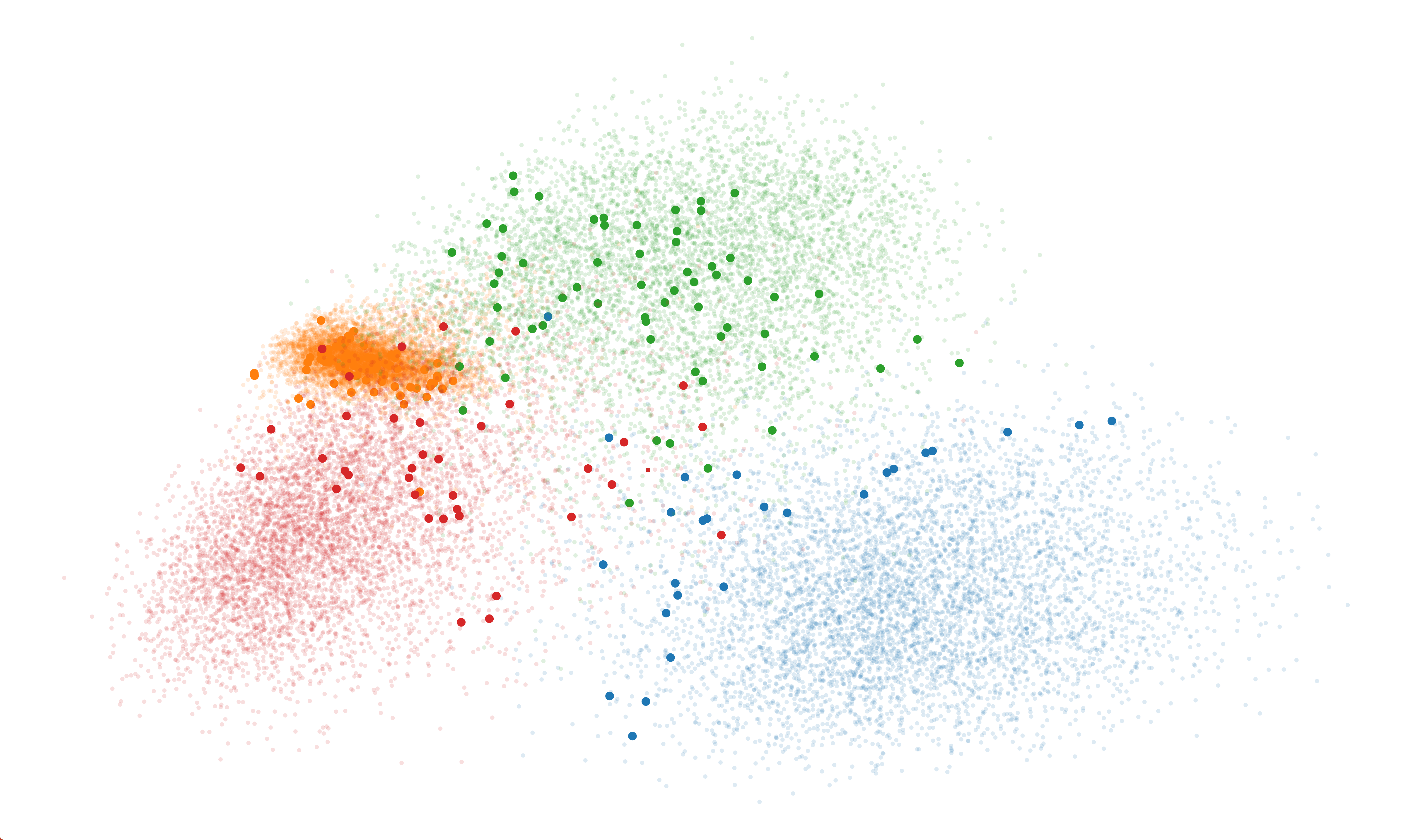}
		\vspace{-5mm}
		\caption{BALD (ICML 2017)~\cite{bald_2017}}
		\label{fig:plot_minimnist_bald}
	\end{subfigure}
	\vspace{-3mm}
	\caption{\small{Visualization of sample selection behaviours of various AL methods in the latent space (see the Appendix for additional methods). The larger dots represent the selected samples to label; smaller dots represent  unlabelled ones. Our approach finds a candidate set (demonstrated by stars in \ref{fig:plot_minimnist_ours}) of unlabelled instances with inconsistencies in their label prediction when interpolated with labelled representations. It selects a diverse set of samples lying close to the all four borders for the labelling (with three zoom-in windows). 
	The demonstration problem is that of identifying 4 classes from MNIST (illustrated above by 4 colours) using a MLP. An initial training set of 200 randomly selected points and their labels was provided, with each method given a budget of 200 additional labels.
	The features are projected to two-dimensions for visualization. }}
	\label{fig:plots_minimnist}
	\vspace{-5mm}
\end{figure*}

Various AL strategies have been proposed differing in predicting (1)~how informative a particular unlabelled instance will be (\ie uncertainty estimation~\cite{margin_2006, entropy_2014, bald_2017, adversarialDFAL_2018})
or (2)~how varied a set of instances will be (\ie diversity estimation~\cite{diversity_2015, coreset_iclr_2018}), or both~\cite{alqir_2010,albl_2015,cdal_2020}. 
Recent deep learning based AL techniques include, for example, the use of an auxiliary network to estimate the loss of unlabelled instances~\cite{llal_2019}, the use of generative models like VAEs to capture distributional differences~\cite{vaal_2019, taskaware_2021_CVPR}, and the use of graph convolutional networks to relate unlabelled and labelled instances~\cite{gcn_2021_cvpr}.

Despite much progress made, current AL methods still struggle when applied to deep neural networks, with high-dimensional data, and in a low-data regime. 
We hypothesised that the representations learned within deep neural networks may be leveraged to reason about the model's uncertainty while alleviating the challenges associated with high-dimensional data.
Some existing methods only consider the model's output, but we believe that this cannot convey a complete picture of the model's current state.
Assessing the uncertainty in the model is particularly important in a low-data regime since the number of available training examples is small.
This motivation has led to methods like BADGE~\cite{badge_iclr_2020} which uses gradients through the classifier layer of the network. Besides its relatively poor performance in lo-data regimes~\cite{badge_iclr_2020}, the drawback is a high computational cost due to the high dimensionality of the gradient embeddings, making the method impractical for deep models with latent representations of high dimensions, large datasets, and large numbers of classes.   

In this paper, we present a novel and efficient AL method, named \underline{A}ctive \underline{L}earning by \underline{F}e\underline{A}ture Mixing (\methodname), based on the manipulation of latent representations of the data. 
We identify informative unlabelled instances by evaluating the variability of the labels predicted for perturbed versions of these instances.
These perturbed versions are instantiated in feature space as convex combinations of unlabelled and labelled instances (see Figure~\ref{fig:approach}). 
This approach effectively explores the neighbourhood surrounding an unlabelled instance by interpolating its features with those of previously-labelled ones.
Convex combinations of features have been already used in other contexts such as data augmentation, using random interpolations~\cite{mixup_2018,manifold_mixup_2019, int_cons_2019, mixstyle_2021} or actual solutions to an optimisation problem \cite{counterfactual_2020,counterfactual_vln_2020}.

We provide a theoretical support for the method.
In particular, under a norm-constraint on the interpolation ratio, we show that the interpolation is equivalent to considering (1)~{the difference between the features of the unlabelled instance and the labelled ones} and (2)~{the gradient of the model w.r.t the features at the unlabelled point}.
Discovering new features considering (1) and (2) leads us to finding {an optimal interpolated point} deterministically, at a minimal computing cost.
Rather than using all the labelled data for these interpolations, we choose a subset we call anchors to capture the common features for each class. Subsequently, we construct a {candidate set} by choosing the instances from the unlabelled set that when mixed with these anchors lead to a change in the model's prediction for those instances.
Then, to ensure selected instances are diverse, we perform a simple clustering in the candidate set and choose their centroids as the points to be queried. 

The contributions of this paper are as follows.
\begin{itemize}[leftmargin=12pt,noitemsep,topsep=0pt,parsep=2pt]
\item Instead of interrogating an unlabelled instance directly, we interpolate its representation features from the labelled instances to uncover its hidden traits. To the best of our knowledge, it is the first of its kind in AL. Unlike existing methods that reply solely on the predicted output, we harness useful information from the feature representations as an indication of which features are novel for the model. 
\item We show that optimal interpolation/mixing for each instance that underscores the novel features with which the model could change prediction, has a closed-form solution making our approach efficient and scalable.
\item We show that our approach outperforms its counterparts
over 9 image, 2 OpenML, and one video datasets in various
settings of architecture, network initialisation, and budget
choice. Our approach consistently achieves higher accuracy than existing methods, with particularly significant
gains in a low-data regime.
\item We provide the first investigation into using AL in vision transformers: we demonstrate the effectiveness of \methodname~on a self-trained vision transformer \cite{dino_iccv_2021}, performing better than random selection in all tests, and 43\% better than the state-of-the-art. In addition, our approach performs significantly better that its counterparts for video classification using transformers \cite{mvit_2021}. 
\end{itemize}

\vspace{-2mm}
\section{Related Work}
\label{related_work}
\vspace{-2mm}

	Active learning strategies can be broadly categorised into three types: diversity-based, uncertainty-based, and hybrid sampling, according to the nature of their acquisition function. Diversity-based approaches aim to select samples that best represent the whole of the available unlabelled set. A variety of approaches have been proposed that cluster the unlabelled samples based on feature representations \cite{diversity_2015}, or construct a core-set over the latent features to identify a suitably diverse set of samples \cite{coreset_iclr_2018}. 
	
	Uncertainty-based methods seek to identify the unlabelled samples that are most ambiguous to the current model that has been trained over the present labelled set based on the target objective function. The assumption here is that having these uncertain samples labelled will add the most value to the next model training round. Entropy and the confidence of the predictions \cite{entropy_2014}, the margin between the confidence of the highest and second highest predicted classes \cite{margin_2006}, the information gain in the model parameters in a Bayesian framework \cite{bald_2017}, and the variance between the predicted probabilities within the ensemble \cite{ensemble_208} have all been proposed as measures of uncertainty. 
	These methods favour points that lie close to the decision boundary, but as they rely entirely on the predicted class likelihoods they ignore the value of the feature representation itself.
	The closest method to that which we propose here is the deep fool attack learning (DFAL) approach~\cite{adversarialDFAL_2018} where 
	the distance to the decision boundary is approximated by perturbation, using techniques originally developed for adversarial attacks~\cite{deep_fool_2016_CVPR}. Adversarial examples may expose vulnerability of the network architecture to particular patterns in the input rather than the distribution of the labels over latent space. That may lead to incorrect selection of instances that have patterns that are easily manipulated rather than helping to shape a more consistent decision boundary. 
	Random perturbations are unlikely to lie within the true data distribution, and thus risk wasting labelling cost on feature values that can never arise in practice.
	Rather than repeatedly adding random noise in the input space, the method we propose here (\methodname) interpolates in latent space.  {\methodname} is not only faster, it also significantly outperforms the DFAL approach.
	
	Recently, a series of model-based active learning have been developed whereby a separate model is trained for active instance selection. Various objectives, either task-agnostic (\eg variational adversarial active learning~\cite{vaal_2019}, graph convolutional active learning~\cite{gcn_2021_cvpr}) or task-aware (\eg target loss prediction~\cite{llal_2019}), have been proposed as for  training  these models. Additionally, \cite{vabal_2021} has married model-based algorithms with conventional ones by combining a variational Bayes network with feature representations from the target model. In addition to sensitivity to hyper-parameters and additional computational cost, these AL methods do not consider the diversity of the selected samples and are prone to selecting samples with repetitive patterns. Moreover, our experiments show their poor performances in low-data regime.
		
	Hybrid AL methods exploit both diversity and uncertainty in their sample selection methodologies. A mini-max strategy was proposed in \cite{alqir_2010}, for example, that maximises both the informativeness and representativeness of the samples.  Interestingly, a method that learns to combine different AL strategies was presented in \cite{albl_2015}.  Additionally, \cite{cdal_2020} exploits the predicted probabilities in images to select samples from diverse contexts (\ie images of objects with varied backgrounds). 
	Recently, \cite{badge_iclr_2020} proposed to cluster the gradients of the final output layer of the target model as the features of the unlabelled samples that implicitly encompass the uncertainty information. Despite their state-of-the-art results on some image and non-image datasets, their approach is not scalable to larger tasks with numerous number of classes. Our approach not only consistently outperforms their method by a large margin in different settings, but it also is extremely efficient and scalable to large tasks.
	
	\vspace{-2mm}
	\section{Methodology}\label{sec:method}
	
	\vspace{-1mm}
	\subsection{Problem Definition}
	\vspace{-1mm}
	
	Without loss of generality, we consider our learning objective to be training a supervised multiclass classification problem with $\lblsize$ classes. A learner is actively trained in iterations of interactions with an oracle.
	At each iteration, this active learner has access to a small set of labelled data 
	$\lblData=\{(\inp_i,\lbl_i)\}_{i=0}^M$
	where $\inp_i\in\mathcal{X}$ represents the input (\eg an image or a video clip) and $\lbl_i\in\{1,\ldots,K\}$ stands for the associated class label. The learner also has access to a set of unlabelled data $\ulblData$ from which $\Bgt$ number of instances are chosen to be labelled by the oracle.
	The labelled samples are then added to $\lblData$ to update the model.
    The performance of the model is evaluated on an unseen test dataset. 
	
	The learner is a deep neural network $f=f_c\odot f_e$ parameterised by $\params=\{\encparams,\clsparams\}$. Here, $f_e: \mathcal{X}\to\R^D$ is the backbone which encodes the input to a $D$-dimensional representation in a latent space, \ie $\ltn=f_e(\inp;\encparams)$. Further, $f_c:\R^D\to\R^K$ is a classifier \eg multi-layer perceptron (MLP) that maps the instances from their representations to their corresponding logits which can be converted to class 
	likelihoods 
	by $p(\lbl\mid\ltn;\params)=\softmax(f_c(\ltn;\clsparams))$.
	We optimise the parameters end-to-end by minimising the cross-entropy loss over the labelled set: $\Expec_{(\inp,\lbl)\sim\lblData}[\loss(f_c\odot f_e(\inp;\params),\lbl)]$.
	The prediction of the label (\ie pseudo-label) for an unseen instance is  $\lblmaxarg{\ltn}=
	\arg\max_\lbl f_c^\lbl(\ltn;\clsparams)$ where $\ltn=f_e(\inp;\encparams)$ and $f_c^\lbl$ is the logit output for class $\lbl$. Additionally, 
	the logit of the predicted label is denoted as $f_c^*(\ltn):=f_c^\lblmaxarg{\ltn}(\ltn)$\footnote{For brevity, when the parameters $\encparams$ and $\clsparams$ are clear from the context, we refrain from explicitly including them.}. We also denote $\Ltnu=\{f_e(\inp), \forall\inp\in\ulblData\}$ the set for representations of the unlabelled data and $\Ltnl$ its labelled counterpart.
    We compute the average representation $\prt$ of the labelled samples per class, and call it anchor. The anchors for all classes form the anchor set $\Prt$, and serve as representatives of the labelled instances. 
	
	\vspace{-1mm}
	\subsection{Feature Mixing} \label{sec:interpolate}
	\vspace{-1mm}
	The characteristics of the latent space plays a crucial role in identifying the most valuable samples to be labelled. Our intuition is that the model's incorrect prediction is mainly due to novel "features" in the input that are not recognisable. 
	Thus, we approach the AL problem by first probing the features learned by the model.  
	To that end, we use a convex combination (\ie interpolation) of the features as a way to explore novel features in the vicinity of each unlabelled point. 
    Formally, we consider our interpolation between the representations of the unlabelled and labelled instances, $\ltnu$ and $\prt$ respectively (we use the labelled anchor here for efficiency) as $\intp=\bnoise\prt+(1-\bnoise)\ltnu$ using an interpolation ratio $\bnoise\in[0,1)^D$. 
    This process can be seen as a way of sampling a new instance without explicitly modelling the joint probability of the labelled and unlabelled instances \cite{mixup_2018,genint,counterfactual_2020,counterfactual_vln_2020}, \ie	{\small\begin{align}
	    \ltn\sim p(\ltn\mid \ltnu, \Prt,\bnoise)\equiv\bnoise\prt+(1-\bnoise)\ltnu,\ \ \prt\sim\Prt.
	    \label{eq:gen}
	\end{align}}
	We consider interpolating an unlabelled instance with all the anchors representing different classes to uncover the sufficiently distinct features by considering how the model's prediction changes. For that, we investigate the change in the pseudo-label (\ie $\lblmax$) for the unlabelled instance and the loss incurred with the interpolation. We expect that a small enough interpolation with the labelled data should not have a consequential effect on the predicted label for each unlabelled point. 
	
	Using a first-order Taylor expansion w.r.t. $\ltnu$, the model's loss for predicting the pseudo-label of an unlabelled instance at its interpolation  with a labelled one can be re-written as\footnote{This statement is true for any given instance and any convex combination of points in the latent space. For AL, we particularly focus on unlabelled instances though. More details are provided in the Supplements.}:
	\begin{align}
	\loss\left(f_c\left(\intp\right),\lblmax\right)\,\approx\,&\loss\left(f_c(\ltnu), \lblmax\right) + \label{eq:pred_change}\\  
	&\left(\bnoise(\prt-\ltnu)\right)^{\intercal} . \nabla_{\ltnu} \loss\left(f_c\left(\ltnu\right), \lblmax\right)\,, \notag
	\end{align}
	which for a sufficiently small $\bnoise$, \eg $\|\bnoise\|\leq \epsilon$ is almost exact.
	Consequently, for the full labelled set, by choosing the max loss from both sides we have:
	\begin{align}
	&\max_{\prt\sim\Prt}\left[\loss\left(f_c\left(\intp\right),\lblmax\right)\right]-\loss\left(f_c(\ltnu), \lblmax\right)\approx\label{eq:loss_change}\\
	&~~~\qquad\max_{\prt\sim\Prt}\left[\left(\bnoise(\prt-\ltnu)\right)^{\intercal} . \nabla_{\ltnu} \loss\left(f_c\left(\ltnu\right),\lblmax\right)\right]. \notag
	\end{align}
	
	Intuitively, when performing interpolation, the change in the loss is proportionate to two terms: (a) the difference of features of $\prt$ and $\ltnu$ proportionate to their interpolation $\bnoise$, and (b) the gradient of the loss w.r.t the unlabelled instance. The former determines which features are novel and how their value could be different between the labelled and unlabelled instance. On the other hand, the later determines the sensitivity of the model to those features. That is, if the features of the labelled and unlabelled instances are completely different but the model is reasonably consistent, there is  ultimately no change in the loss, and hence those features are not considered novel to the model. 
	
	The choice of $\bnoise$ is input specific and determines the features to be selected. 
	As such, in Sec~\ref{sec:interpolation_learn} we introduce a closed form solution for finding a suitable value for $\bnoise$.
	Finally, we note that the interpolations utilised here have some interesting properties that are further discussed in the supplements.

\vspace{-1mm}
	\subsection{Optimising the Interpolation Parameter $\bnoise$} \label{sec:interpolation_learn}
	\vspace{-2mm}
	Since manually choosing a value for $\bnoise$ is non-trivial, we devise a simple optimisation approach to choose the appropriate value for a given unlabelled instance. To that end, we note that, as observed from Eq.~\eqref{eq:loss_change}, the worst case of maximum change in the loss is when we choose $\bnoise$ that maximises the loss at the interpolation point (details are in the supplement).  However, using the r.h.s of the Eq.~\eqref{eq:loss_change}, we devise the objective for choosing $\bnoise$ as:
	\begin{align}
	\bnoiseOpt=&\argmax_{\|\bnoise\|\leq\noiseOptMax} \left(\bnoise(\prt-\ltnu)\right)^\intercal.\nabla_{\ltnu}\loss(f_c(\ltnu),\lblmax), \label{eq:noise_max}
 	\end{align}
 	where $\epsilon$ is a hyper-parameter governing the magnitude of the mixing.
	Intuitively, this optimisation chooses the hardest case of $\bnoise$ for each unlabelled instance and anchor. 
	We approximate the solution to this optimisation using dual norm formulation, which in the case of using 2-norm yields:
	\begin{equation}
	\bnoiseOpt\approx\noiseOptMax\frac{\|(\prt-\ltnu)\|_2\nabla_{\ltnu}\loss(f_c(\ltnu),\lblmax)}{\|\nabla_{\ltnu}\loss(f_c(\ltnu),\lblmax)\|_2} \oslash (\prt-\ltnu),  \label{eq:dual_opt_noise}
	\end{equation}
	where $\oslash$ represents element-wise division (further details in the Supplement). This approximation makes the optimisation of the interpolation parameter efficient and our experiments show that it will not have significant detrimental effects on the final results compared to directly optimising for $\bnoise$ to maximise the loss.
	
    \setlength{\textfloatsep}{2pt}
	{\footnotesize
		\begin{algorithm}[t]
			\caption{Our active learning algorithm.}\label{alg:main}
			
			{\bfseries Inputs:}  initial labelled set $\lblData$; unlabelled pool $\ulblData$; 
			labelling budget at each round $\Bgt$; mixing parameter $\epsilon$; 
			
			\For{$i=1$ {\bfseries to} $\text{max\_rounds}$} 
			{
				Train the model $f$ using the labelled data $\lblData$. \\
				Initialise $\Prt$ based on the representations of $\lblData$.\\
				$\Uncertain=\{\}$. \\
				\For{$\inpu\in\ulblData$} 
				{
					$\ltnu=f_e(\inpu)$.\\ 	
					\For{$\prt\in\Prt$}
					{	
    					Calculate $\bnoiseOpt$ using $\epsilon$ and  Eq.~\ref{eq:dual_opt_noise}.\\
    					$\ltnC=\bnoiseOpt\prt+(1-\bnoiseOpt)\ltn^u$.\\	\If{$\argmax_{\lbl}(f_c^\lbl(\ltn_u))\neq\argmax_{\lbl}(f_c^\lbl(\ltnC)\,)$}
    					{
    						$\Uncertain=\Uncertain\cup(\inpu,\ltnu)$. \\
    						Break
    					}
				    }
				}
				
				Cluster the samples in $\Uncertain$ into $\Bgt$ clusters.\\
				Select samples at the centre of each cluster ($\Centroid$).\\
				$\Lbl^{new}=\text{Query}(\Centroid)$.\\	$\lblData=\lblData\cup(\Centroid,\Lbl^{new}),~ \ulblData=\ulblData\,\text{\textbackslash}\,\Centroid$.
			}
		\end{algorithm}
	}
	\subsection{Candidate Selection} \label{sec:candidate}
	
	For AL it is reasonable to choose instances to be queried whose loss  substantially change with interpolation according to Eq.~\eqref{eq:loss_change}. This corresponds to those instances for which the model's prediction change and have novel features. Intuitively, as depicted in Figure.~\ref{fig:plot_minimnist_ours}, these samples are placed close to the decision boundary in the latent space. 
	Alternatively, we expect a small interpolation should not affect the model's loss when it is reasonably confident in its recognition of the features of the input. 
	We, then, create our candidate set as:
	{\small\begin{align}
	 \Uncertain = \bigg\lbrace \ltnu\in\Ltnu\,\bigg|\, \exists\prt\in\Prt,\, f_c^*( \intp)\neq\lblmaxarg{\ltnu} \bigg\rbrace.\label{eq:in_set} 
	\end{align}}
	
	The size of the selected set $\Uncertain$ could potentially be larger than the budget $\Bgt$. In addition, ideally we seek \emph{diverse} samples since most instances in $\Uncertain$ could be chosen from the same region (\ie they might share the same novel features). To that end, we propose to cluster the instances in $\Uncertain$ into $\Bgt$ groups based on their feature similarities and further choose the closest samples to the centre of each cluster to be labelled by oracle. This ensures the density of the space represented by $\Uncertain$ samples, is reasonably approximated using $\Bgt$ instances.
	We simply use $k$-MEANS which is widely accessible.
	Similar strategy is also used by \cite{badge_iclr_2020} to encourage diversity.
	Our approach is summarised in Algorithm~\ref{alg:main}.

    \begin{figure*}
        \centering
        \includegraphics[width=\textwidth]{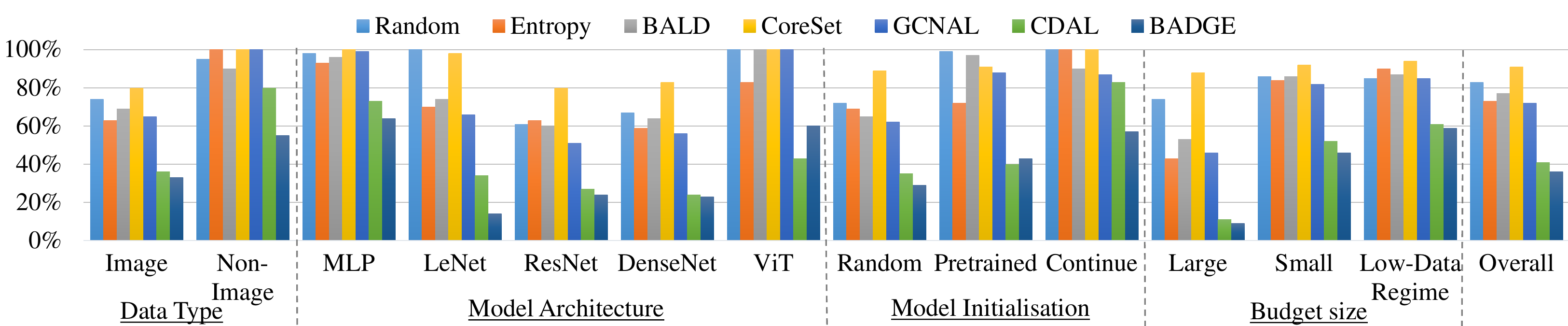}
        \vspace{-6mm}
        \caption{\small{A summary of the performance of our proposed AL method ({\methodname}) compared with state-of-the-art across all the 30 settings considered. Each bar represents the percentage of AL rounds in which our approach outperforms others (lower indicates stronger baseline). It is worth noting that our approach ({\methodname}) under-performs others in close to zero cases.}}
        \label{fig:ablation_chart}
        \vspace{-6mm}
    \end{figure*}

	\vspace{-2mm}
	\section{Experiments and Results} \label{sec:experiments}
	
	\vspace{-1mm}
	\subsection{Baselines}
	\vspace{-2mm}
	We compare {\methodname} with the following AL baselines:
	\begin{itemize}[leftmargin=6pt,noitemsep,topsep=0pt,parsep=0pt,labelsep=1pt]
		\item[--]\textbf{Random}: a simple baseline that randomly selects $\Bgt$ samples from the unlabelled pool at each round.  
		\item[--] \textbf{Entropy} \cite{entropy_2014}: A conventional AL approach that picks unlabelled instances with highest entropy.
		\item[--] \textbf{BALD} \cite{bald_2017}: An uncertainty model relying on Bayesian approaches that selects a set of samples with the highest mutual information between label predictions and posterior of the model approximated using dropout (Figure~\ref{fig:plot_minimnist_bald}).
		\item[--] \textbf{Coreset} \cite{coreset_iclr_2018}: An approach based on the core-set technique that chooses a batch of diverse representative samples of the whole unlabelled set (Figure.~\ref{fig:plot_minimnist_coreset}). 
		\item[--] \textbf{Adversarial Deep Fool} \cite{adversarialDFAL_2018}: An uncertainty method that utilises deep fool attacks to select a batch of unlabelled samples whose predictions flip with small perturbations. 
		\item[--] \textbf{GCNAL} \cite{gcn_2021_cvpr}: A model-based approach that learns a graph convolutional network to measures the relation between labelled and unlabelled instances (Figure.~\ref{fig:plot_minimnist_gcn}){\footnote{We employed CoreGCN variation in our experiments as results reported in \cite{gcn_2021_cvpr} show its superiority over the UncertainGCN version.}}.
		\item[--] \textbf{BADGE} \cite{badge_iclr_2020}: A hybrid approach that queries the centroids obtained from the clustering of the gradient embeddings (Figure.~\ref{fig:plot_minimnist_badge}).
		\item[--] \textbf{CDAL} \cite{cdal_2020}: A hybrid approach that exploits the contextual information in the predicted probabilities to choose samples with varied contexts (Figure.~\ref{fig:plot_minimnist_cdal})
	\end{itemize}
	
	\vspace{-1mm}
    \subsection{Experiment Settings}
    \vspace{-1mm}
    
    \noindent\textbf{Setting and Datasets:}
    We conducted a comprehensive set of experiments in 30 different settings on multiple datasets to evaluate how {\methodname} compares to its counterparts. We define an AL setting as a combination of a specific dataset, a limited set of initial labelled samples, a particular type of deep neural network, a limited number of AL rounds, and a fixed labelling budget (batch) for each round. 
    
    Specifically, we experimented on MNIST~\cite{mnist_1998}, EMNIST~\cite{emnist_2017}, CIFAR10~\cite{cifar10_2009}, CIFAR100~\cite{cifar10_2009}, Mini-ImageNet~\cite{miniimagenet_iclr_2017}, DomianNet-Real~\cite{domain_net_2019} as well as two subsets of DomainNet-Real for image datasets. Additionally, we extended our experiments to two more non-visual datasets from the OpenML{\footnote{\url{https://www.openml.org}}} repository.
    Furthermore, to reveal the effectiveness of each AL method in different data regimes, we utilised both small ($10 \times \lblsize$
    ) and large ($100 \times \lblsize$) budget sizes. More importantly, the network architecture and its initial parameters are two more factors that we considered in our experiments. As for the choice of the architecture, we employed different common deep neural networks; including Multi-Layer Perceptron (MLP)~\cite{badge_iclr_2020}, ResNet-18 \cite{resnet}, DenseNet-121 \cite{densenet}, as well as Vision Transformers~\cite{vit_iclr_2021}. Regarding the network initialisation, we considered three scenarios where at the start of each AL round{\footnote{After a new batch of samples are selected by AL method and added to the labelled set and before the model training.}}, the parameters are initialised randomly, from the model trained in the previous round (denoted as "Continue" in Figure.~\ref{fig:ablation_chart}), or using pre-trained models (either from supervised or self-supervised~\cite{dino_iccv_2021} pre-training on ImageNet~\cite{imagenet_cvpr09}). Please find for more details in the Appendix.
    
	We followed the supervised training setting proposed in \cite{badge_iclr_2020} and optimised the network using all the labelled set (without any validation set) based on a cross-entropy loss and an Adam optimiser with a learning rate of $1e-3$ and $1e-4$ for image and non-image datasets, respectively. Similarly, we continued the training using a batch size of 64 until the model reaches a certain early stopping condition (\ie reaching a training accuracy above $99\%$~\cite{badge_iclr_2020}). 
	
	 We set the number of rounds for each setting to 10, except for DomainNet-Real where we continue for 5 rounds. Additionally, to eliminate the effect of randomness in the results, we repeated each experiment 5 times with different random seeds.
	To have a better understanding about the performance of each method, in addition to the quantitative results, we provided the penalty matrix \cite{badge_iclr_2020} that facilitates the pairwise comparisons between different approaches across all the settings.

    \noindent\textbf{Video classification:} Since video classification is a more challenging task with higher annotation cost, we compare the AL performance on video classification tasks.
    All the experiments are conducted on HMDB~\cite{hmdb_iccv_2011}, a widely used dataset consisting of 5,412 training videos belonging to 51 classes representing different actions. For each video, we randomly sampled a video clip with 32 frames of size $224\times 224$ using a temporal stride of $2$. Regarding the network architecture, we employed the state-of-the-art Multi-Scale Vision Transformer (MViT) backbone pre-trained on Kinetics-600~\cite{kinetics600_cvpr_2017}. Starting with a labelled set consisting of two labelled instances from each class (a total of 102 video clips), we provide each AL method with budget of the varied sizes ($2\times \lblsize$, $4\times \lblsize$, $7\times \lblsize$ and $15\times \lblsize$) in the next AL rounds. At each AL round, we train the network for 50 epochs with a batch size of 8 using AdamW~\cite{adamw_2018} optimiser with a base learning rate of $1e-4$ that warms up linearly for the first 30 epochs and then decays to $5e-5$ for the rest of the iterations using a cosine scheduler~\cite{cosinelr_iclr_2017}. We repeated each experiment twice to cancel out the effect of random selection of the initial labelled set.
    
	\noindent \begin{figure}
	\centering
    \includegraphics[width=0.9\linewidth]{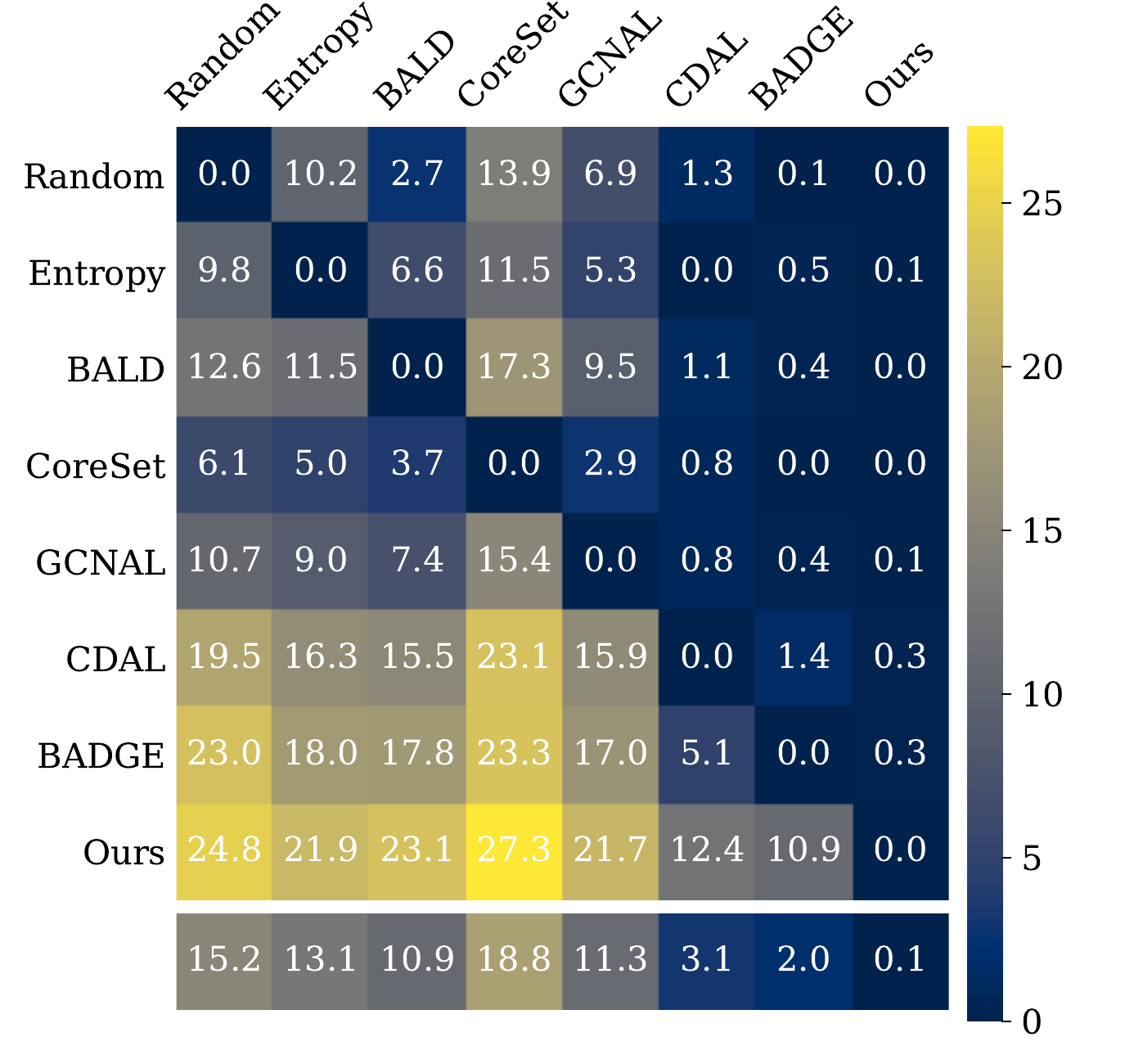}
    \vspace{-5mm}
    \caption{{\small{Pairwise comparison \cite{badge_iclr_2020} of different approaches. 
    Lower values shown at each column reveal the better performances of that AL method across all the experiments. Maximum value of each cell is 30, which represents the number of experimental settings.
    }}}
    \label{fig:comp_all}
    \end{figure}
    
	\begin{figure*}[t]
	    \vspace{-4mm}
	    \centering
		\begin{subfigure}{0.31\textwidth}
			\centering
			\includegraphics[width=1.\textwidth]{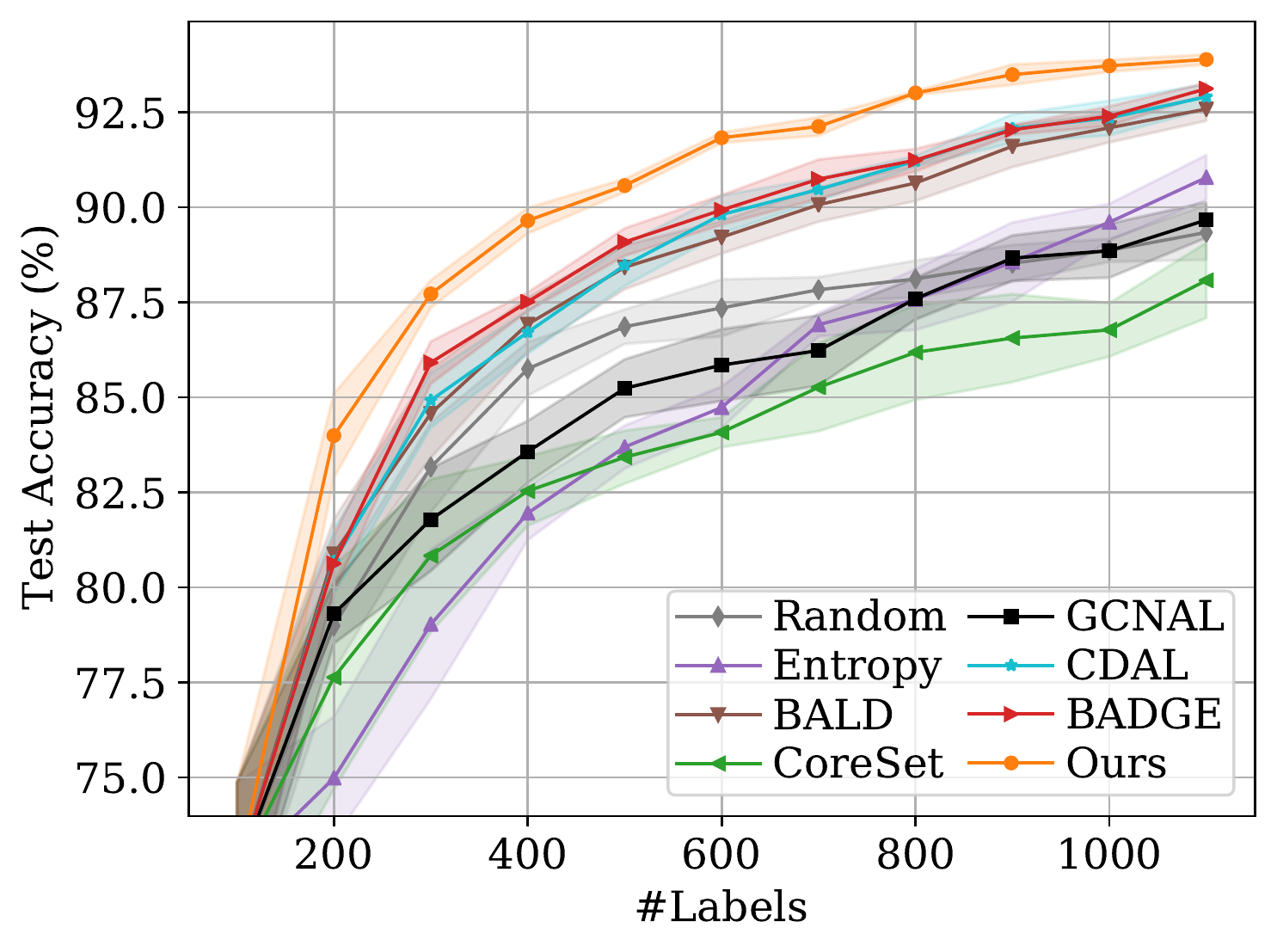}
			\vspace{-4mm}
			\caption{\small{MNIST (MLP)}}
			\label{fig:plot_mnist_mlp}
		\end{subfigure}	
		\begin{subfigure}{0.31\textwidth}
			\centering
			\includegraphics[width=1.\textwidth]{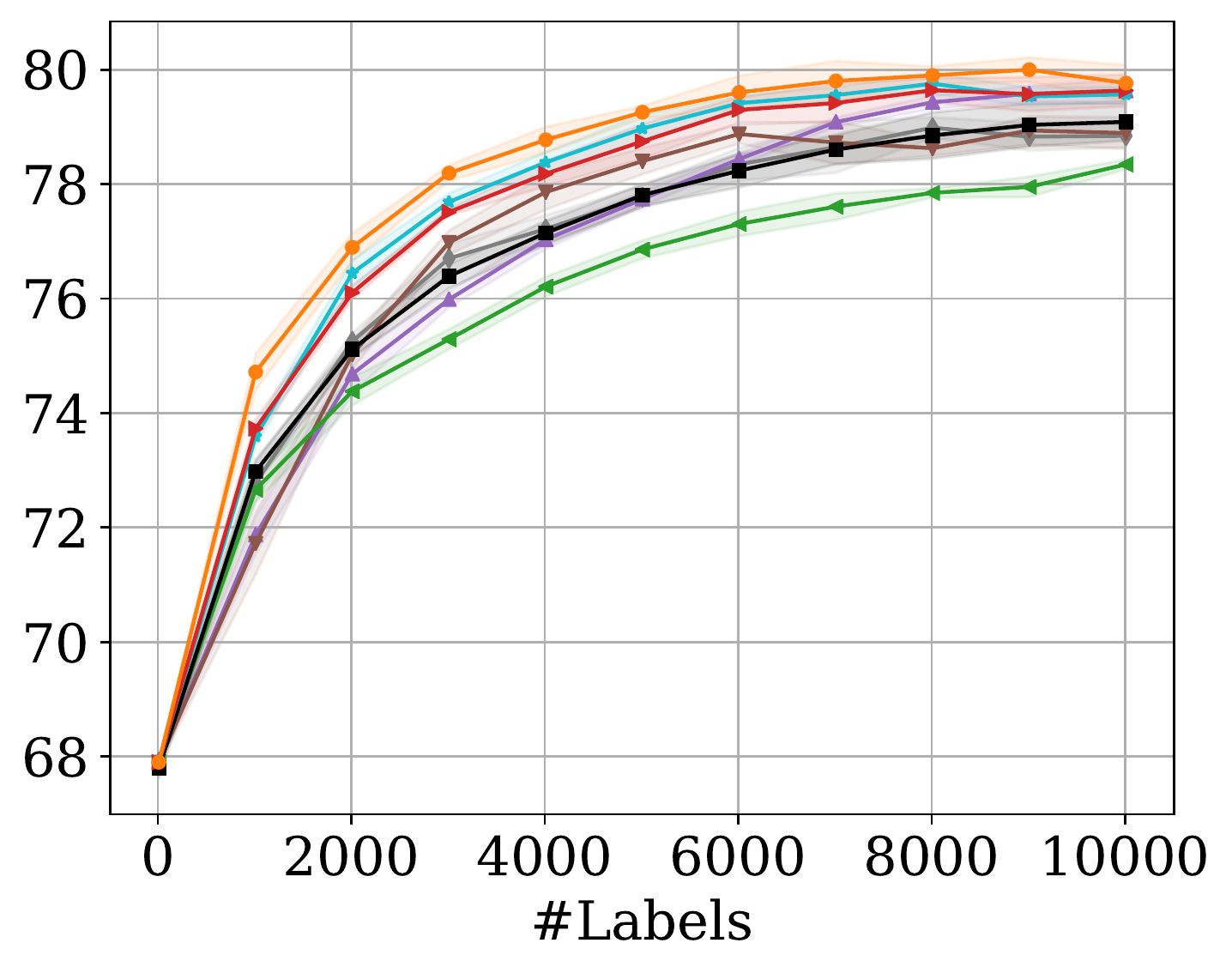}
			\vspace{-5mm}
			\caption{\small{MiniImageNet (ViT-Small)}}
			\label{fig:plot_cifar10_resnet}
		\end{subfigure}	
		\begin{subfigure}{0.31\textwidth}
			\centering
			\includegraphics[width=1.\textwidth]{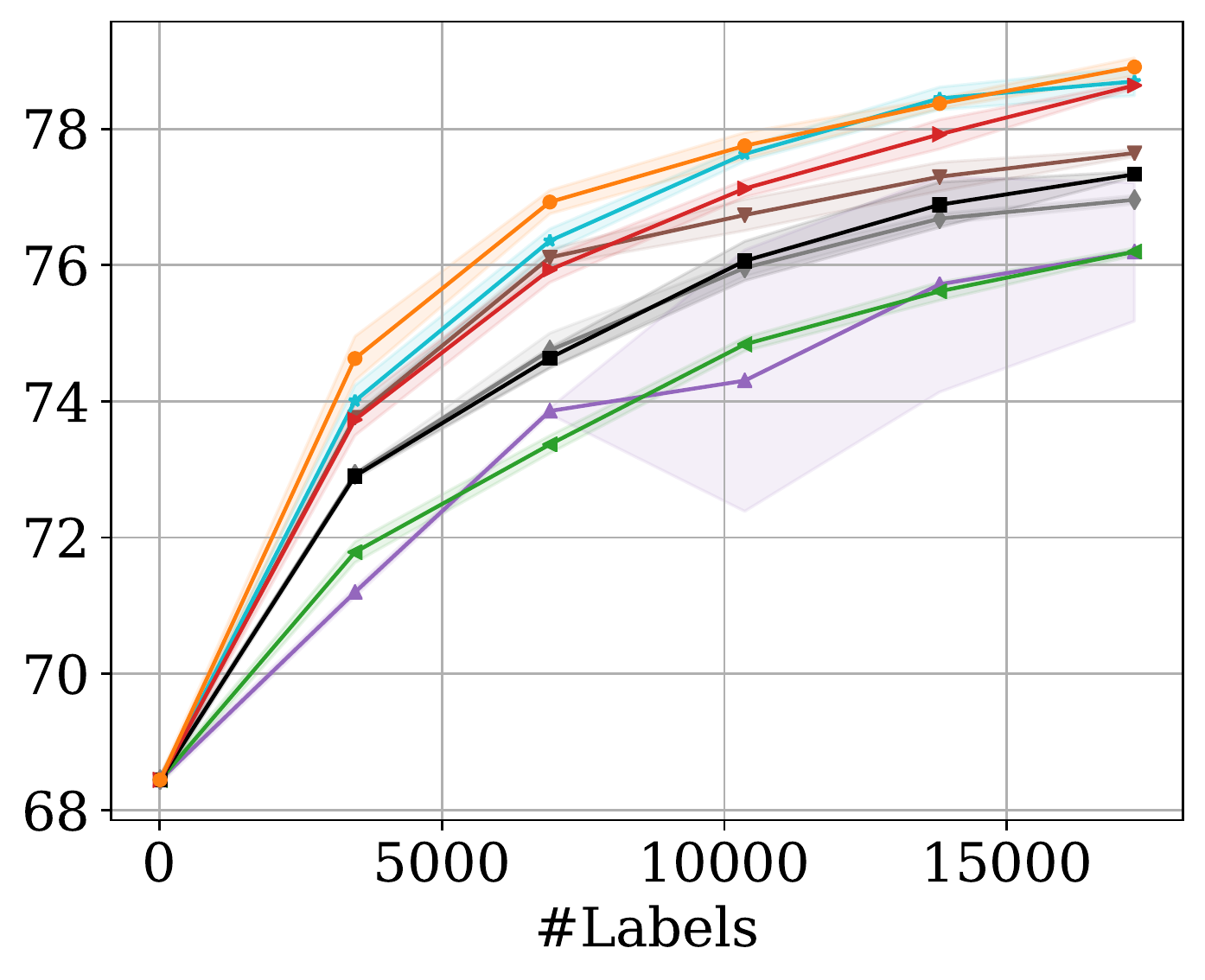}
			\vspace{-4mm}
			\caption{\small{DomainNet-Real (ViT-Base)}}
			\label{fig:plot_domain_net_densenet}
		\end{subfigure}
		\vspace{-3mm}
		\caption{\small{Test accuracy plots across some of the employed settings. Each experiment has been repeated 5 times. 
		}}
		\label{fig:acc_plots}
		\vspace{-4mm}
	\end{figure*}
	
	\vspace{-2mm}
    \noindent\textbf{Interpolation optimisation:}
	In our approach, we set $\epsilon=\frac{0.2}{\sqrt{D}}$, where $D$ is the dimentionality of $\bnoise$ vector. 
	Considering the norm condition in Eq.~\ref{eq:noise_max}, we relate the scale of $\epsilon$ to $D$ to easily utilise the same hyper-parameter across different networks with representations of variable dimensions.
	
    \vspace{-2mm}
	\subsection{Overall Results}
	\vspace{-2mm}
	
	\noindent \textbf{Image and non-image results.} In Figure.~\ref{fig:comp_all} we summarise all the results across various datasets, budget sizes and architectures (30 different settings in total) for image and non-image tasks into a matrix $C$. 
	While each element $C_{i,j}$ in the matrix reveals in how many experiments the method shown in row $i$ outperforms the one in column $j$ in terms of accuracy of an unseen test set (higher is better for the approach shown in the row). The last row indicates the average number of experiments in which the method in the column has been outperformed by others (lower is better). The maximum value for each cell in the matrix is $30$. This matrix clearly shows the superior performance of our approach compared to the baselines.
	In particular, we outperform CDAL~\cite{cdal_2020} and BADGE~\cite{badge_iclr_2020} in a significant number of experiments (12.3 and 10.6 out of 30, respectively) but ours under-performed in only 0.3 of the times. Generally as shown in the last column, our approach is rarely outperformed (lower than 0.3).
    In other words, except in 3 AL rounds, for the rest of 282 ones (around $99\%$ of the rounds), our approach is capable of matching or outperforming the best-performing baselines (BADGE and CDAL).

	\begin{table}
	 	\setlength{\tabcolsep}{2.pt}
	 	\centering
	 	\resizebox{0.89\linewidth}{!}{
	 		\noindent
	 		\begin{tabular}
	 			{p{2.2cm}>{\centering\arraybackslash}p{1.35cm}>{\centering\arraybackslash}p{1.35cm}>{\centering\arraybackslash}p{1.35cm}>{\centering\arraybackslash}p{1.35cm}}
	 			\toprule
	 			& \multicolumn{4}{c}{\textbf{AL Rounds}} \tabularnewline
	 			\cmidrule(lr){2-5} 
	 			\textbf{Method}&\textbf{204*}&\textbf{408}&\textbf{765}&\textbf{1530}\\
	 			\midrule
	 			\multicolumn{5}{l}{\underline{\textbf{MViT}} (initial accuracy with 102 instances: $50.9\scriptstyle\pm1.2$)}\\
	 			\quad Random	& $56.7\scriptstyle\pm1.4$	& $64.1\scriptstyle\pm1.2$	& $72.0\scriptstyle\pm1.1$	& $75.3\scriptstyle\pm0.4$ \\
	 			\quad Entropy~\cite{entropy_2014}	& $55.5\scriptstyle\pm0.6$	& $65.5\scriptstyle\pm0.3$	& $70.2\scriptstyle\pm2.0$	& $76.5\scriptstyle\pm0.7$ \\ 
	 			\quad BALD~\cite{bald_2017}	& $56.7\scriptstyle\pm0.4$	& $65.5\scriptstyle\pm0.6$	& $72.4\scriptstyle\pm1.3$	& $76.6\scriptstyle\pm1.8$ \\ 
	 			\quad CoreSet~\cite{coreset_iclr_2018}	& $59.3\scriptstyle\pm1.3$	& $65.8\scriptstyle\pm1.2$	& $72.8\scriptstyle\pm1.6$	& $78.5\scriptstyle\pm0.7$ \\
	 			\quad GCNAL~\cite{gcn_2021_cvpr}	& $54.9\scriptstyle\pm1.4$	& $63.3\scriptstyle\pm2.2$	& $70.8\scriptstyle\pm1.4$	& $77.0\scriptstyle\pm1.3$ \\
	 			\quad CDAL~\cite{cdal_2020}	    & $60.9\scriptstyle\pm0.1$	& $67.2\scriptstyle\pm0.4$	& $74.6\scriptstyle\pm0.2$	& $78.4\scriptstyle\pm0.5$ \\
	 			\quad BADGE~\cite{badge_iclr_2020}	    & $60.6\scriptstyle\pm1.3$	& $67.3\scriptstyle\pm0.2$	& $73.2\scriptstyle\pm1.1$	& $\boldsymbol{78.7}\scriptstyle\pm0.2$ \\
	 			\midrule
	 			\quad Ours	    & $\boldsymbol{62.5}\scriptstyle\pm0.6$	& $\boldsymbol{69.4}\scriptstyle\pm0.7$	& $\boldsymbol{75.1}\scriptstyle\pm0.3$	& $78.3\scriptstyle\pm0.1$ \\
	 			
	 			
	 			\bottomrule
	 	\end{tabular}
	 	}
	 	\vspace{-2mm}
	 	\caption{\small{Top-1 test accuracy of various AL methods on HMDB~\cite{hmdb_iccv_2011}. \newline * Values on top of each column reveal the size of the labelled set at the end of each round.
	 	}
	 	}\label{tab:vc_hmdb_results}
	 	\vspace{1mm}
	 \end{table}
	 
	\noindent \textbf{Video Classification results.} Table.~\ref{tab:vc_hmdb_results} summarises the results for applying various AL methods for the activity recognition in videos where our approach outperforms the baselines. 
    Interestingly, compared to the Random sampling, we are able to improve the Top-1 test accuracy by more than $5\%$ in the first two AL rounds and $3\%$ in the last ones. This signifies the effectiveness of our proposed approach in reducing the labelling cost which is particularly an obstacle for video classification tasks. Moreover, {\methodname} outperforms all other strong baselines with a large margin (more than $2\%$) in the first three AL rounds. Interestingly, this is similar to what we observe from our experiments on other data types and show the effectiveness of our approach when applied to pre-trained transformers and/or in low-data regimes. 
    
	\subsection{Ablation Study}
	\vspace{-1mm}
	
	\begin{figure*}
        \centering
        \begin{subfigure}{0.31\textwidth}
            \centering
     		\includegraphics[width=1\textwidth]{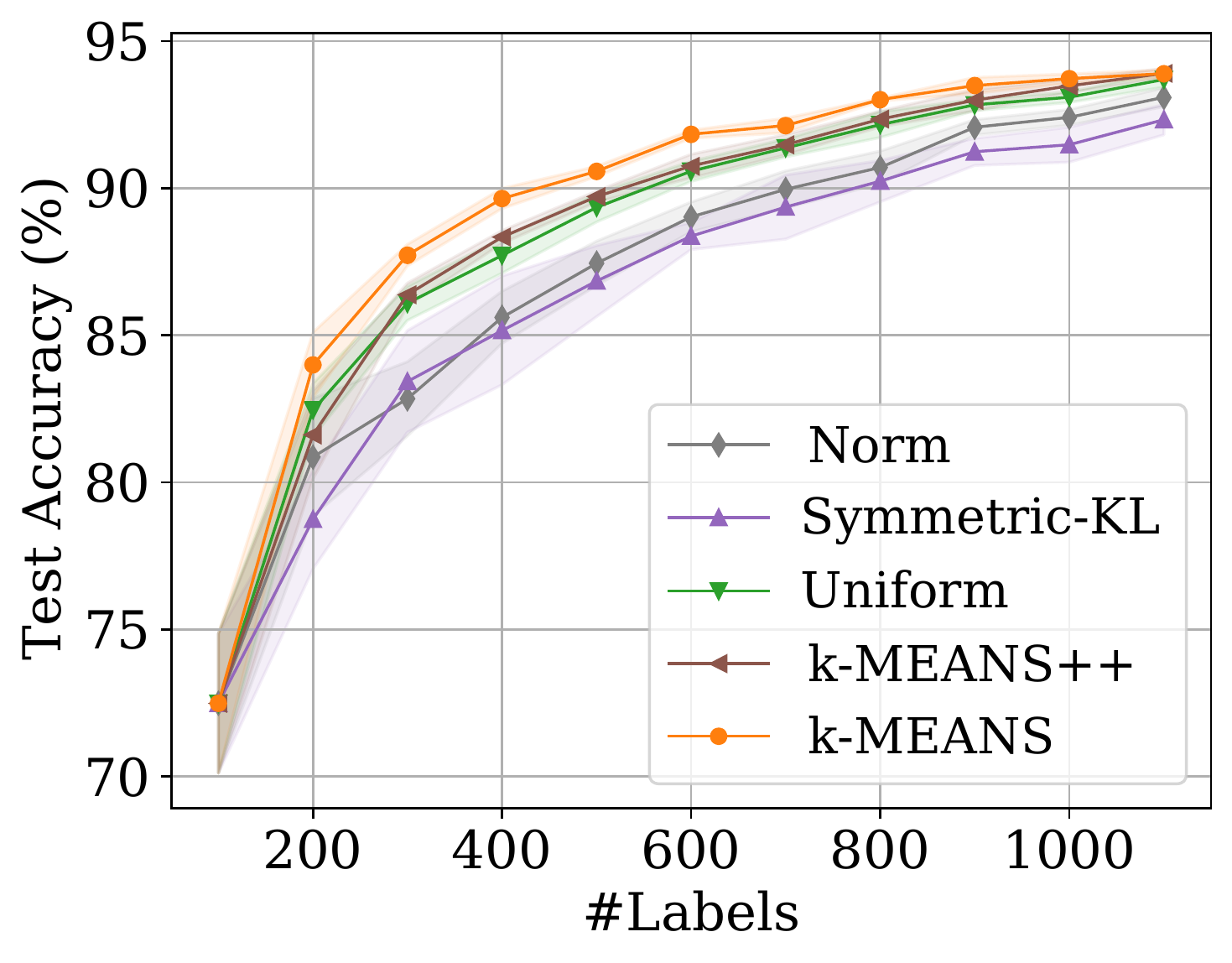}
    	 	\vspace{-5mm}
    	 	\caption{\small{Diversity impact of the sample selection from the candidate set ($\Uncertain$). \newline*$k$-MEANS is our proposed full model.}}
	 	    \label{fig:ablation_sampling}
        \end{subfigure}
		\begin{subfigure}{0.33\textwidth}
		    \centering
            \includegraphics[width=1.0\textwidth]{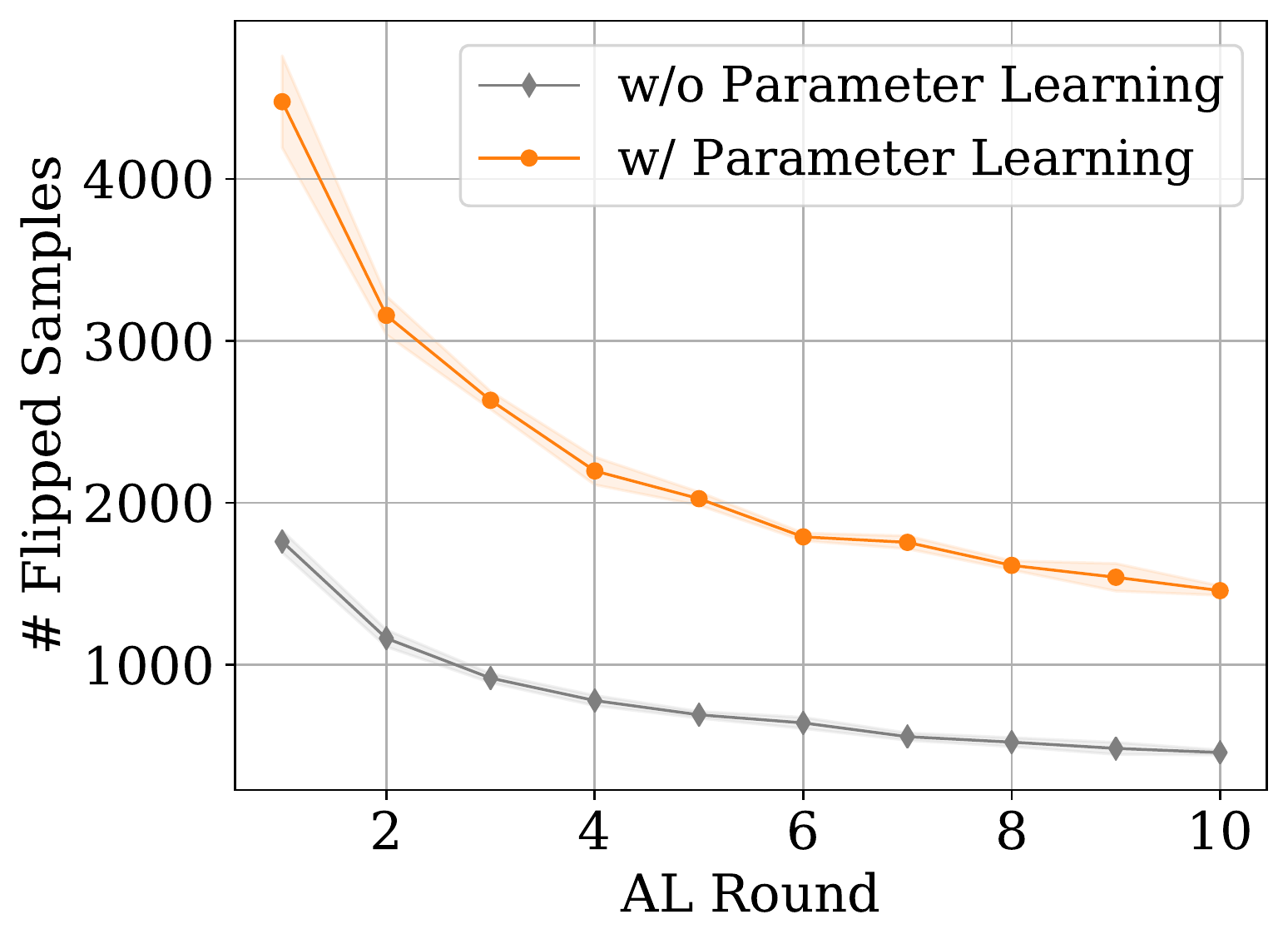}
            \caption{{\small{Number of unlabelled samples whose predictions flip with and without learning the interpolation parameter $\bnoise$.}}}
            \label{fig:ablation_learn_changes}
        \end{subfigure}
		\begin{subfigure}{0.33\textwidth}
		    \centering
    		\includegraphics[width=1.0\textwidth]{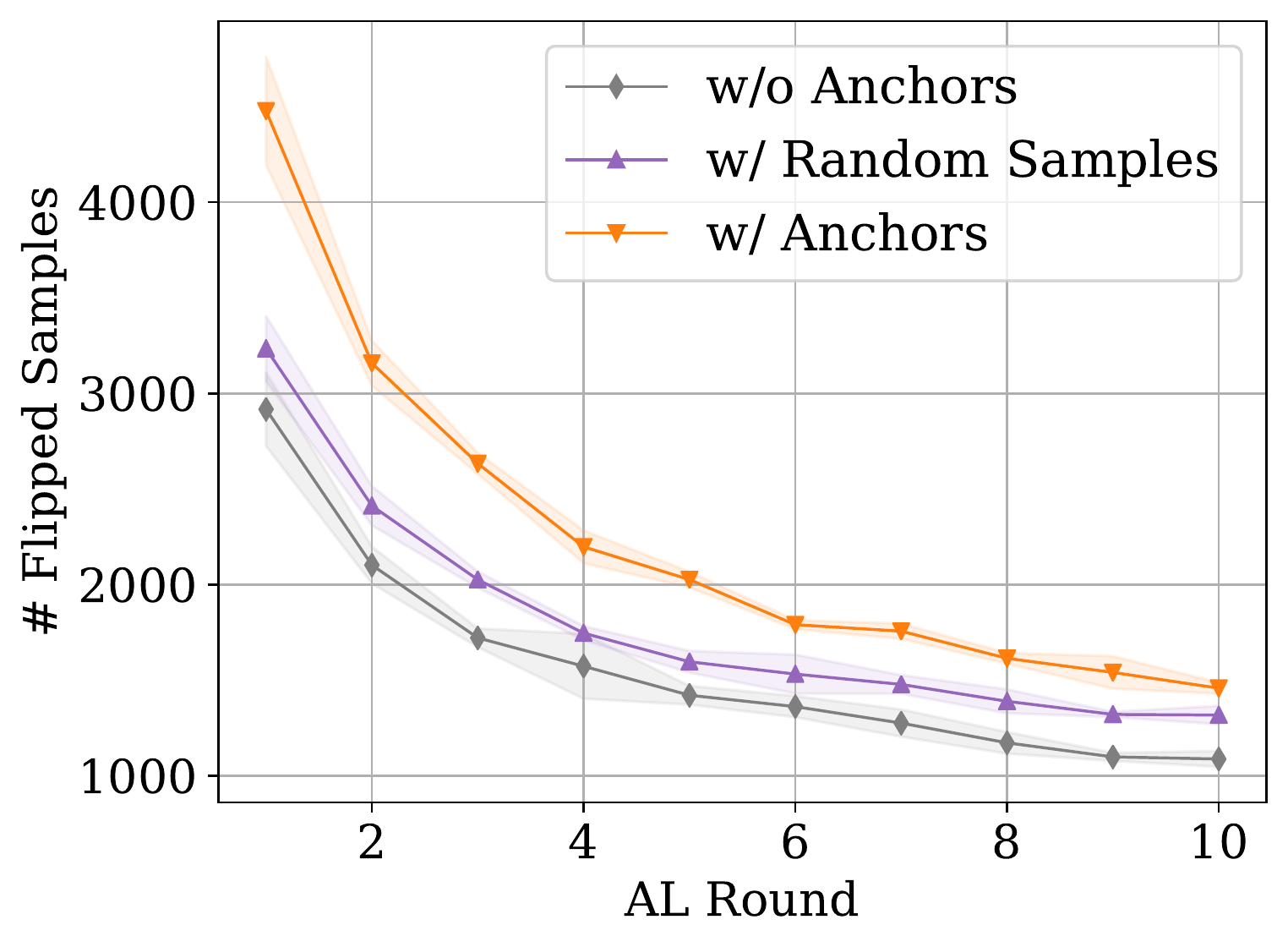}
    		\caption{\small{The impact of anchors on identifying samples whose labels flip during the interpolation.
    		}}
    		\label{fig:ablation_prototype}
		\end{subfigure}
		\vspace{-2mm}
		\caption{\small{Ablations of our AL approach. Experiments are conducted on MNIST datasets using an MLP model and a small AL budget.}}
    		\label{fig:ablation}
		\vspace{-4mm}
	\end{figure*}
	
	\textbf{Learning Ablations.} 
	Figure.~\ref{fig:ablation_chart} demonstrates the percentage of AL rounds where {\methodname} performs better than the baselines considering input data type, network architecture, network parameter initialisation and the budget size. The results indicate our approach, irrespective of other factors, consistently outperforms other AL baselines. Interestingly, when employing pre-trained networks, which is a common practice for transferring learnt representations to new tasks, {\methodname} $99\%$ of occasions assists the model to generalise better than random sampling. Additionally, in these settings, our approach surpasses the strongest baselines (CDAL and BADGE) in more than $40\%$ of the rounds. Above all, using Vision Transformer networks pre-trained in a self-supervised manner, {\methodname} not only outperforms Random, BALD, CoreSet and GCNAL in all the AL settings, it also significantly improves on BADGE and CDAL in $60\%$ and $43\%$ of the rounds respectively.
	
	Interestingly, we observe a significant advantage from our proposed AL method when it is applied on small budget setting (Figure.~\ref{fig:ablation_chart}). In fact, the test performance of our approach exceeds BADGE (the best performing baseline) in $46\%$ of the small budget experiments. 
	Moreover, we observe a more evident gap between our approach and others when it comes to AL in low-data regime. For that, we consider the performance in the first 5 rounds of AL using a small budget; \ie starting from $10\times\lblsize$ randomly selected labelled samples, each method queries for the maximum of $50\times\lblsize$ unlabelled samples overall during 5 AL iterations.
	Figure.~\ref{fig:ablation_chart} demonstrates the dominance of our approach in this setting as it eclipses all other approaches in at least $60\%$ of the experiments. 
	When using a large budget, our approach is able to slightly surpass BADGE which previously has shown great success in this setting.
	
	\noindent\textbf{Diversification.} Figure.~\ref{fig:ablation_sampling} illustrates the effectiveness of the batch diversification on selecting final instances from the set of samples whose predictions have been changed ($\Uncertain$) during the interpolation process. 
    In addition to \emph{uniformly} sampling instances from the candidate set, we consider two heuristics: (1) the \emph{norm} of the interpolation parameter $\|\bnoise\|_2$ in which a lower norm indicates with smaller intervention the model changed prediction; and, (2) the \emph{symmetric KL-Divergence} between the predicted label distributions from the unlabelled instance $p(\lbl|\ltnu; \clsparams)$ and that of the interpolated variant $p(\lbl|\intp; \clsparams$). The latter evaluates the distributions change in the output (\ie prefers samples with highest values of symmetric KL-Divergence). 
	Interestingly, both heuristics show poor performances even in comparison with the uniform selection from the candidate set. While this highlights how hard the candidate selection could be, one explanation is that these approaches might use a considerable proportion of the budget on samples that reside in a small region of the space. Consequently, the selected batch does not carry the whole information obtained through the interpolation process.
	
	In addition to the heuristic measures, we considered $k$-MEANS++, a simpler variation of $k$-MEANS that has shown better performance in \cite{badge_iclr_2020}, as another contender. In contrast to what found in \cite{badge_iclr_2020}, in our experiments, $k$-MEANS outperforms $k$-MEANS++ considerably as it better representations found using interpolation. 
	
	\setlength{\intextsep}{2mm}%
	\setlength{\columnsep}{3mm}%
    \noindent \begin{wraptable}{o}{0.53\columnwidth}
 		\centering\vspace{-2mm}
    		\setlength{\tabcolsep}{0.5pt}
    		\resizebox{1.\linewidth}{!}{
    			\noindent\begin{tabular}
    			 {p{1.7cm}>{\centering\arraybackslash}p{1.7cm}>{\centering\arraybackslash}p{1.7cm}}
    				\toprule
    				& \multicolumn{2}{c}{\textbf{Time} (seconds)} \tabularnewline
    				\cmidrule(lr){2-3} 
    				\textbf{Method}&\textbf{MNIST} (MLP)&\textbf{SVHN} (DenseNet)\\
    				\midrule
    				Entropy~\cite{entropy_2014}	& $1\scriptstyle\pm0$	& $169\scriptstyle\pm44$\\
    				BALD~\cite{bald_2017}	& $16\scriptstyle\pm4$	& $1723\scriptstyle\pm445$\\
    				Coreset~\cite{coreset_iclr_2018}	& $7\scriptstyle\pm2$	& $185\scriptstyle\pm49$\\
    				DFAL~\cite{adversarialDFAL_2018}	& $242\scriptstyle\pm69$ &--\\
    				GCNAL~\cite{gcn_2021_cvpr}	& $12\scriptstyle\pm4$  & $187\scriptstyle\pm65$\\
    				CDAL~\cite{cdal_2020}	& $5\scriptstyle\pm2$ & $179\scriptstyle\pm52$\\
    				BADGE~\cite{badge_iclr_2020}	& $50\scriptstyle\pm13$	& $523\scriptstyle\pm135$\\
    				\midrule
    				Ours	& $5\scriptstyle\pm7$	& $210\scriptstyle\pm50$\\					
    				\bottomrule
    		\end{tabular}}
    		\vspace{-4mm}
    		\caption{\small{Label acquisition run times of different methods. Our approach is significantly faster than BADGE and about 50x quicker than its Adversarial counterpart. 
    		}}\label{tab:ablation_time}
	 \end{wraptable}
	\noindent\textbf{Learning the Interpolation Parameter.} 
	 As it is evident in Figure.~\ref{fig:ablation_learn_changes}, skipping the learning process for the interpolation parameter $\bnoise$ (see section \ref{sec:interpolation_learn})  significantly reduces the number of samples chosen in the candidate set. This can have detrimental consequence on the diversity of samples that are selected during the clustering.

	\noindent\textbf{Anchors.} Figure.~\ref{fig:ablation_prototype} shows the impact of
	 using different anchors $\Prt$. 
    Evidently, the proposed method based on anchors outperforms other plausible alternatives including picking random samples from the labelled set and removing $\prt$ during the interpolation.
    The latter resembles adding noise to the sample and is similar to applying adversarial attack in the latent space.

	\noindent\textbf{Acquisition Time.} 
	We measured the time required to choose instances for labelling during each AL round. As demonstrated in Table~\ref{tab:ablation_time}, using a simple MLP network or a deep DenseNet-121, our approach performs competitive with the fastest baselines. 
	This is mainly because of the fact that we only back-propagates to a latent representation layer (not the whole network). 
	 Additionally, our approaches reduces the time required for BADGE (the best performing baseline) by a factor of more than $2$ when applied to datasets with a small number of classes.
	 We should note that running BADGE on large-scale datasets with numerous classes requires a considerable time and computing resources. The main reason is the large dimensionality of the gradient embedding in tasks with large number of classes and instances. More importantly, Table~\ref{tab:ablation_time} shows the time needed for DFAL method for MNIST dataset, which makes it impossible to apply to deep models and large datasets in a reasonable time.
	
	\vspace{-2mm}
	\section{Conclusions and Limitations}
	\vspace{-1mm}
	In this paper, we proposed a simple AL method based on the interpolation between labelled and unlabelled samples. 
	We effectively applied {\methodname} to a wide variety of image, non-image and video datasets and demonstrate its state-of-the-art results across various settings. Attractively, when the labelled set is small and the budget is limited, our approach is able to gain the most performance boost--it surpassed all other baselines in around 60\% of all evaluated rounds.
	
	Further, the feature representations are not generally disentangled \cite{locatello20a,genesis_iclr_2020} and interpolation in the high dimensional space may yield representations for unexpected inputs. Nevertheless, our approach indicates such interpolations highlight reasonable variations in the input that may otherwise remain unexplored. For future, we consider using disentangled representations to explore novel factors of variations.

	\noindent\textbf{Limitations}:
	AL consciously selects a small subset of a large pool of unlabelled samples to be labelled and used to train a model. 
    AL will be essential in catastrophes, like pandemics, where the time to reach a model at a particular level of accuracy becomes vital and would directly impact the lives of people. In spite of that, its a common practice to evaluate AL in a simulated environment mainly due to financial constraints. However, AL community at large and our approach in particular could heavily benefit from real-world evaluations.

    {\small
    \bibliographystyle{ieee_fullname}
    \bibliography{main}
    }
	
	\clearpage
	\newpage
	
	\setcounter{section}{0}

    {
    \centering
	\section*{\huge \textit{Supplements}}
	}

    \section{Methodology}
    
    \noindent\textbf{Details of Eq. (2) in the main text.}
    We can write the first-order Taylor expansion of the loss for an interpolation w.r.t. $\ltnu$ as:
    \begin{align}
	\loss\left(f_c\left(\intp\right), \lblmax\right)\approx\,&\loss\left(f_c(\ltnu), \lblmax\right)+ \label{eq:taylor_1}\\  
	&\left(\intp-\ltnu\right)^{\intercal} . \nabla_{\ltnu} \loss\left(f_c\left(\ltnu\right), \lblmax\right)\,. \notag
	\end{align}
	We also know that considering $\intp=\bnoise\prt+(1-\bnoise)\ltnu$, we will have
	\begin{align}
	\intp-\ltnu&=\left(\bnoise\prt+(1-\bnoise)\ltnu\right)-\ltnu \notag\\
	    & =\bnoise\prt+\ltnu-\bnoise\ltnu-\ltnu \notag\\
        & =\bnoise\prt-\bnoise\ltnu \notag\\
	    & =\bnoise(\prt-\ltnu)\,. \label{eq:difference}
	\end{align}
	By replacing this in Eq.~(\ref{eq:taylor_1}), we have
	\begin{align}
	\loss\left(f_c\left(\intp\right), \lblmax\right)\approx\,&\loss\left(f_c(\ltnu), \lblmax\right)+ \label{eq:intp_taylor_expansion}\\  
	&\left(\bnoise\left(\prt-\ltnu\right)\right)^{\intercal} . \nabla_{\ltnu} \loss\left(f_c\left(\ltnu\right), \lblmax\right)\,. \notag
	\end{align}
	which uncovers Eq. (2) in the main text.
	
	\vspace{2mm}
	\noindent\textbf{Details of Eq. (5) in the main text.}
	As stated in section 3.3 of the main text, using a 2-norm constraint on $\bnoise$, we approximate the optimum interpolation ratio as
	\begin{align}
	\bnoiseOpt=&\argmax_{\|\bnoise\|_2\leq\noiseOptMax} \left(\bnoise(\prt-\ltnu)\right)^\intercal.\nabla_{\ltnu}\loss(f_c(\ltnu),\lblmax). \label{eq:argmax}
 	\end{align}
 	By multiplying both sides of the constraint in Eq.~\ref{eq:argmax} by $\|(\prt-\ltnu)\|_2$, we have
 	\begin{align}
	\|\bnoise\|_2\,\|(\prt-\ltnu)\|_2\leq\noiseOptMax\|(\prt-\ltnu)\|_2. \notag
 	\end{align}
 	Based on Cauchy-Schwartz inequality, we know that $\|\bnoise(\prt-\ltnu)\|_2\leq\|\bnoise\|_2\,\|(\prt-\ltnu)\|_2$. Thus, we can infer 
 	\begin{equation}
 	    \|\bnoise(\prt-\ltnu)\|_2\leq\noiseOptMax\|(\prt-\ltnu)\|_2=\noiseOptMax'.\notag
 	\end{equation}
 	Therefore, we can change the optimisation problem to 
 	\begin{align}
	\bnoiseOpt=&\argmax_{\|\bnoise(\prt-\ltnu)\|_2\leq\noiseOptMax'} \left(\bnoise(\prt-\ltnu)\right)^\intercal.\nabla_{\ltnu}\loss\left(f_c(\ltnu),\lblmax\right). \notag
 	\end{align}
 	We can use the dual norm \cite{convex_optimization} of the above equation to approximate the optimum value for $\boldsymbol{u}=\bnoise(\prt-\ltnu)$, which is
    \begin{align}
        \boldsymbol{u}^*=\noiseOptMax'\frac{\nabla_{\ltnu}\loss\left(f_c(\ltnu),\lblmax\right)}{\|\nabla_{\ltnu}\loss\left(f_c(\ltnu),\lblmax\right)\|_2}.
    \end{align}
    After replacing the actual values for $\boldsymbol{u}$ and $\noiseOptMax'$, we have
    \begin{equation}
        \bnoiseOpt\approx\noiseOptMax\frac{\|(\prt-\ltnu)\|_2\nabla_{\ltnu}\loss(f_c(\ltnu),\lblmax)}{\|\nabla_{\ltnu}\loss(f_c(\ltnu),\lblmax)\|_2} \oslash (\prt-\ltnu), \label{eq:opt_noise_dual}
    \end{equation}
    which reveals Eq. (5) in the main text ($\oslash$ indicates element-wise division).

    \begin{figure*}
        \centering
        \begin{subfigure}{0.55\linewidth}
    		\centering
    		\includegraphics[width=\linewidth]{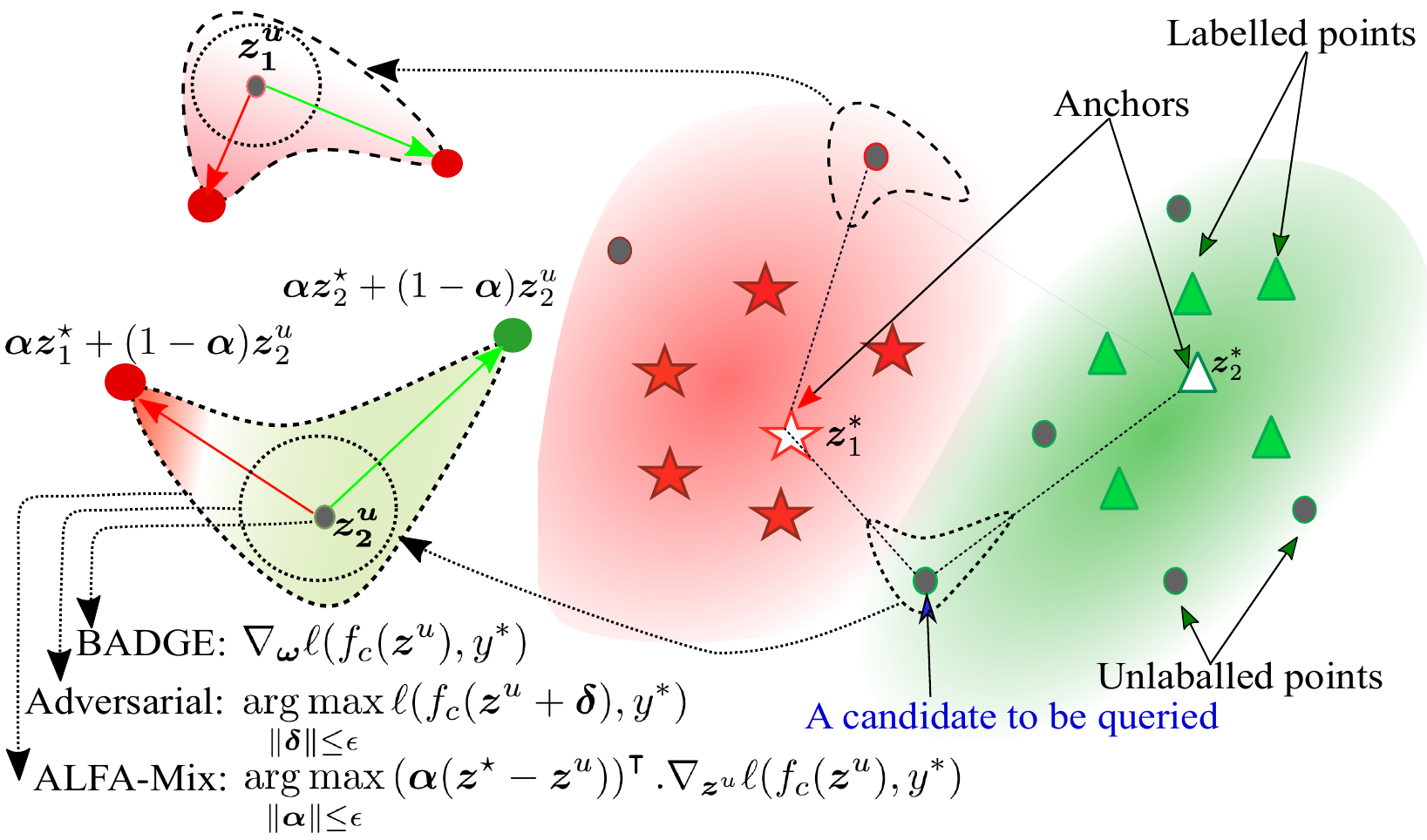}
    		\vspace{-5mm}
    		\caption{\small{Sampling strategies.}}
    		\label{fig:sub_fig1}
    	\end{subfigure}	
    	\begin{subfigure}{0.40\linewidth}
    		\centering
    		\includegraphics[width=\linewidth]{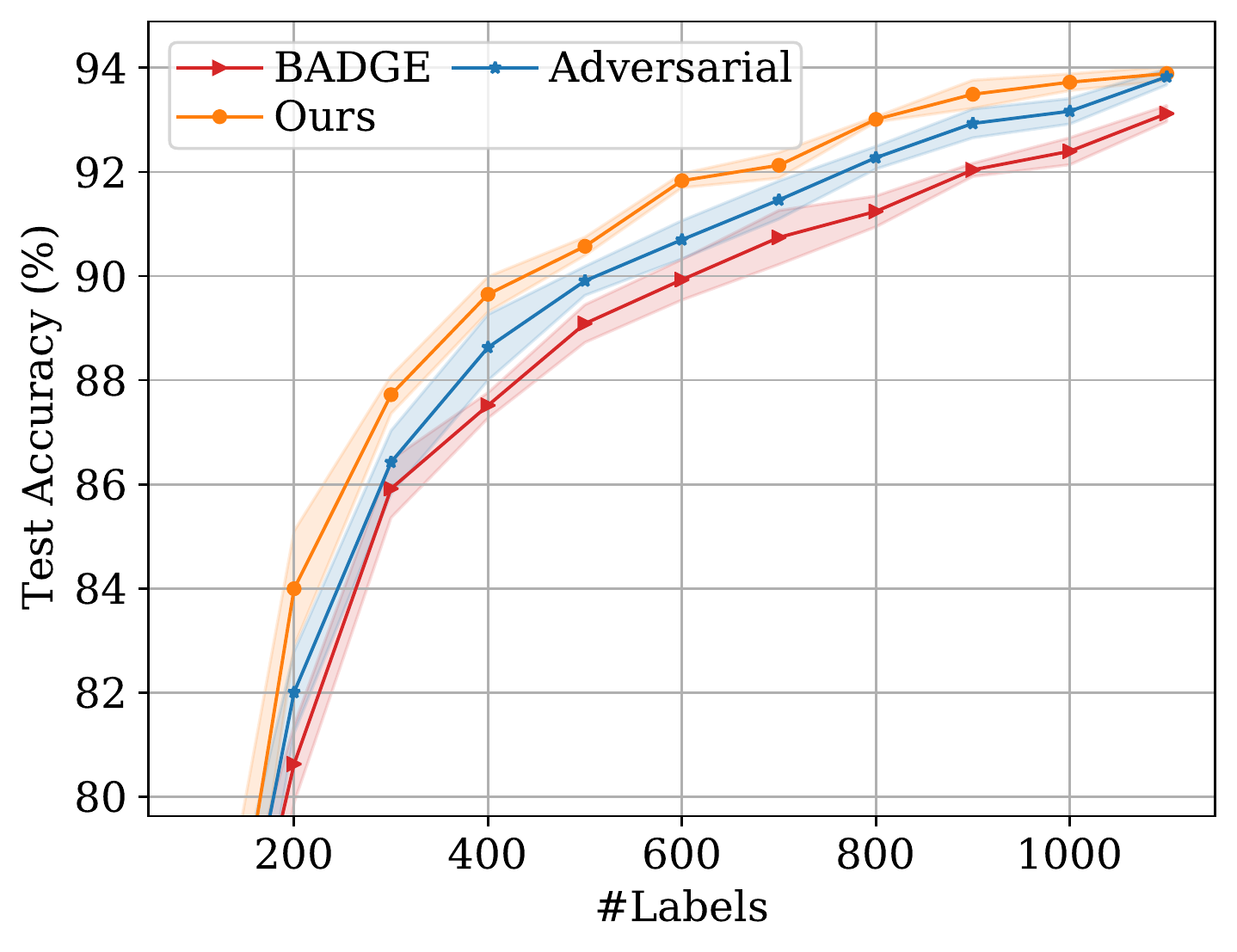}
    		\vspace{-5mm}
    		\caption{\small{Results on MNIST dataset using an MLP and a small budget of size 100 at each round.}}
    		\label{fig:adversarial}
    	\end{subfigure}	
        \vspace{-2mm}
        \caption{\small{A comparative depiction of our approach (\methodname) vs. BADGE vs. adversarial in the latent space: Since ours considers interpolations in the direction of the anchor points and proportional to their distance, it better evaluates the consistency of the predictions in the latent space. When points are less consistent, it is more intuitive to consider them as candidates to be queried (\eg $\ltnu_2$ in this figure is inconsistent after the interpolation, and hence likely to be queried).}}
        \label{fig:fig1}
        \vspace{-4mm}
    \end{figure*}

	\subsection{Relations Between {\methodname} and Other Baselines}
	\noindent\textbf{Using gradients in BADGE:} From Eq.~(3) in the main text we can understand that when the prediction is accurate and confident, small movements of the latent representation towards different directions (declared by anchors) should not change the prediction. Otherwise, as per right-hand-side of the equation, either the surface has changed dramatically or the unlabelled features is far from the labelled representations (\ie the features of the unlabelled instance are novel). This is one of the major differences of our approach when compared with BADGE that only relies on the gradients of the unlabelled instances (Figure.~\ref{fig:fig1}). 
	
	\vspace{2mm}
	\noindent{\textbf{Adversarial perturbation of features:}} To show the importance of the feature interpolations with labelled representations in our approach, we also considered using adversarial noise as an alternative perturbation mechanism. For that, we examined adding small values of noise $\boldsymbol\delta$ to the latent representations of each unlabelled point (instead of using interpolations with anchors) to find inconsistencies in their predicted labels. Following Eq. (3) and Eq. (4) in the main text, we set the objective for finding the optimum noise vector $\boldsymbol\delta^*$ as:
	\begin{align}
    	\boldsymbol\delta^*=\argmax_{\|\boldsymbol\delta\|\leq\epsilon} \loss(f_c(\ltnu + \boldsymbol\delta),\lblmax).
 	\end{align}
 	Similarly, using a first-order Taylor expansion w.r.t. $\ltnu$ and its dual norm, we can approximate the optimum noise as
 	\begin{equation}
        \boldsymbol\delta^*\approx\noiseOptMax\frac{\nabla_{\ltnu}\loss(f_c(\ltnu),\lblmax)}{\|\nabla_{\ltnu}\loss(f_c(\ltnu),\lblmax)\|_2}\,. \label{eq:opt_noise_dual_adversarial}
    \end{equation}
    After constructing a candidate set of unlabelled samples whose predicted labels are not consistent after the adversarial feature perturbation, we conduct clustering to sample a diverse set from the candidate set (similar to \methodname). Interestingly, as depicted in Figure.~\ref{fig:adversarial}, although the adversarial approach shows better performance in comparison to BADGE, it falls behind considerably when compared to {\methodname}. We believe that the main advantage of {\methodname} is the consideration of both the novelty of the features and the extent of gradient at each unlabelled point. It is worth mentioning that {\methodname} is able to identify more inconsistencies all over the decision boundary (Figure. (6c) in the main text).

    \begin{table*}[t]
	 	\setlength{\tabcolsep}{2.pt}
	 	\centering
	 	\resizebox{\linewidth}{!}{
	 		\noindent
	 		\begin{tabular}
	 			{lccccccc}
	 			\toprule
	 			\textbf{Dataset}&\textbf{Pool Size}&\textbf{Label Size}&\textbf{Input}&\textbf{Initial Instances}&\textbf{Budgets}&\textbf{Architectures}&\textbf{Initialisations}\\
	 			\midrule
	 			MNIST~\cite{mnist_1998}           & $50,000$	& $10$	& $28\times28$	& $100$	    & $100$, $1000$  & MLP, LeNet-5    & Random, Continue** \\
	 			EMNIST~\cite{emnist_2017}	        & $124,800$	& $26$	& $28\times28$	& $260$  	& $260$, $2650$  & MLP, LeNet-5    & Random, Continue \\
	 			SVHN~\cite{svhn_2011}	            & $50,000$	& $10$	& $32\times32$	& $100$	    & $100$, $1000$  & ResNet-18, DenseNet-121    & Random \\
	 			CIFAR10~\cite{cifar10_2009}	    & $50,000$	& $10$	& $32\times32$	& $100$	    & $100$, $1000$  & ResNet-18, DenseNet-121    & Random \\
	 			DomainNet-Real-10*	                & $4,673$	& $10$	& $224\times224$	& $100$	& $100$  & ResNet-18, DenseNet-121    & Pre-trained \\
	 			DomainNet-Real-20*	                & $8,615$	& $20$	& $224\times224$	& $200$	& $200$  & ResNet-18, DenseNet-121    & Pre-trained \\
	 			CIFAR100~\cite{cifar10_2009}	                    & $50,000$	& $100$	& $32\times32$	& $1000$ 	    & $1000$ & ViT-Small    & Pre-trained \\
	 			Mini-ImageNet~\cite{miniimagenet_iclr_2017}	    & $48,000$	& $100$	& $84\times84$	& $1000$	    & $1000$  & ViT-Small    & Pre-trained \\
	 			DomainNet-Real~\cite{domain_net_2019}	            & $122,563$	& $345$	& $224\times224$	& $3450$	& $3450$  &  ViT-Base, ResNet-18, DenseNet-121    & Pre-trained \\
	 			\midrule
	 			OpenML\_6	        & $18,000$	& $26$	&  $16$	& $100$	& $100$  & MLP & Random \\
	 			OpenML\_155	    & $50,000$	& $9$	&  $10$	& $100$	& $100$  & MLP & Random \\
	 			\bottomrule
	 	\end{tabular}
	 	}
	 	\vspace{-2mm}
	 	\caption{\small{A summary of diverse AL settings that we used in our image and non-image experiments. Overall, 30 different settings were utilised in our experiments to compare AL methods in various conditions.
	 	\newline* These are two small subsets of DomainNet-Real that has been used to compare AL methods on small datasets with high-resolution images.
	 	\newline**"Continue" represents the setting where the weights of the network initialise from those of the network trained in the previous round.
	 	}
	 	}\label{tab:experimental_settings}
	 	\vspace{-2mm}
	 \end{table*}
	 
	\vspace{2mm}
    \noindent\textbf{Distribution matching.}
    Denote
    $\Delta=\Expec_{p(\ltnl\mid\lblData)}\left[\ltnl\right]-\Expec_{p(\ltnu\mid\ulblData)}\left[\ltnu\right]$ if we had the distributions in the latent space. 
    We know that based on the definition of the interpolation between a pair of labelled and unlabelled samples (\ie $\intp=\bnoise\ltnl+(1-\bnoise)\ltnu$), we can have
    \begin{align}
        \ltnu=\frac{1}{1-\bnoise}\left(\intp-\bnoise\ltnl\right). \notag
    \end{align}
    By taking the expectation from both side of the above equation for all the labelled samples we have
    \begin{align}
        \ltnu=\Expec_{p(\ltnl\mid\lblData)}\left[\frac{1}{1-\bnoise}\left(\intp-\bnoise\ltnl\right)\right]. \notag
    \end{align}
    After replacing this in the definition of $\Delta$, it is easy to show that:
    \begin{equation}
    \Delta=\frac{1}{(1-\bnoise)}\left(
    \Expec_{p(\ltnl\mid\lblData)}\left[\ltnl\right]-
    \Expec_{p(\ltnu\mid\ulblData)}\left[\Expec_{p(\ltnl\mid\lblData)}\left[\intp\right]\right]\right). \notag
    \end{equation}
    That is, the interpolation operation we used here only affects difference of the expectation of distributions with a constant factor. When seen in light of Eq. (1) in the main text, it acts as a simple surrogate for a divergence measure. In fact, this relates our approach to other AL methods that their focus is on finding the distributional difference between labelled and unlabelled samples \cite{vaal_2019,gcn_2021_cvpr}. 
    
    \vspace{2mm}
    \noindent\textbf{Gradient-based interpolation optimisation.} Following \cite{counterfactual_2020,counterfactual_vln_2020}, we could have utilised iterative gradient-based optimisation  to find the optimum interpolation ratios (instead of the closed-form solution used in {\methodname}). For that, motivated by the condition in the Eq. (6) in the main text where we are interested in instances whose predictions flip with an interpolation in the latent space, we can choose $\bnoise$ as a solution to the following:
    \begin{align}
	\bnoiseOpt =&\argmax_{\bnoise \,\in\, [0,\noiseMax]^D} 
	\loss(f_c(\bnoise\prt+(1-\bnoise)\ltnu),\lblmax), \label{eq:opt_noise}\\
    \text{s.t.}& \quad\lblmax=\argmax_{k\in\{1,\ldots,\lblsize\}} f^k_c(\ltnu),\quad
    \forall\ltnu\in\Ltnu
    ,\quad \prt\in\Prt
	\notag,
	\end{align}
	where $\noiseMax$ is a hyper-parameter governing the feature mixing ratios.
	Intuitively, this optimisation chooses the hardest case of $\bnoise$ for each unlabelled instance and anchor. 
	We perform few iterations of projected gradient descent to optimise $\bnoise$. 
	Our empirical studies reveal similar performances when using this objective in comparison to the closed-form one. However, the time required for the iterative gradient-based approach is much more than the closed-form one (\ie when using 5 iterations of gradient update, it is 5x slower than {\methodname}).

	\begin{table}
	 	\setlength{\tabcolsep}{1.pt}
	 	\centering
	 	\resizebox{\linewidth}{!}{
	 		\noindent
	 		\begin{tabular}
	 			{llc|ccccccc}
	 			\toprule
	 			\textbf{Factor}&\textbf{Variety}&\rotatebox{90}{\textbf{\#Settings}}&\rotatebox{90}{\textbf{Random}}&\rotatebox{90}{\textbf{Entropy}}&\rotatebox{90}{\textbf{BALD}}&\rotatebox{90}{\textbf{CoreSet}}&\rotatebox{90}{\textbf{GCNAL}}&\rotatebox{90}{\textbf{CDAL}}&\rotatebox{90}{\textbf{BADGE}} \\
	 			\midrule
	 			\multirow{2}{*}{Data Type}    & Image & 28 &	$74\scriptstyle\%$ &	$63\scriptstyle\%$ &	$69\scriptstyle\%$ &	$80\scriptstyle\%$ &	$65\scriptstyle\%$ &	$36\scriptstyle\%$ &	$33\scriptstyle\%$ \\
	 			    & OpenML & 2 &	$95\scriptstyle\%$ &	$100\scriptstyle\%$ &	$90\scriptstyle\%$ &	$100\scriptstyle\%$ &	$100\scriptstyle\%$ &	$80\scriptstyle\%$ &	$55\scriptstyle\%$ \\
 			    \midrule
	 			\multirow{5}{*}{Architecture}    & MLP & 8 &	$98\scriptstyle\%$ &	$93\scriptstyle\%$ &	$96\scriptstyle\%$ &	$100\scriptstyle\%$ &	$99\scriptstyle\%$ &	$73\scriptstyle\%$ &	$64\scriptstyle\%$ \\
	 			    & LeNet-5 & 5 &	$100\scriptstyle\%$ &	$70\scriptstyle\%$ &	$74\scriptstyle\%$ &	$98\scriptstyle\%$ &	$66\scriptstyle\%$ &	$34\scriptstyle\%$ &	$14\scriptstyle\%$ \\
	 			    & ResNet-18 & 7 & $61\scriptstyle\%$ &	$63\scriptstyle\%$ &	$60\scriptstyle\%$ &	$80\scriptstyle\%$ &	$51\scriptstyle\%$ &	$27\scriptstyle\%$ &	$24\scriptstyle\%$ \\
	 			    & DenseNet-121 & 7 & $67\scriptstyle\%$ &	$59\scriptstyle\%$ &	$64\scriptstyle\%$ &	$83\scriptstyle\%$ &	$56\scriptstyle\%$ &	$24\scriptstyle\%$ &	$23\scriptstyle\%$ \\
	 			    & ViT & 3 & $100\scriptstyle\%$ &	$83\scriptstyle\%$ &	$100\scriptstyle\%$ &	$100\scriptstyle\%$ &	$100\scriptstyle\%$ &	$43\scriptstyle\%$ &	$60\scriptstyle\%$ \\
 			    \midrule
	 			\multirow{3}{*}{Initialisation}    & Random & 18 &	$72\scriptstyle\%$ &	$69\scriptstyle\%$ &	$65\scriptstyle\%$ &	$89\scriptstyle\%$ &	$62\scriptstyle\%$ &	$35\scriptstyle\%$ &	$29\scriptstyle\%$ \\
	 			    & Pre-Training & 9 &	$99\scriptstyle\%$ &	$72\scriptstyle\%$ &	$97\scriptstyle\%$ &	$91\scriptstyle\%$ &	$88\scriptstyle\%$ &	$40\scriptstyle\%$ &	$43\scriptstyle\%$ \\
	 			    & Continue & 3 &	$100\scriptstyle\%$ &	$100\scriptstyle\%$ &	$90\scriptstyle\%$ &	$100\scriptstyle\%$ &	$87\scriptstyle\%$ &	$83\scriptstyle\%$ &	$57\scriptstyle\%$ \\
 			    \midrule
	 			\multirow{2}{*}{Budget}    & Small & 22 &	$86\scriptstyle\%$ &	$84\scriptstyle\%$ &	$86\scriptstyle\%$ &	$92\scriptstyle\%$ &	$82\scriptstyle\%$ &	$52\scriptstyle\%$ &	$46\scriptstyle\%$ \\
	 			    & Large & 8 &	$74\scriptstyle\%$ &	$43\scriptstyle\%$ &	$53\scriptstyle\%$ &	$88\scriptstyle\%$ &	$46\scriptstyle\%$ &	$11\scriptstyle\%$ &	$9\scriptstyle\%$ \\
 			    \midrule
	 			\multicolumn{2}{c}{\textbf{Overall}}   & 30 &	$83\scriptstyle\%$ &	$73\scriptstyle\%$ &	$77\scriptstyle\%$ &	$91\scriptstyle\%$ &	$72\scriptstyle\%$ &	$41\scriptstyle\%$ &	$36\scriptstyle\%$ \\
	 			\bottomrule
	 	\end{tabular}
	 	}
	 	\vspace{-2mm}
	 	\caption{\small{The percentage of the AL rounds in different settings where {\methodname} outperforms other baselines, considering their victory scores~\cite{badge_iclr_2020}. The chart of the same results is depicted in Figure. 3 of the main text.}}\label{tab:ablations}
	 \end{table}

	\section{Experiments}
    
	\subsection{Comparison matrix}
	We demonstrate the performance comparison between every pair of AL methods over various settings in a penalty matrix proposed in \cite{badge_iclr_2020}. Each cell of the matrix reveals the number of settings in which the method shown in the column is outperformed by the ones indicated in the row. 
	It should be noted that each setting consists of conducting $R$ rounds of AL with a specific labelling budget size $\Bgt$ and using a particular model architecture on a single dataset. Since we repeat each setting with 5 different random seeds, at each round $r$ in the setting we use $t$-score of the difference between the test performances ($d_{i,j}^r=a_i^r-a_j^r$) of each pair of AL methods $(i, j)$ over the 5 repeats:
	\begin{align}
	c_{i,j}^r &= \dfrac{\sqrt{5} \mu^r}{\sigma^r},  \\ 
	\mu^r=\dfrac{1}{5}\sum_{m=1}^5d_{i,j}^r, &\quad  \sigma^r=\sqrt{\frac{1}{5}\sum_{m=1}^5(d_{i,j}^r-\mu^r)^2}, \notag 
	\end{align}
	where $a_i^r$ and $a_j^r$ are the test performances of methods $i$ and $j$ respectively at AL round $r$. Similar to \cite{badge_iclr_2020}, we also used a threshold of 2.776 for this score to decide if method $i$ wins over method $j$. After clarifying the winner at each round of the setting, we calculate
	$C_{i,j}=\sum_{r=1}^R\mathbbm{1}_{c_{i,j}^r>2.776}/R$
	as the final victory score of AL method $i$ over method $j$ in that specific setting. Additionally, to compute the matrix over multiple settings, we simply report the element-wise sum of all the individual matrices.
	
	\begin{figure*}
    	\centering
    	\begin{subfigure}{0.34\linewidth+2mm}
    		\centering
    		\includegraphics[width=\linewidth]{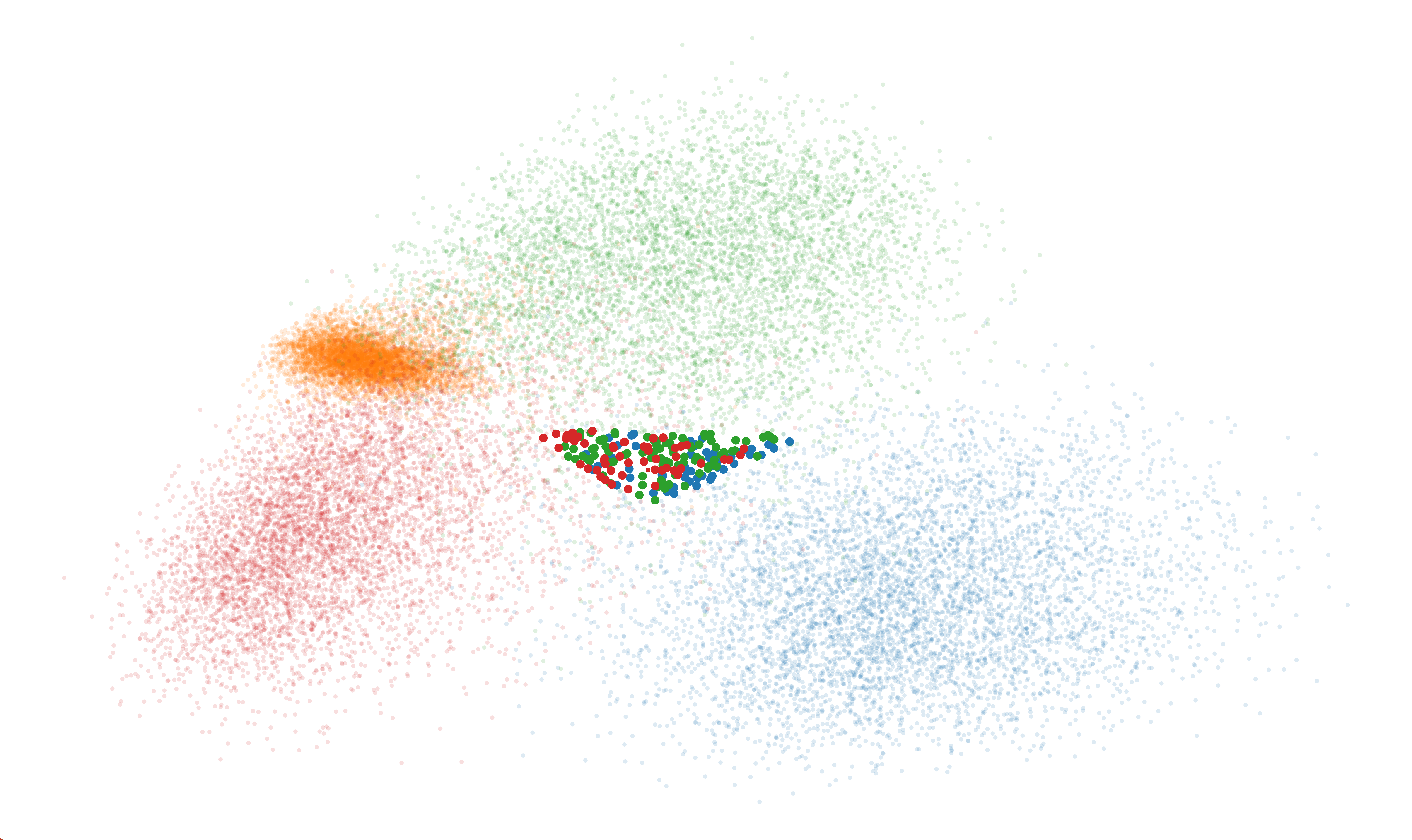}
    		\vspace{-5mm}
    		\caption{Entropy~\cite{entropy_2014}}
    		\label{fig:plot_entropy}
    	\end{subfigure}	
    	\hspace{-2mm}	
    	\begin{subfigure}{0.34\linewidth+0mm}
    		\centering
    		\includegraphics[width=\linewidth]{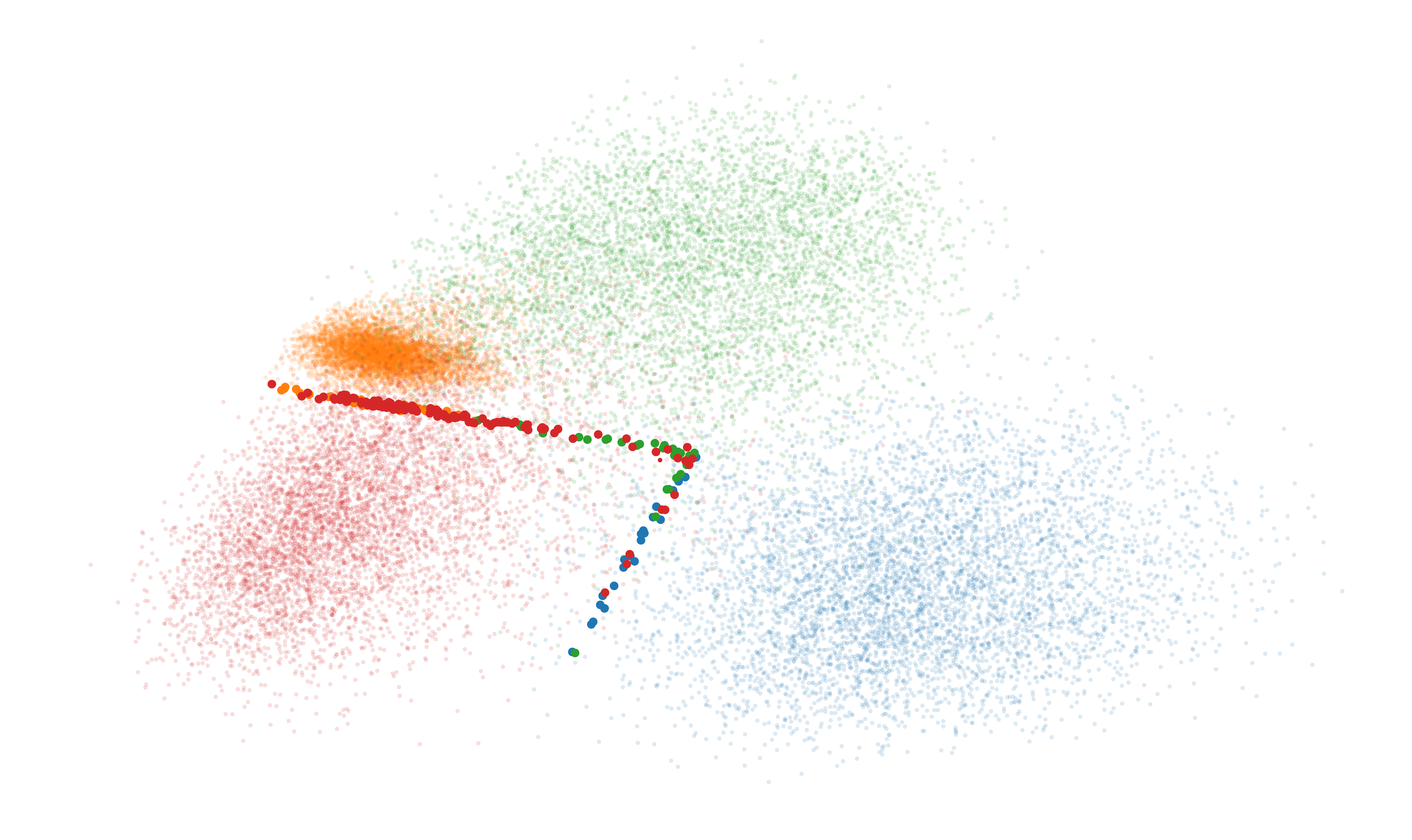}
    		\vspace{-5mm}
    		\caption{DFAL~\cite{adversarialDFAL_2018}}
    		\label{fig:plot_minimnist_dfal}
    	\end{subfigure}	
    	\vspace{-3mm}
    	\caption{\small{Visualization of sample selection behaviours of some AL methods in the latent space (other methods can be found in Figure (2) of the main text). 
    	}}
    	\label{fig:plots_minimnist}
    	\vspace{-5mm}
    \end{figure*}

	\begin{figure*}
		\begin{subfigure}{0.5\textwidth}
			\centering
			\includegraphics[width=.7\textwidth]{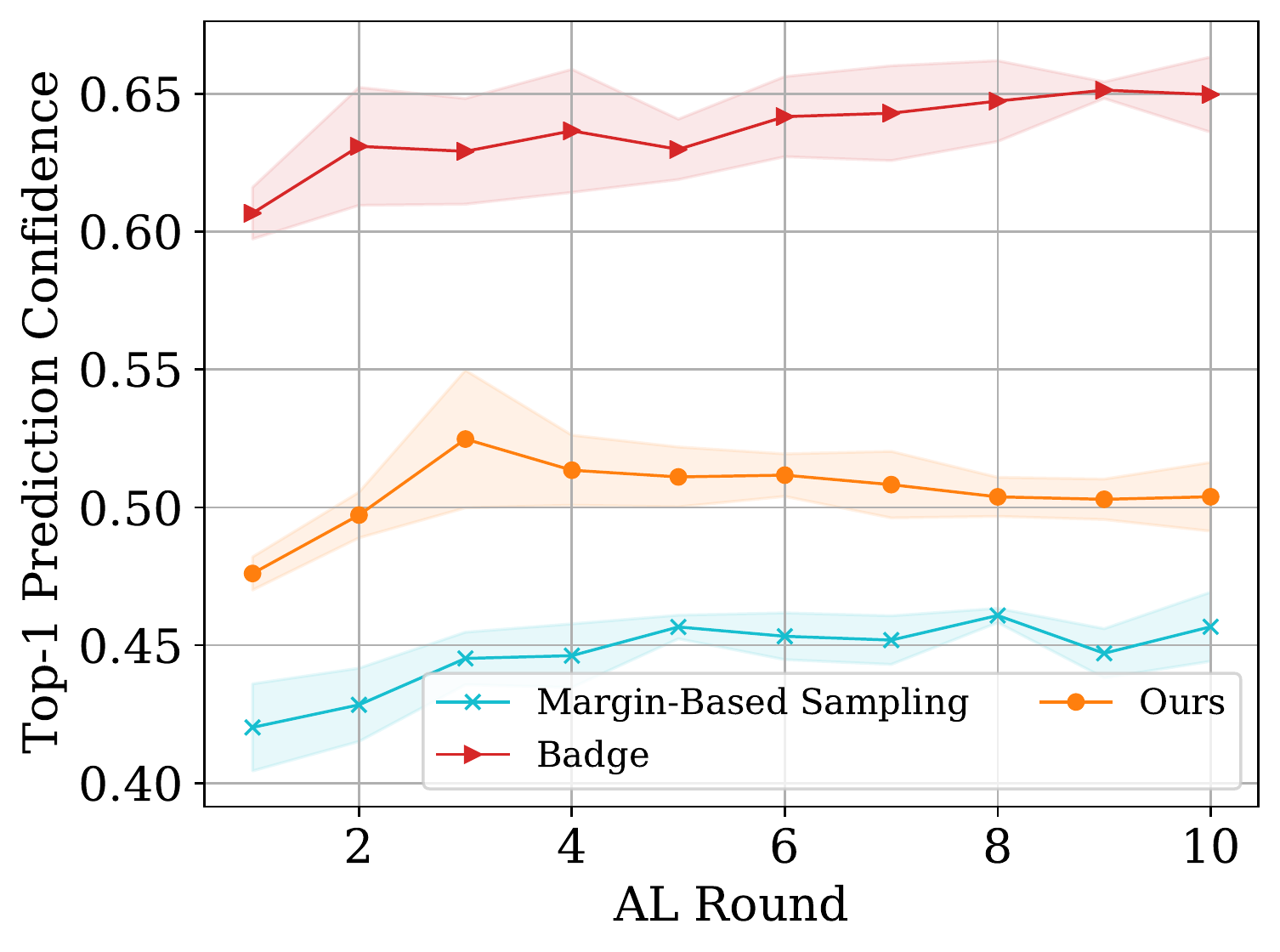}
			\vspace{-2mm}
			\caption{\small{The confidence of the predicted Top-1 class.}} \label{fig:query_conf}
		\end{subfigure}		
		\begin{subfigure}{0.5\textwidth}
			\centering
			\includegraphics[width=.7\textwidth]{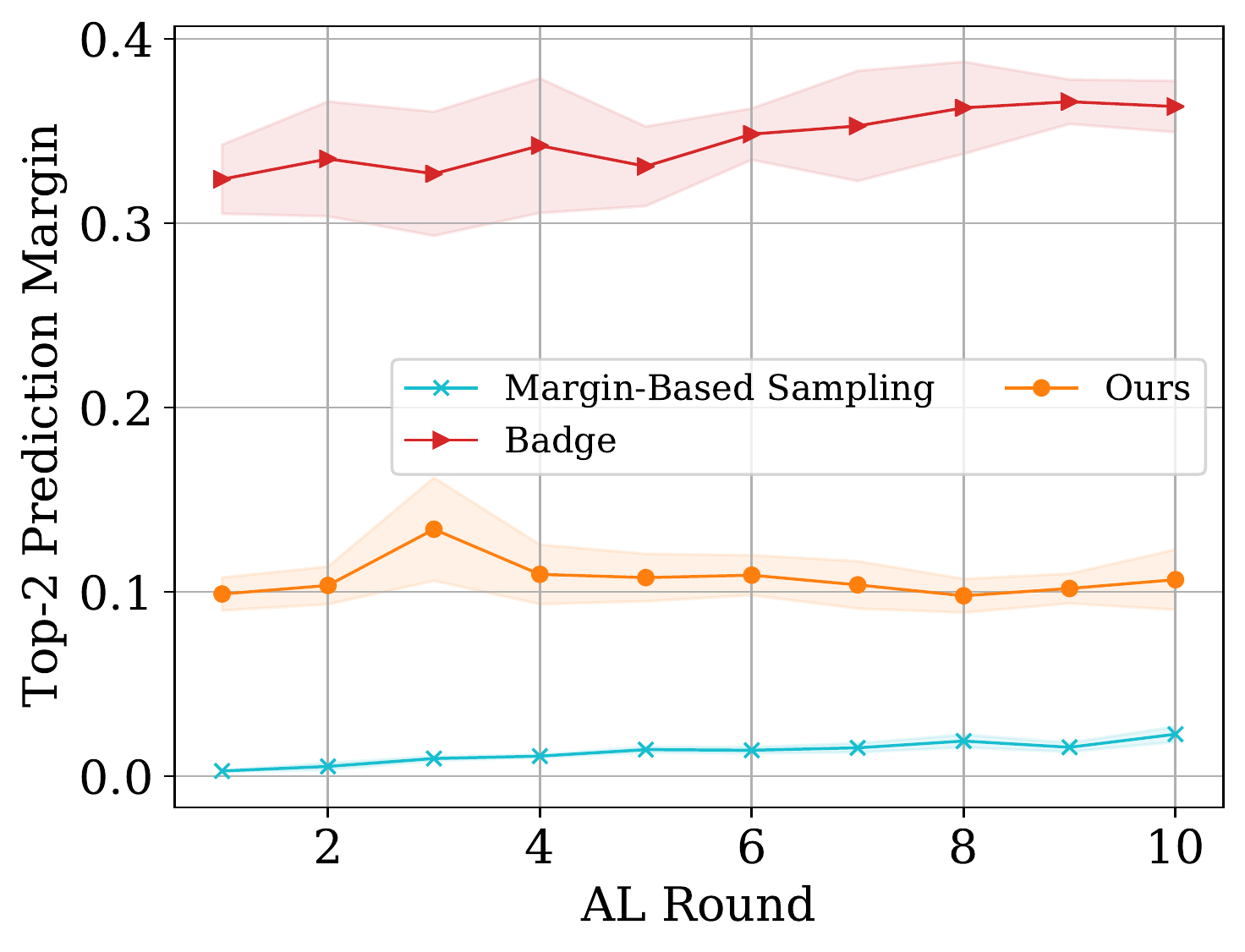}
			\vspace{-2mm}
			\caption{\small{The margin (distance) between the predicted probabilities of the Top-2 classes.}} \label{fig:query_margin}
		\end{subfigure}		
		\begin{subfigure}{0.5\textwidth}
			\centering
			\includegraphics[width=.7\textwidth]{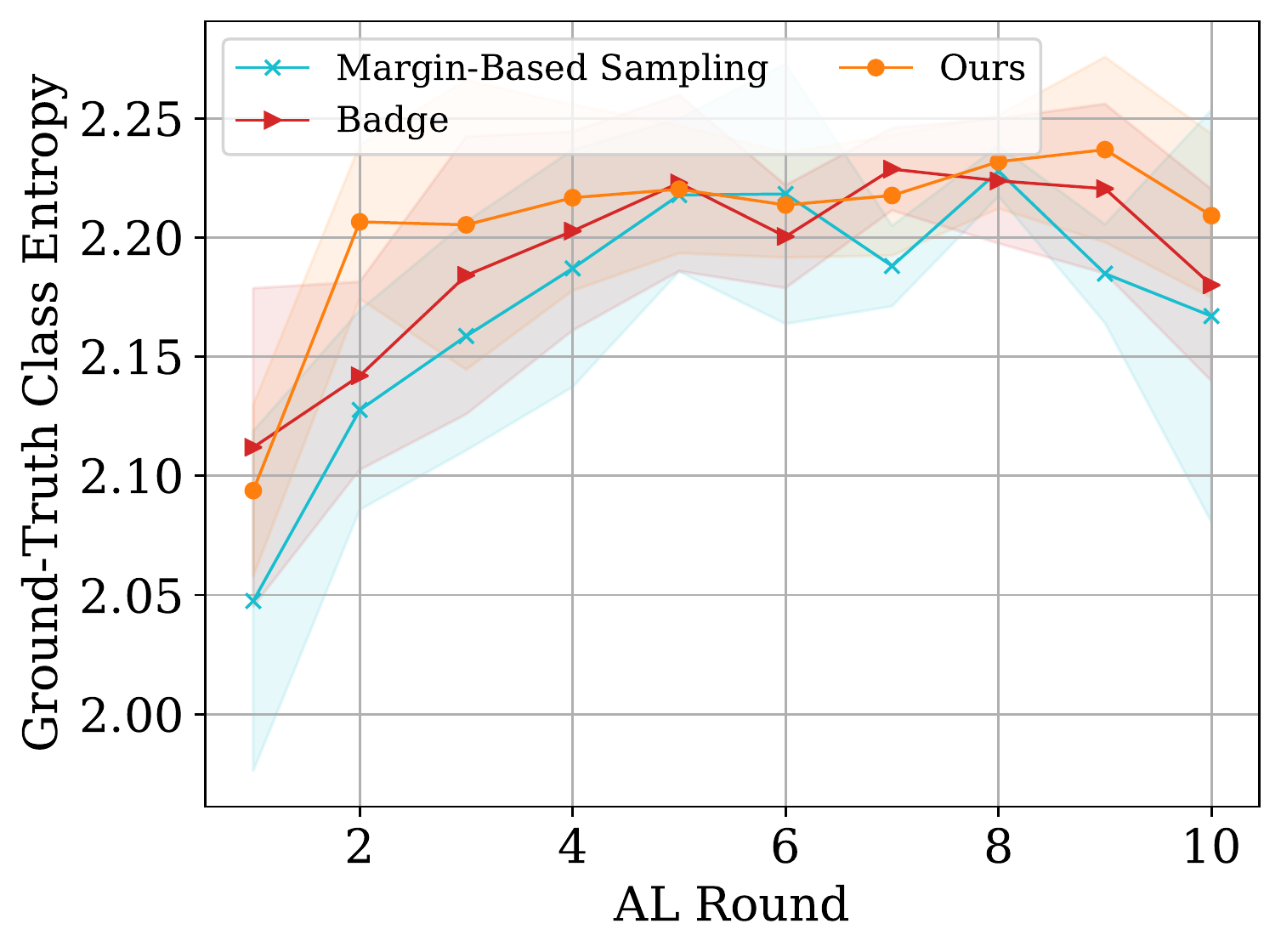}
			\vspace{-2mm}
			\caption{\small{The entropy of the revealed ground-truth labels.}}
		\end{subfigure}	
		\begin{subfigure}{0.5\textwidth}
			\centering
			\includegraphics[width=.7\textwidth]{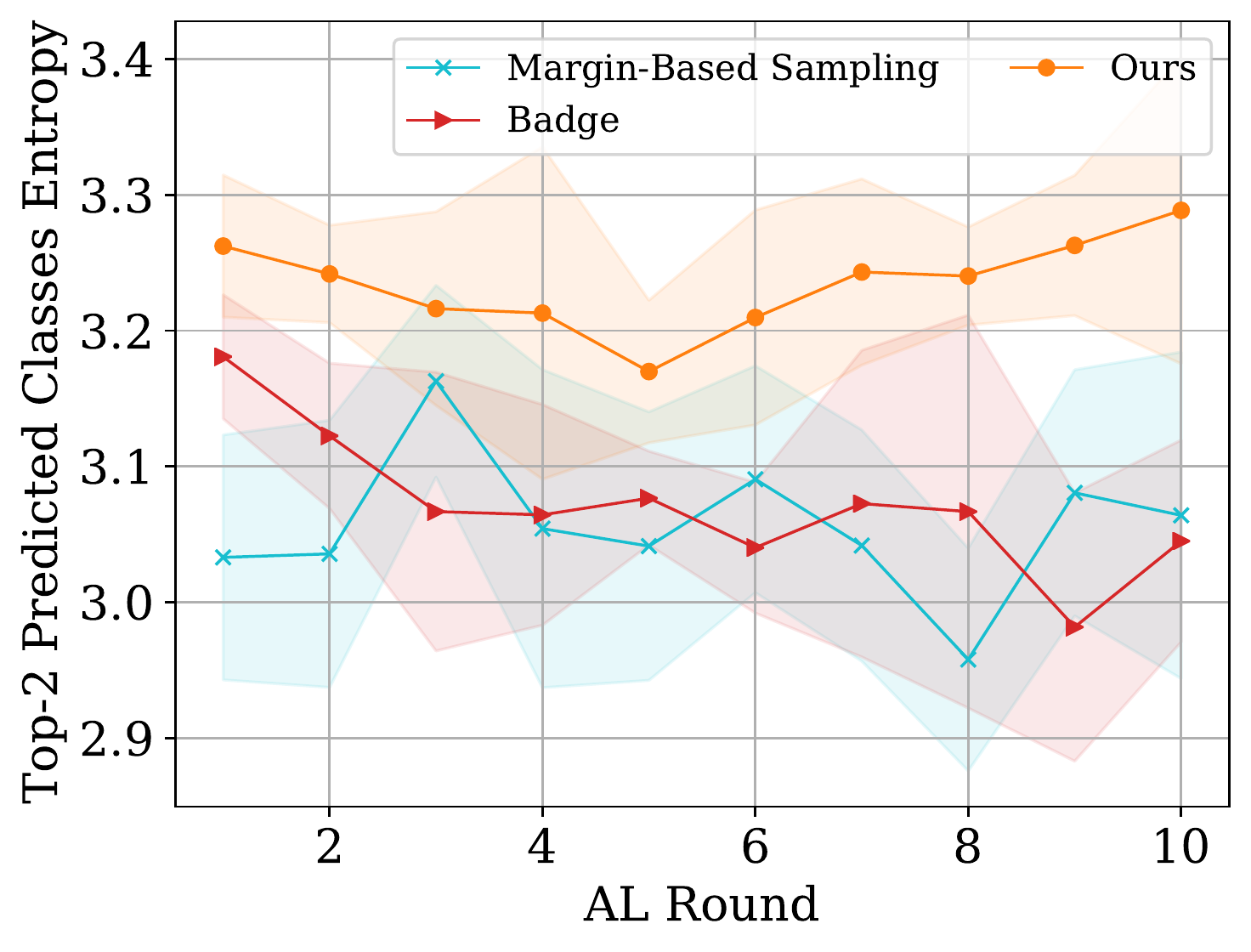}
			\vspace{-2mm}
			\caption{\small{The entropy of the predicted Top-2 classes (ignoring the order of them). }}
		\end{subfigure}
		\vspace{-4mm}
		\caption{\small{Uncertainty and diversity of the selected samples for labelling. All experiments are done on MNIST dataset using LeNet-5 model and a small budget of size 100. }} \label{fig:query_stats}
	    \vspace{-4mm}
	\end{figure*}
	
	\subsection{Sampling Diversity and Uncertainty}
	To have a better understanding with regards to the effectiveness of our approach in selecting an uncertain and diverse set of samples for labelling, we compare some characteristics of the selected batch of instances at each AL round comparing our method with those from BADGE \cite{badge_iclr_2020} and Margin-Based Sampling\footnote{Margin-Based Sampling is another AL method based on uncertainty. It selects samples with the lowest distance between the predicted probabilities for the Top-2 classes (called margin). It should be noted that BADGE has shown significantly better performance compared to Margin-Based Sampling in prior works~\cite{badge_iclr_2020}.}~\cite{margin_2006} (Figure~\ref{fig:query_stats}). 
	
\begin{figure*}
		\centering
		\includegraphics[width=0.90\linewidth]{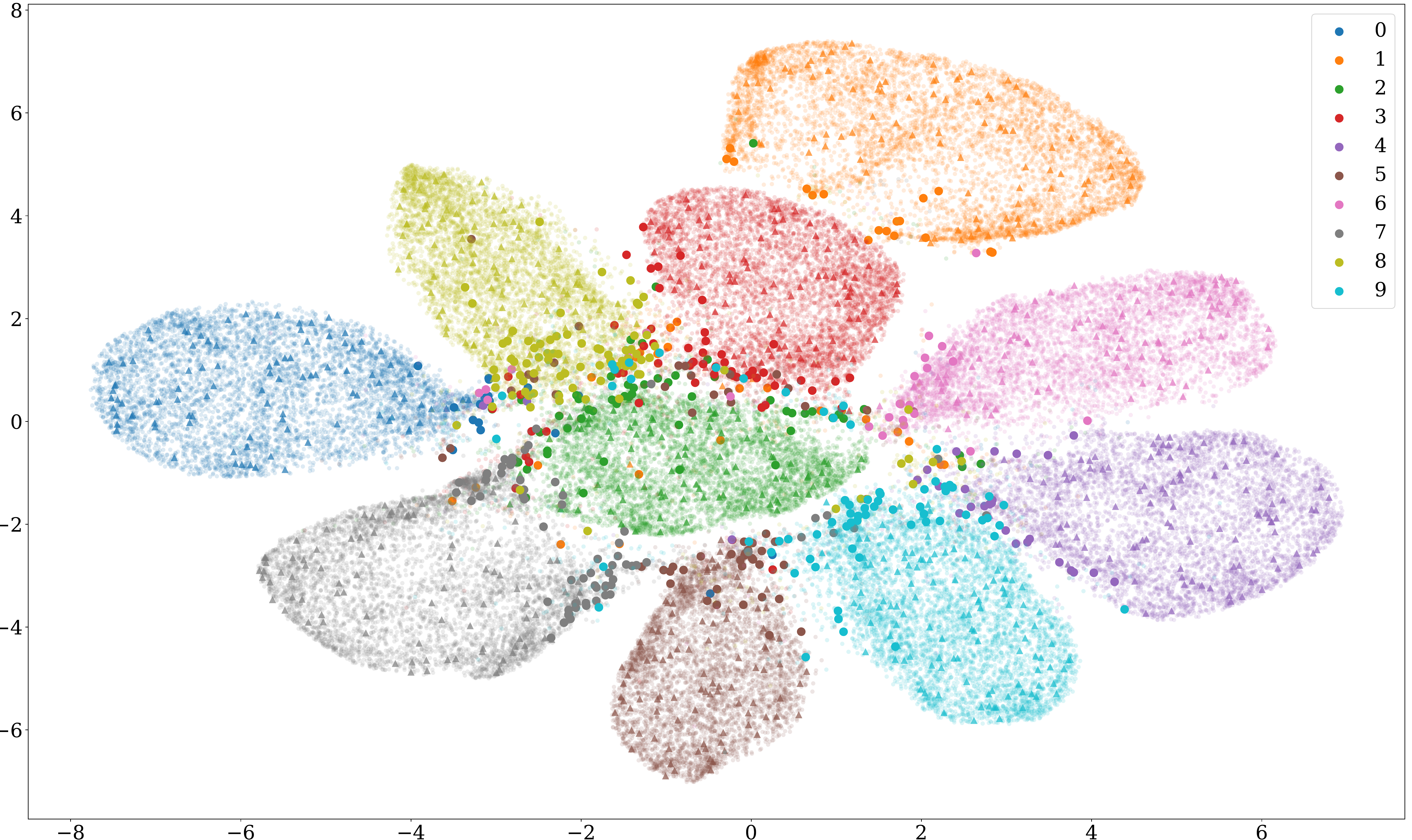}
		\vspace{-2mm}
		\caption{\small{The t-SNE visualisation of the sample selection of our proposed method on MNIST dataset using LeNet-5. The model is trained based on 500 random labelled set (shown as triangles) and is provided with a budget of size 500 to (depicted as bold circles).}}
		\label{fig:MNIST_visualisation}
		\vspace{-2mm}
	\end{figure*}

	Comparing the confidence and Top-2 prediction margins of the selected unlabelled samples, depicted in Figures \ref{fig:query_conf} and \ref{fig:query_margin} respectively, we can see that the uncertainty level of the selected samples by our method is closer to the highest possible value in comparison to BADGE sampling.
	Please note that in contrast to what Margin-Based Sampling is doing, we do not explicitly enforce our approach to select samples close to the decision boundaries. 
	On the other hand, considering the higher entropy values in the ground-truth labels of the selected set and their Top-2 predicted classes, we can realise the capability of our proposed method in selecting a diverse set of unlabelled samples in terms of their true class labels and their position with regard to the decision boundaries. All in all, as depicted in Fig.~\ref{fig:MNIST_visualisation}, our method is able to exploit both uncertainty and diversity concepts to select a diverse set of samples that lie close to decision boundaries, which leads to significantly higher performances.
	
	\begin{figure*}
		\begin{subfigure}{0.5\textwidth}
			\centering
			\includegraphics[width=.90\textwidth]{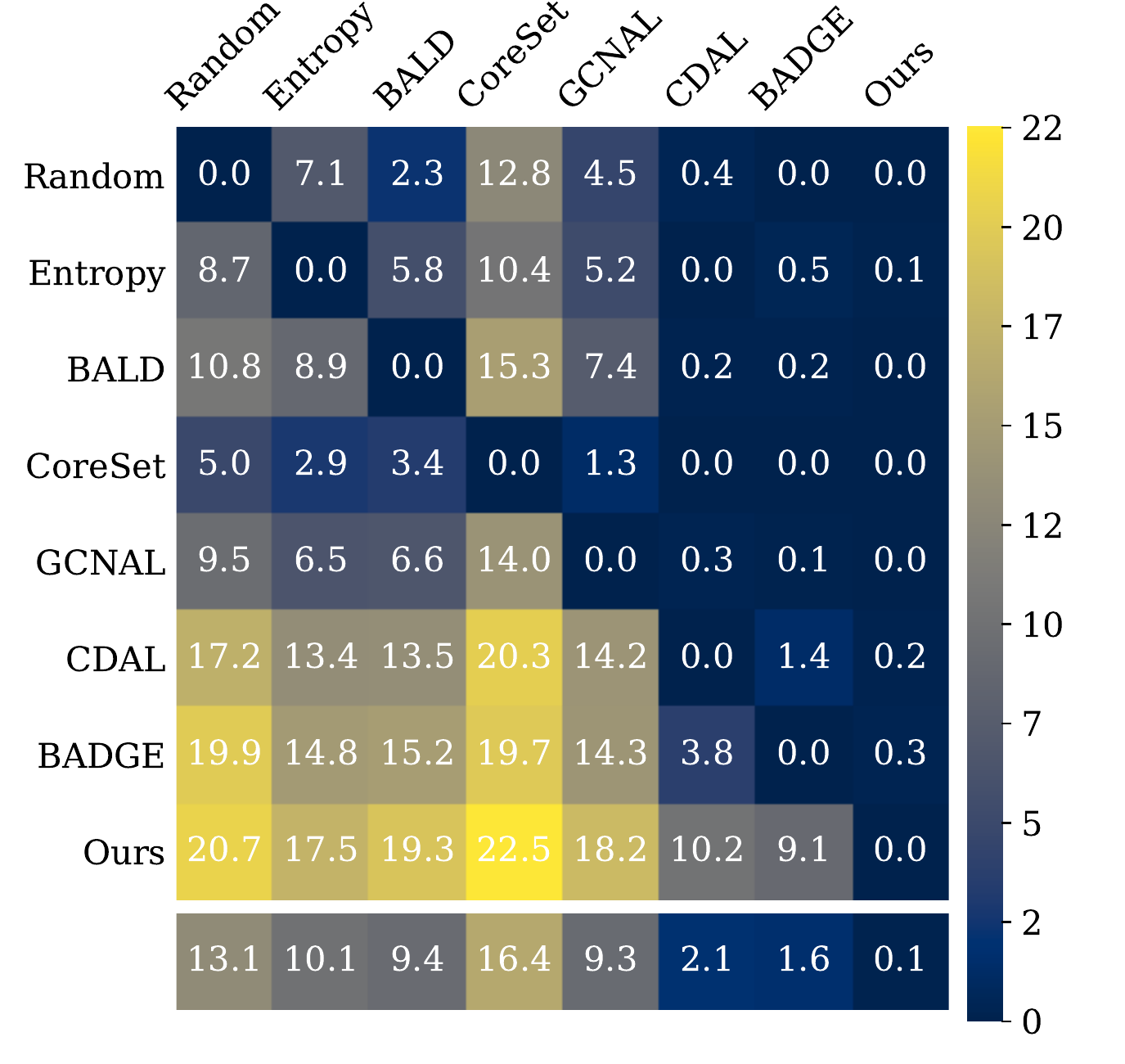}
			\vspace{-4mm}
			\caption{\small{Image (maximum value: 28)}} \label{fig:comp_random}
		\end{subfigure}	
		\begin{subfigure}{0.5\textwidth}
			\centering
			\includegraphics[width=.90\textwidth]{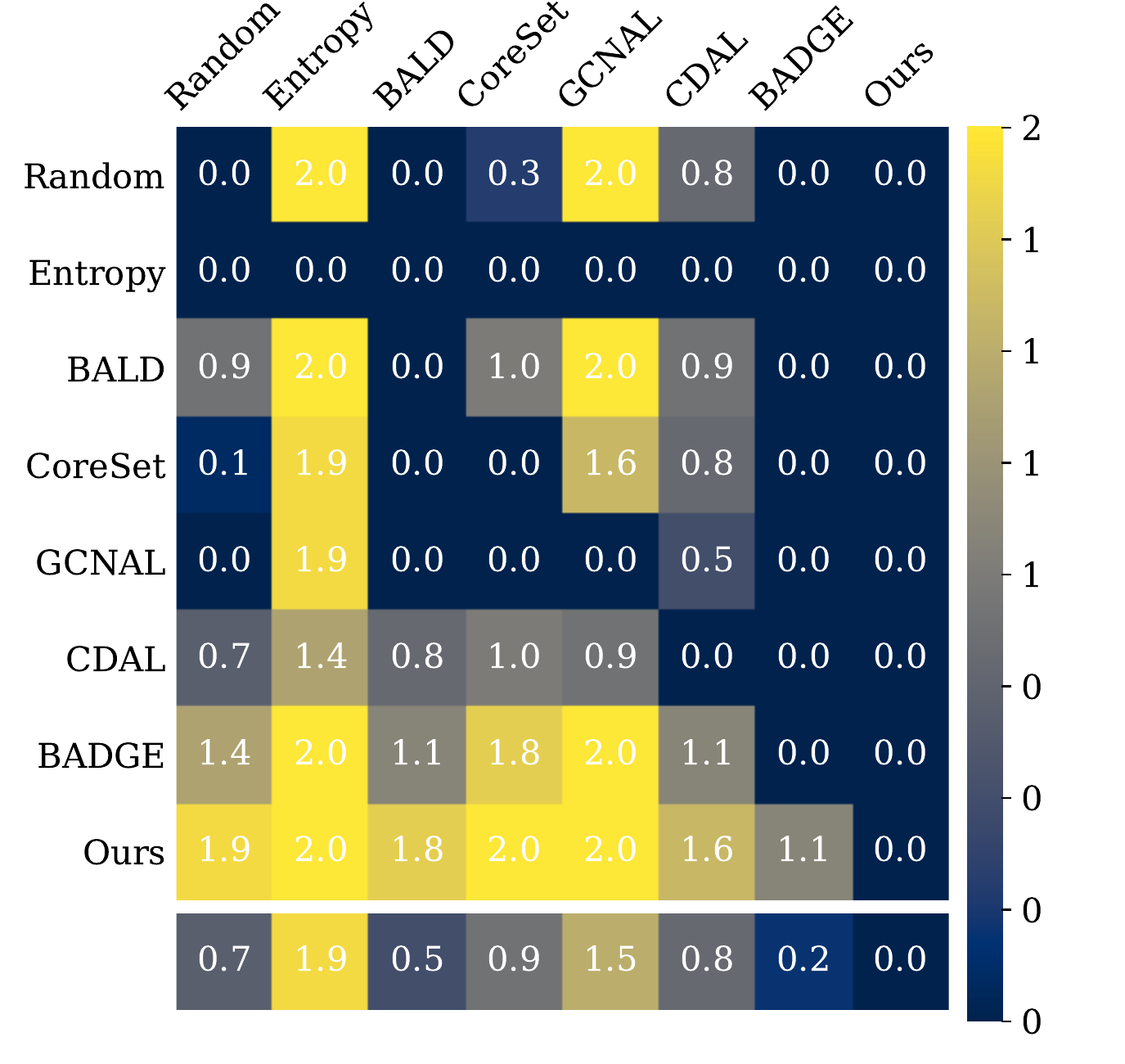}
			\vspace{-4mm}
			\caption{\small{OpenML (maximum value: 2)}} \label{fig:comp_pre_train}
		\end{subfigure}	
		\vspace{-4mm}
		\caption{\small{Pairwise comparison of different AL approaches based on the type of data. The maximum value of each cell for each setting is also provided in the captions.}} \label{fig:ablation_data_type}
	    \vspace{-6mm}
	\end{figure*}

	\begin{figure*}[t]
		\begin{subfigure}{0.5\textwidth}
			\centering
			\includegraphics[width=.90\textwidth]{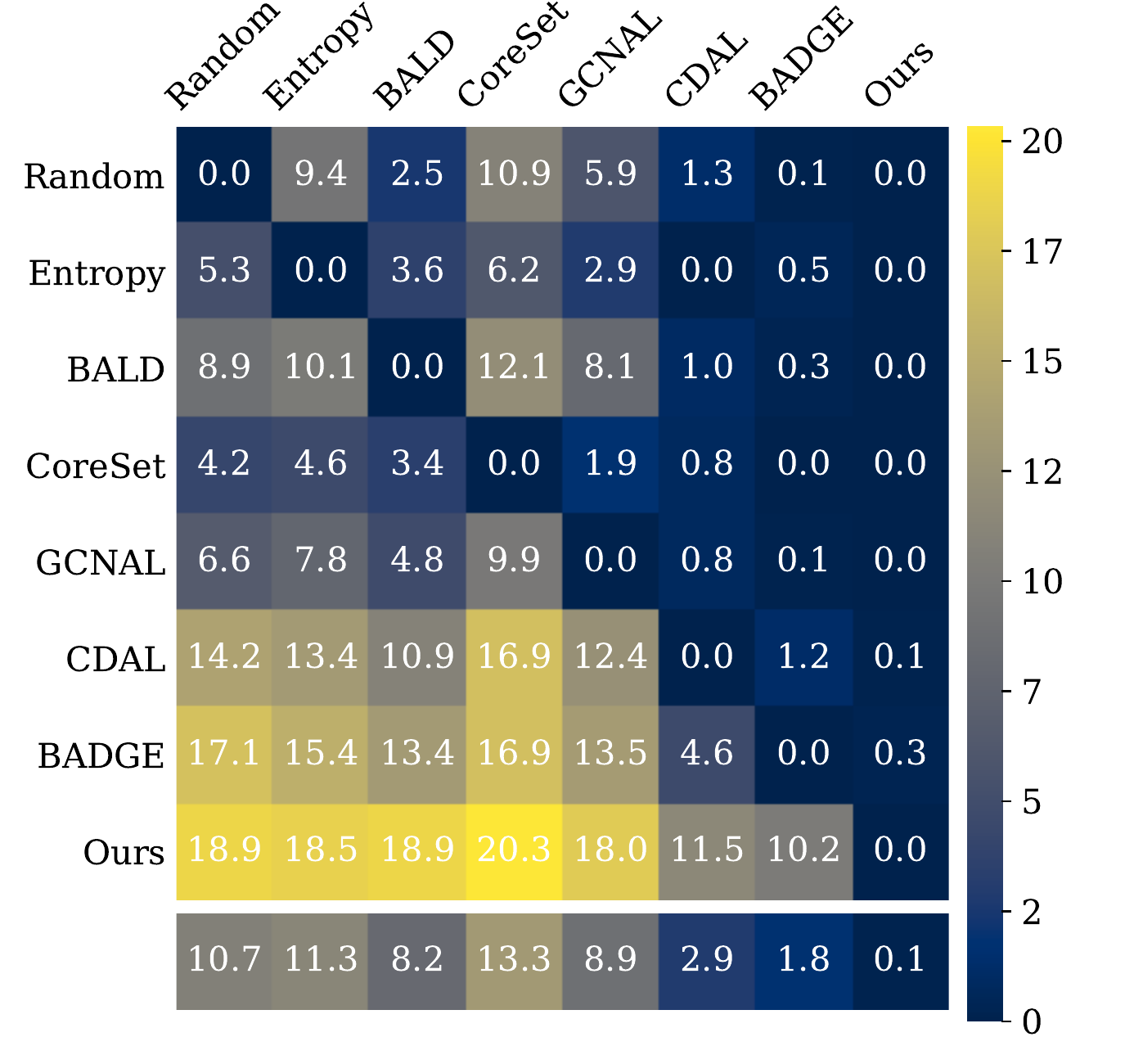}
			\vspace{-4mm}
			\caption{\small{Small budget (maximum value: 22)}} \label{fig:comp_small}
		\end{subfigure}		
		\begin{subfigure}{0.5\textwidth}
			\centering
			\includegraphics[width=.90\textwidth]{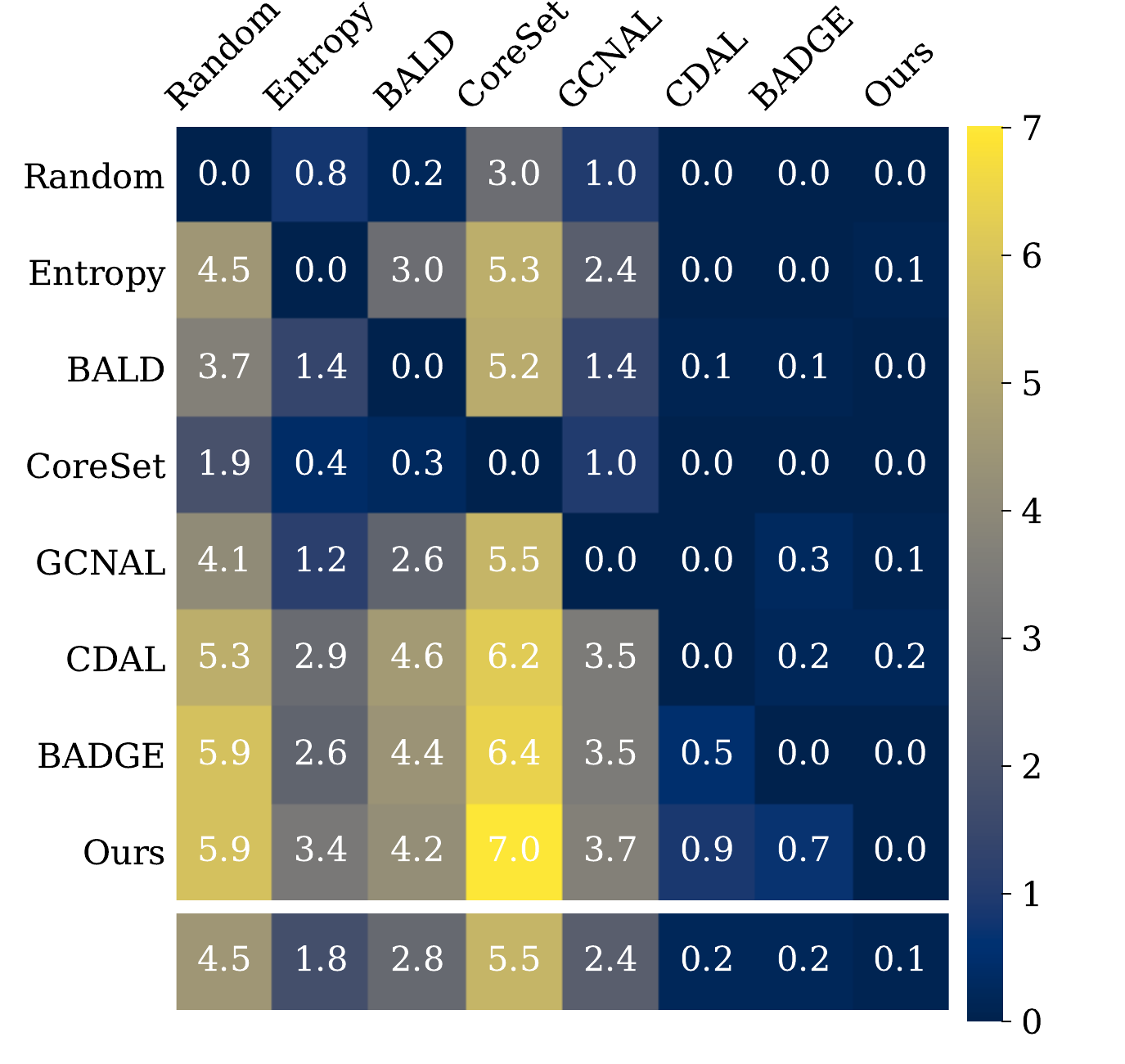}
			\vspace{-4mm}
			\caption{\small{Large budge (maximum value: 8)}} \label{fig:comp_large}
		\end{subfigure}
		\begin{subfigure}{0.5\textwidth}
			\centering
			\includegraphics[width=.90\textwidth]{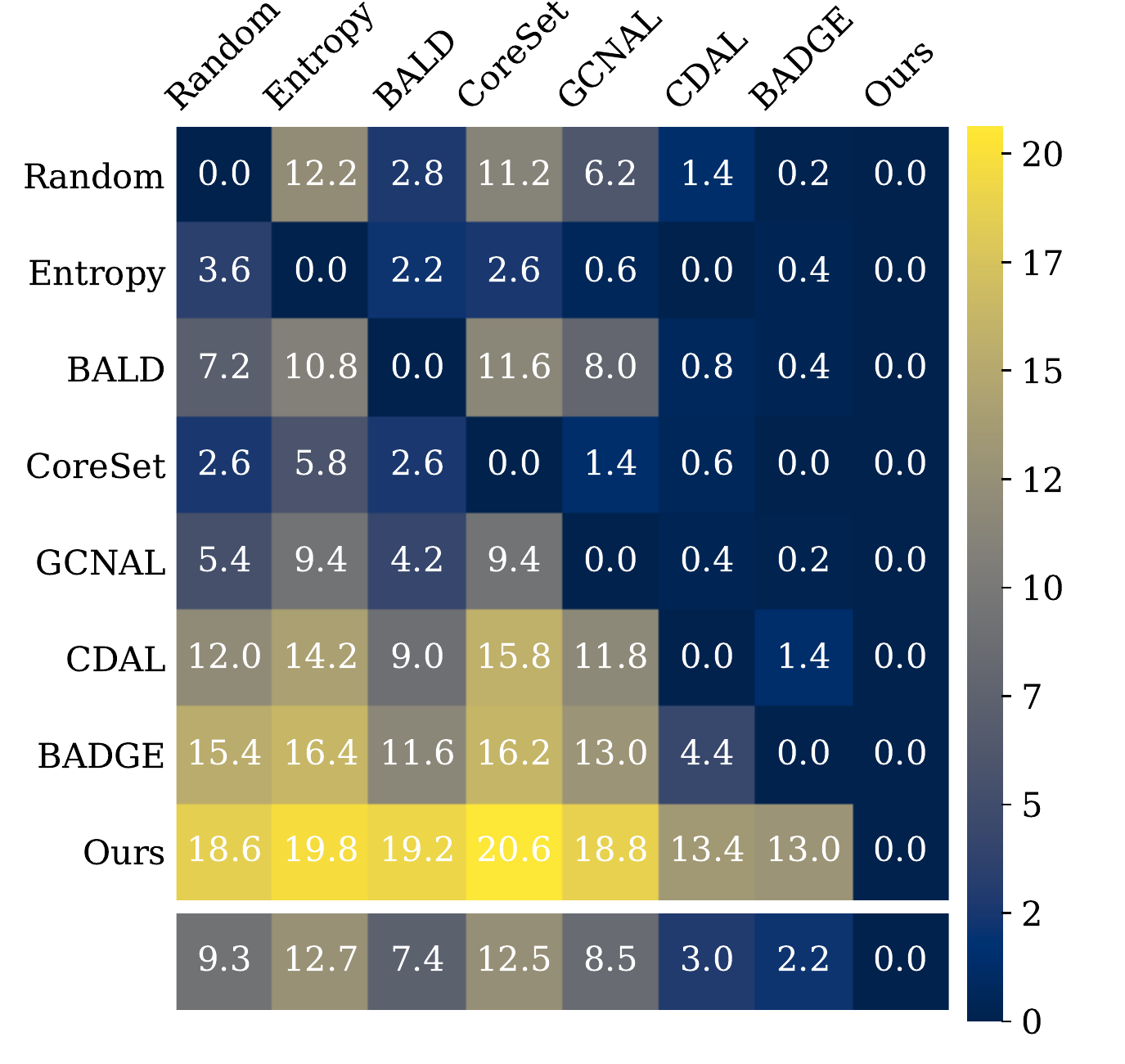}
			\vspace{-4mm}
			\caption{\small{Low-data regime (maximum value: 22)}} \label{fig:comp_low}
		\end{subfigure}	
		\vspace{-4mm}
		\caption{\small{Pairwise comparison of different AL approaches based on different sizes of budget. The maximum value of each cell for each setting is also provided in the captions.}} \label{fig:ablation_budget_size}
	    \vspace{-6mm}
	\end{figure*}
	
	\begin{figure*}[t]
		\begin{subfigure}{0.5\textwidth}
			\centering
			\includegraphics[width=.90\textwidth]{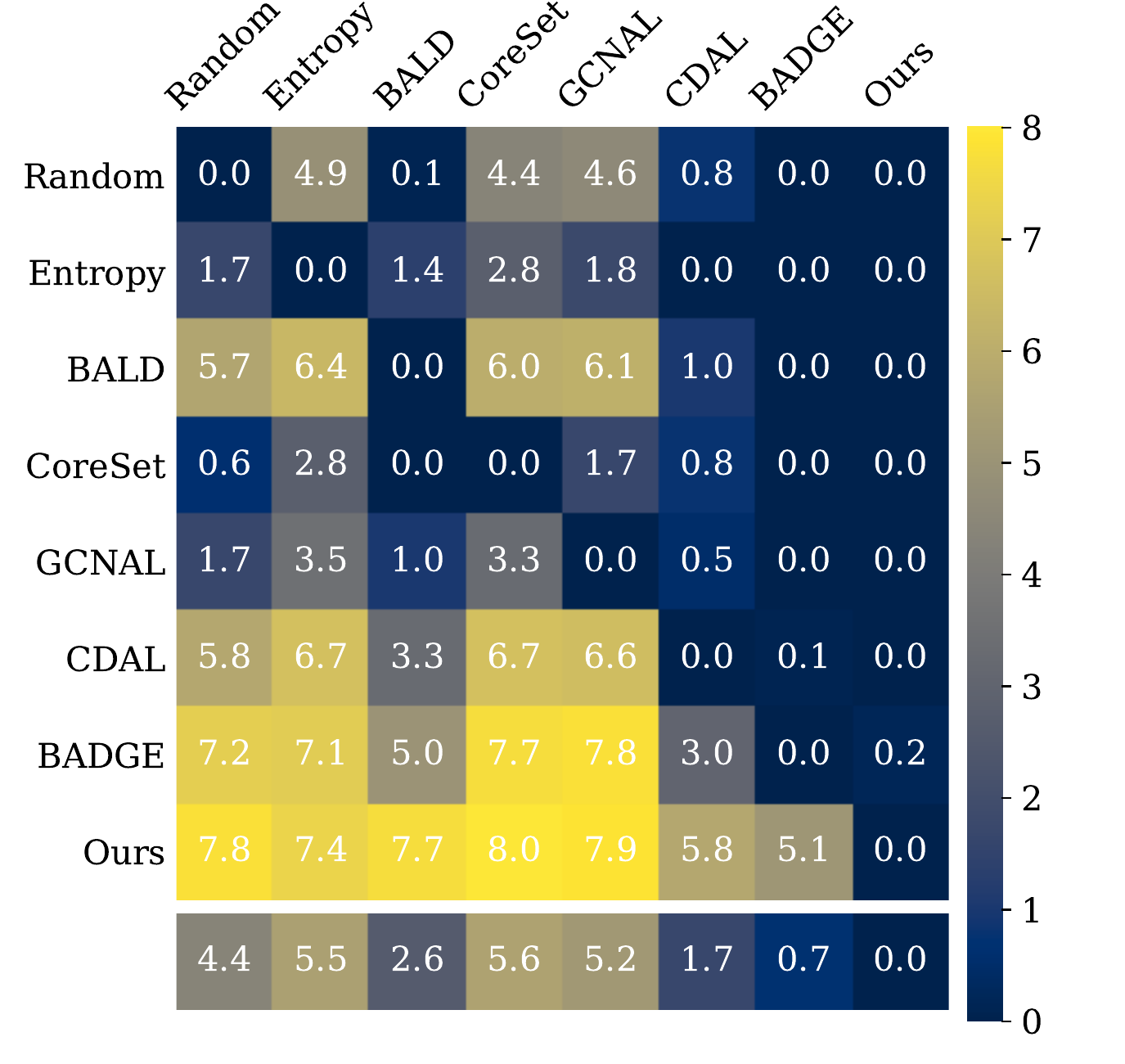}
			\vspace{-4mm}
			\caption{\small{Two-layer MLP (maximum value: 8)}} \label{fig:comp_mlp}
		\end{subfigure}		
		\begin{subfigure}{0.5\textwidth}
			\centering
			\includegraphics[width=.90\textwidth]{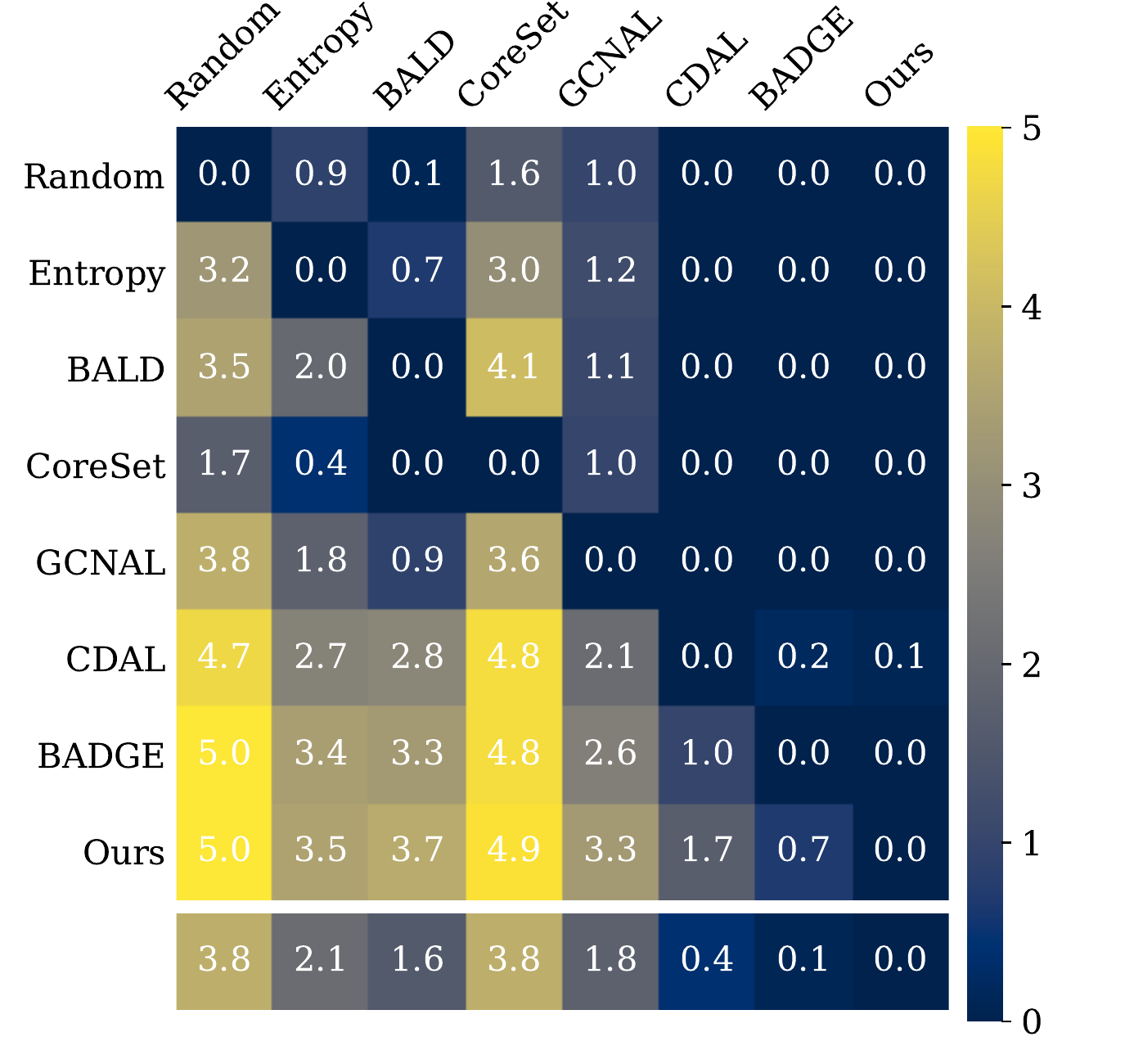}
			\vspace{-4mm}
			\caption{\small{LeNet-5 (maximum value: 5)}} \label{fig:comp_lenet}
		\end{subfigure}
		\begin{subfigure}{0.5\textwidth}
			\centering
			\includegraphics[width=.90\textwidth]{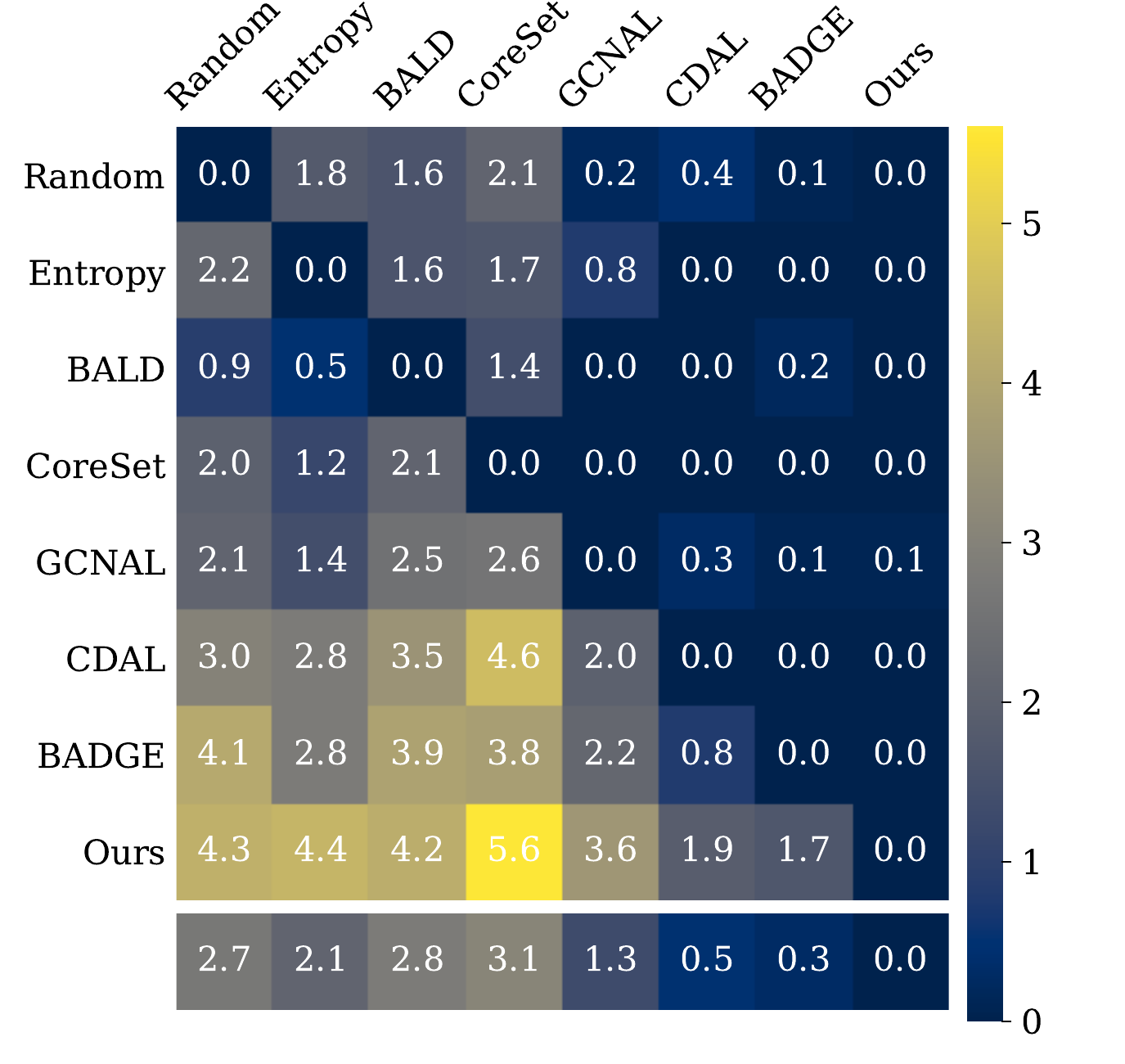}
			\vspace{-4mm}
			\caption{\small{ResNet-18 (maximum value: 7)}} \label{fig:comp_resnet}
		\end{subfigure}	
		\begin{subfigure}{0.5\textwidth}
			\centering
			\includegraphics[width=.90\textwidth]{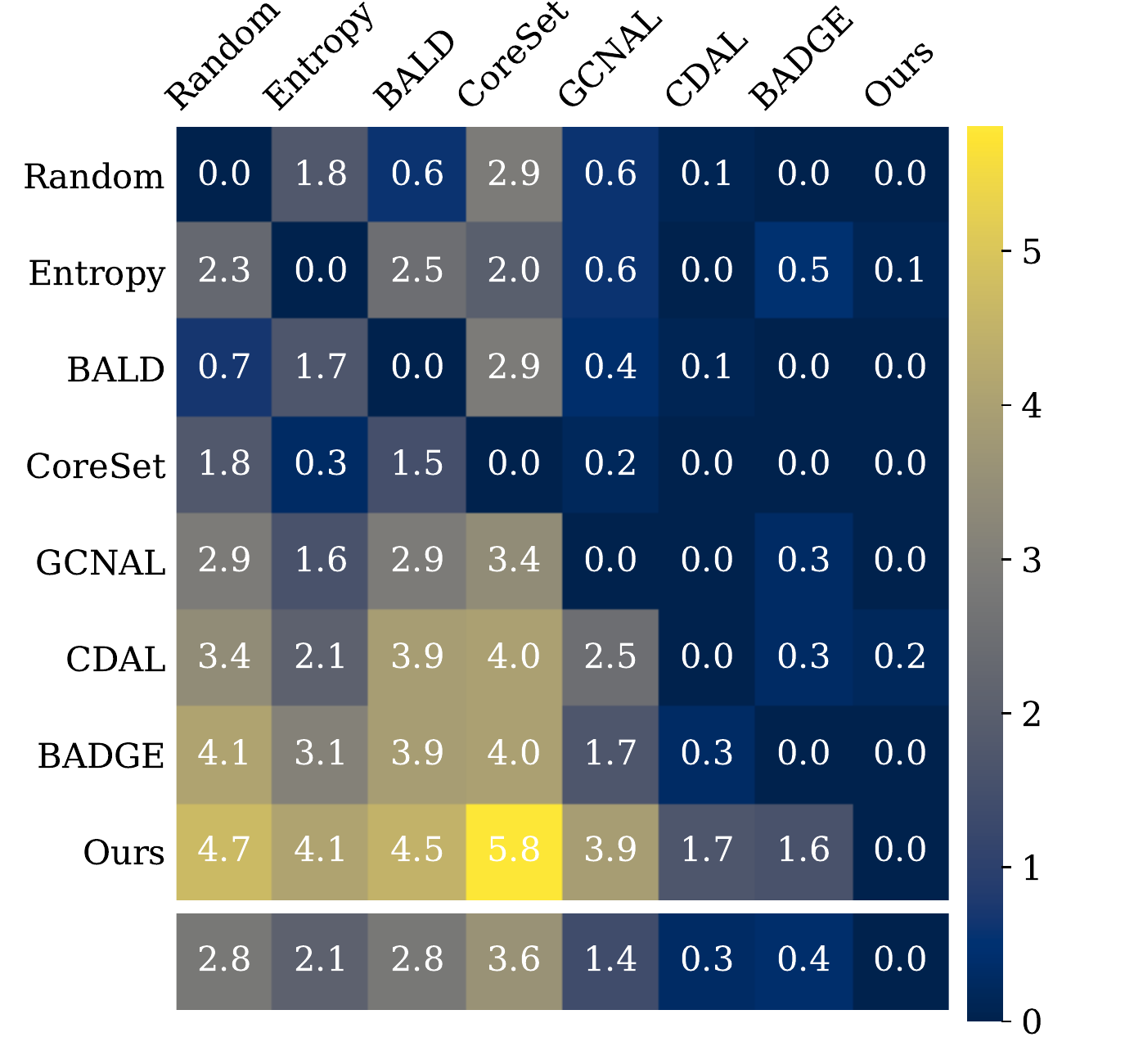}
			\vspace{-4mm}
			\caption{\small{DenseNet-121 (maximum value: 7)}} \label{fig:comp_densenet}
		\end{subfigure}
		\begin{subfigure}{0.5\textwidth}
			\centering
			\includegraphics[width=.90\textwidth]{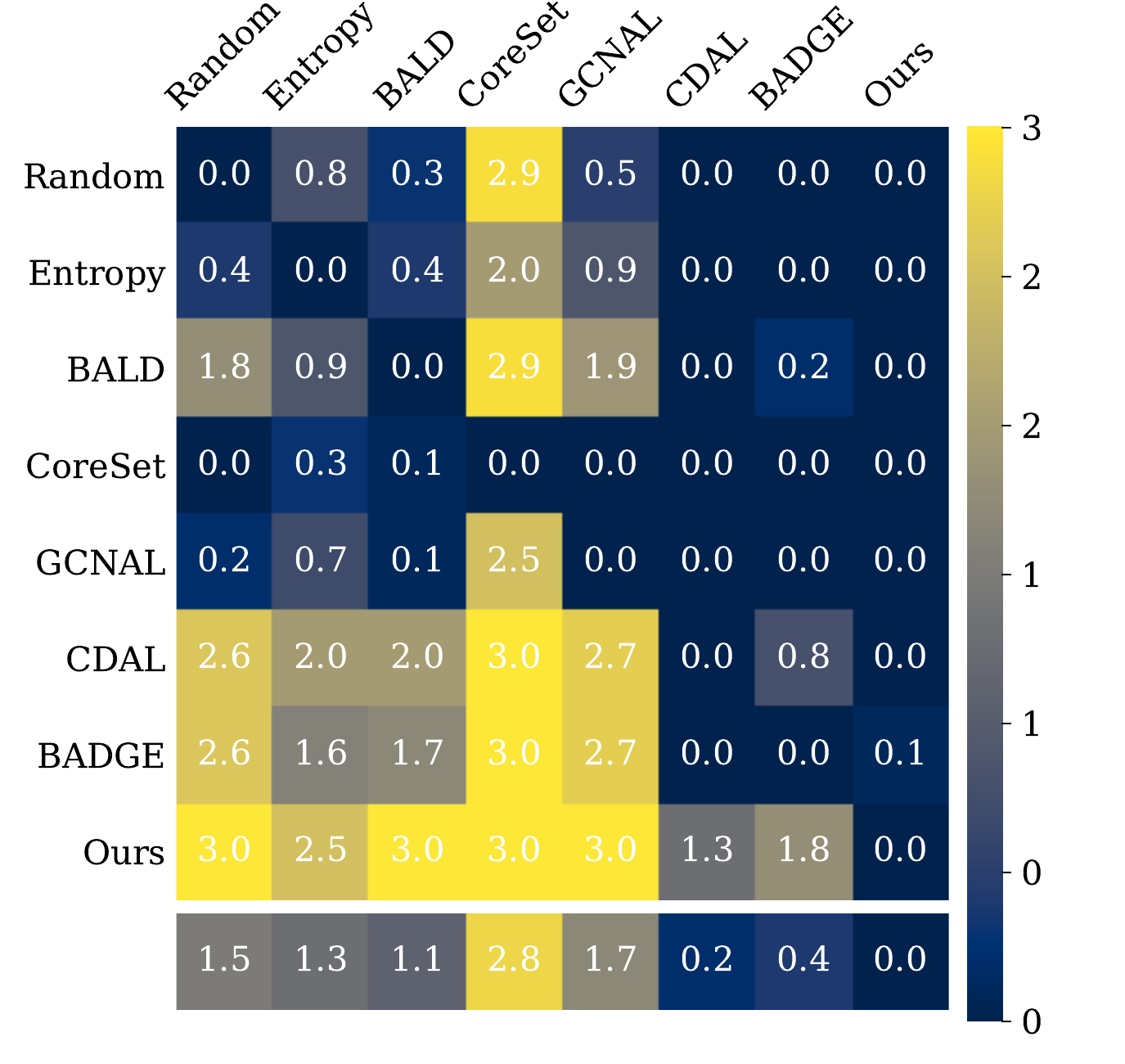}
			\vspace{-4mm}
			\caption{\small{ViT (maximum value: 3)}} \label{fig:comp_densenet}
		\end{subfigure}
		\vspace{-4mm}
		\caption{\small{Pairwise comparison of different AL approaches based on different model architectures. The maximum value of each cell for each setting is also provided in the captions.}} \label{fig:ablation_arch}
	\end{figure*}
	
	\begin{figure*}[t]
		\begin{subfigure}{0.5\textwidth}
			\centering
			\includegraphics[width=.90\textwidth]{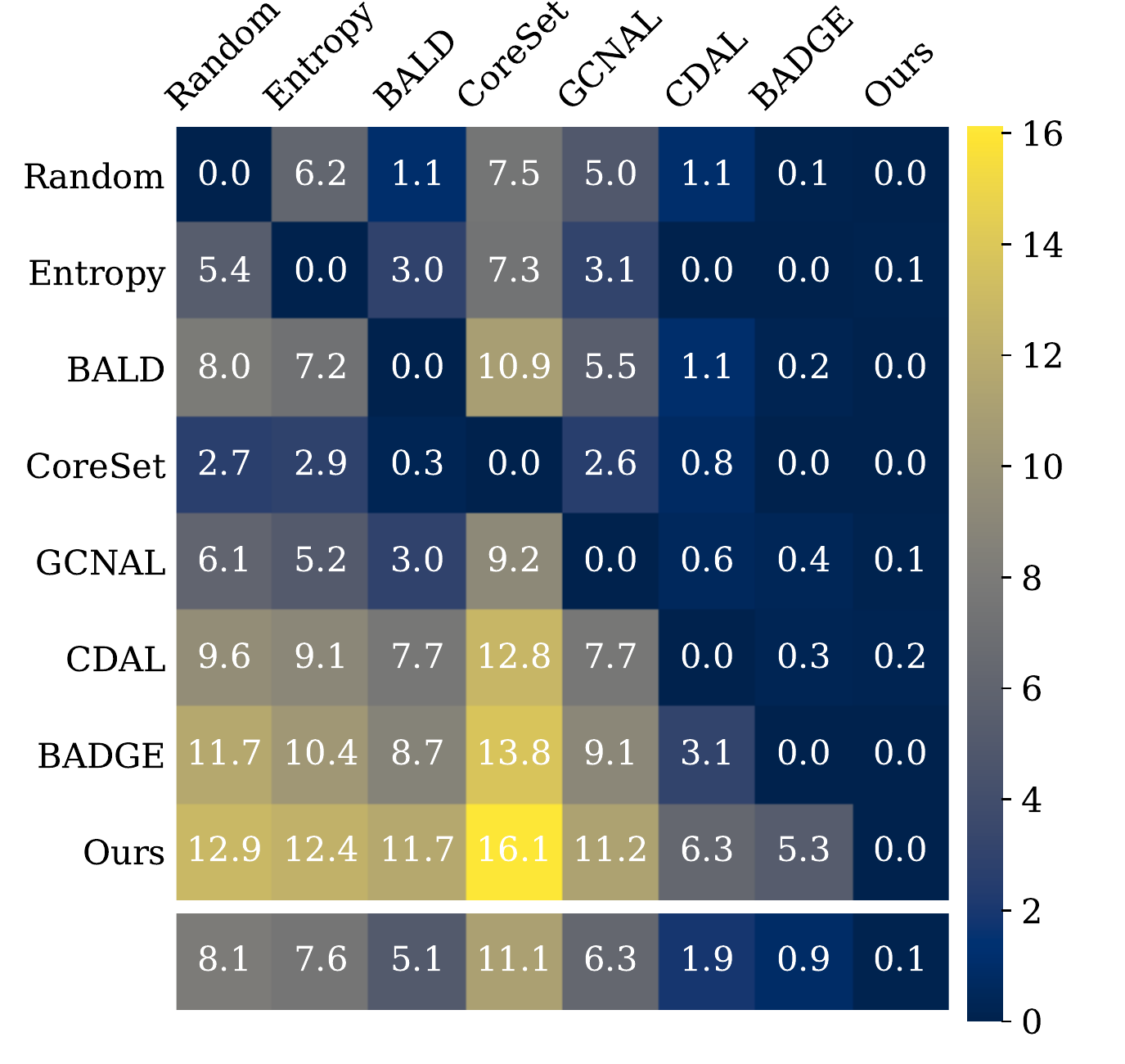}
			\vspace{-4mm}
			\caption{\small{Random (maximum value: 18)}} \label{fig:comp_random}
		\end{subfigure}	
		\begin{subfigure}{0.5\textwidth}
			\centering
			\includegraphics[width=.90\textwidth]{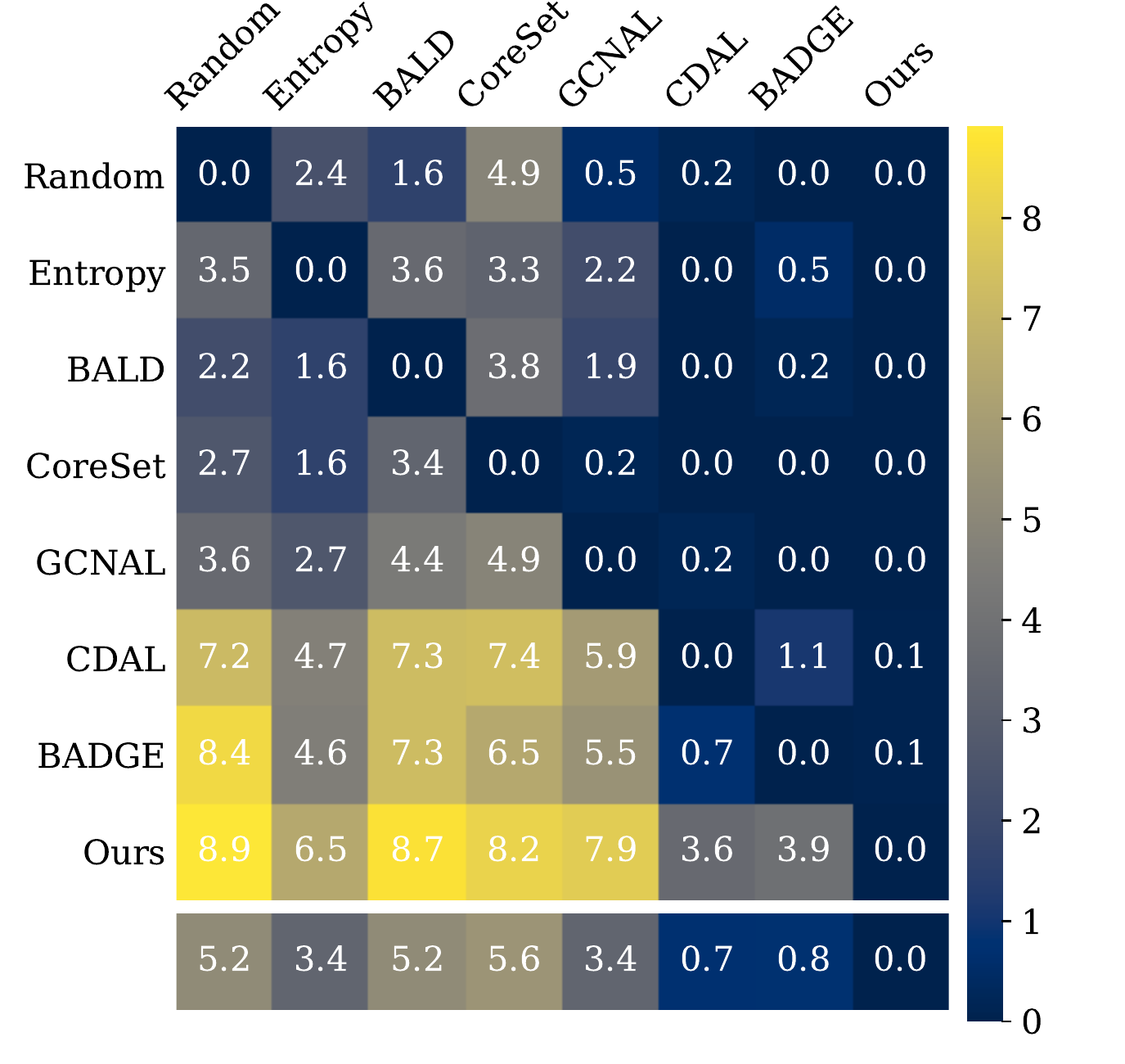}
			\vspace{-4mm}
			\caption{\small{Pre-Training (maximum value: 9)}} \label{fig:comp_pre_train}
		\end{subfigure}		
		\begin{subfigure}{0.5\textwidth}
			\centering
			\includegraphics[width=.90\textwidth]{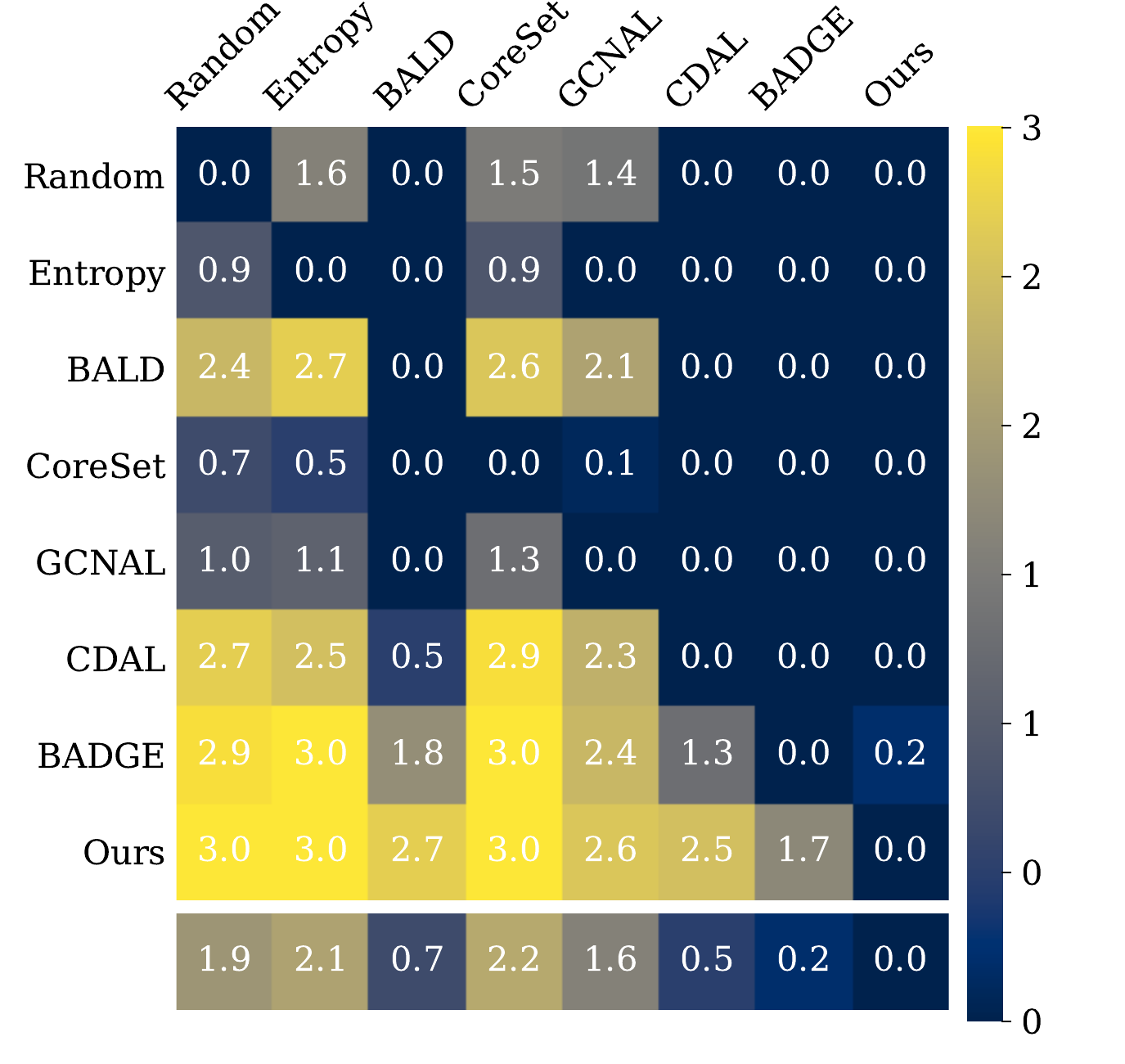}
			\vspace{-4mm}
			\caption{\small{Continue (maximum value: 3)}} \label{fig:comp_continue}
		\end{subfigure}
		\vspace{-4mm}
		\caption{\small{Pairwise comparison of different AL approaches based on different sizes of budget. The maximum value of each cell for each setting is also provided in the captions.}} \label{fig:ablation_initialisation}
	\end{figure*}

	\subsection{More Ablations}
	
	\vspace{-2mm}
	In addition to providing the percentage with which our approach outperforms others in each setting (Table.~\ref{tab:ablations}), we report the pairwise comparison of all the AL methods across various choices of data (Fig.~\ref{fig:ablation_data_type}), budget size (Fig.~\ref{fig:ablation_budget_size}), model architecture (Fig.~\ref{fig:ablation_arch}) and network initialisation method (Fig.~\ref{fig:ablation_initialisation}. Further, in Figure~\ref{fig:comp_low}, we provide the pairwise comparisons in low-data regimes. Considering the values in the rows and columns corresponding to our approach, we can infer that our approach consistently outperforms all other baselines regardless of the architecture, dataset selection, network initialisation and budget size and is rarely beaten by others.
	
	\vspace{-2mm}
	\subsection{All the Experiments}
	\vspace{-2mm}
	
	We compare our approach with other baselines over a total of 285 AL rounds in 30 different settings, with each setting identified by a specific combination of dataset, budget size, model architecture, and model initialisation method.
	Table~\ref{tab:ablations} demonstrates details of each setting we employed in our experiments.
	
	All the experiments for small datasets were carried out on a NVIDIA GEFORCE GTX 1080 Ti, while for larger datasets we used an NVIDIA QUADRO RTX 8000. It is worth mentioning that for the video experiments, we utilised two NVIDIA V100 GPUs in parallel.
	
	We borrowed the implementations of the baselines from their publicly provided codes. The MLP network we employed in our experiments follows the architecture proposed in \cite{badge_iclr_2020}: a two-layer Perceptron with ReLU activations and an embedding dimension of size 256 for image datasets (\ie MNIST and EMNIST). Similarly, we expanded the embedding dimensionality to 1024 for OpenML datasets. We include the accuracy curves over the unseen test set for all the settings.

	\begin{figure}[t]
		\centering
		\includegraphics[width=1.\linewidth]{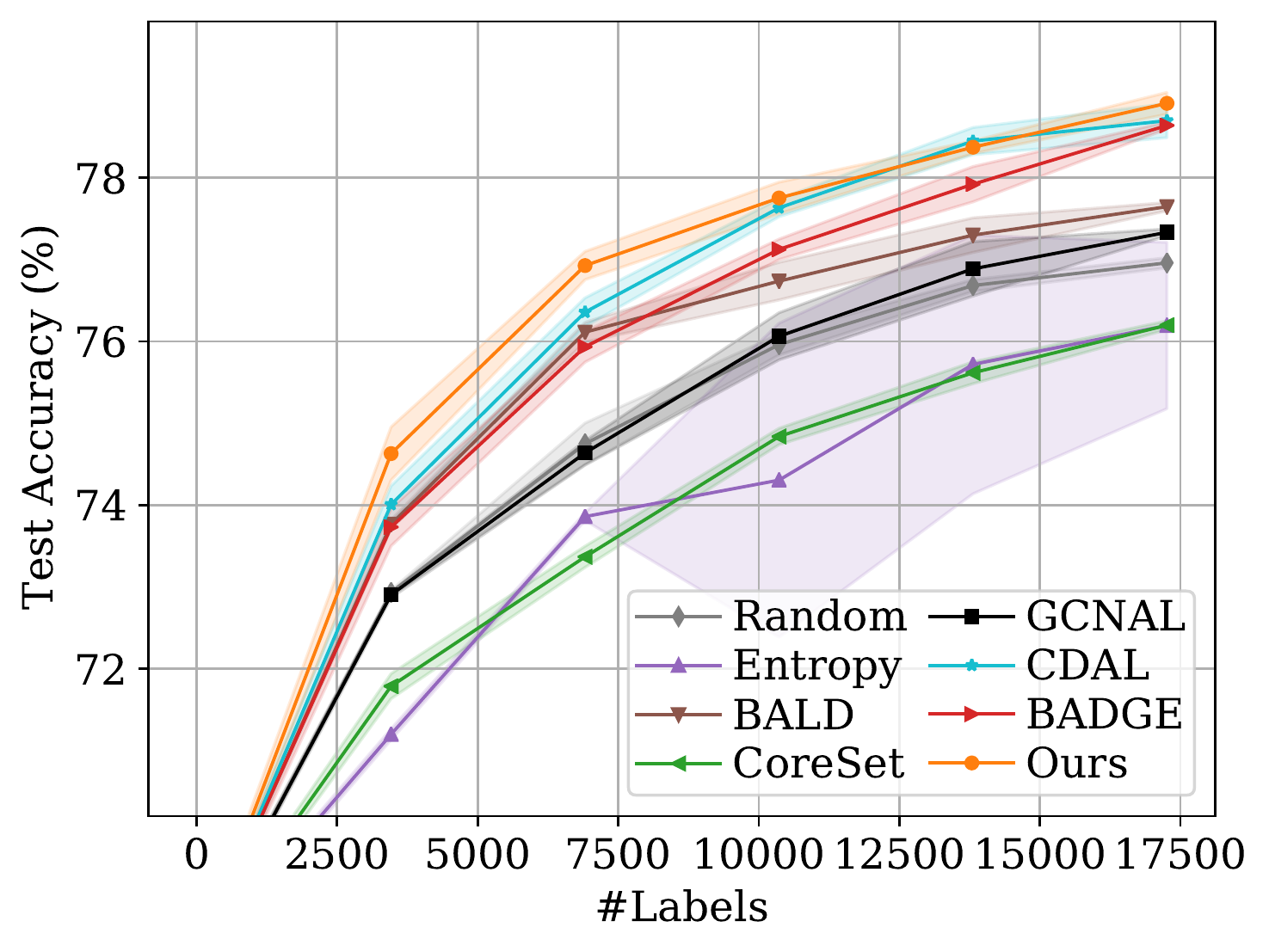}
		\vspace{-8mm}
		\caption{Small Budget, ViT-Base, DomainNet-Real}
		\vspace{-2mm}
	\end{figure}
	\begin{figure}[t]
		\centering
		\includegraphics[width=1.\linewidth]{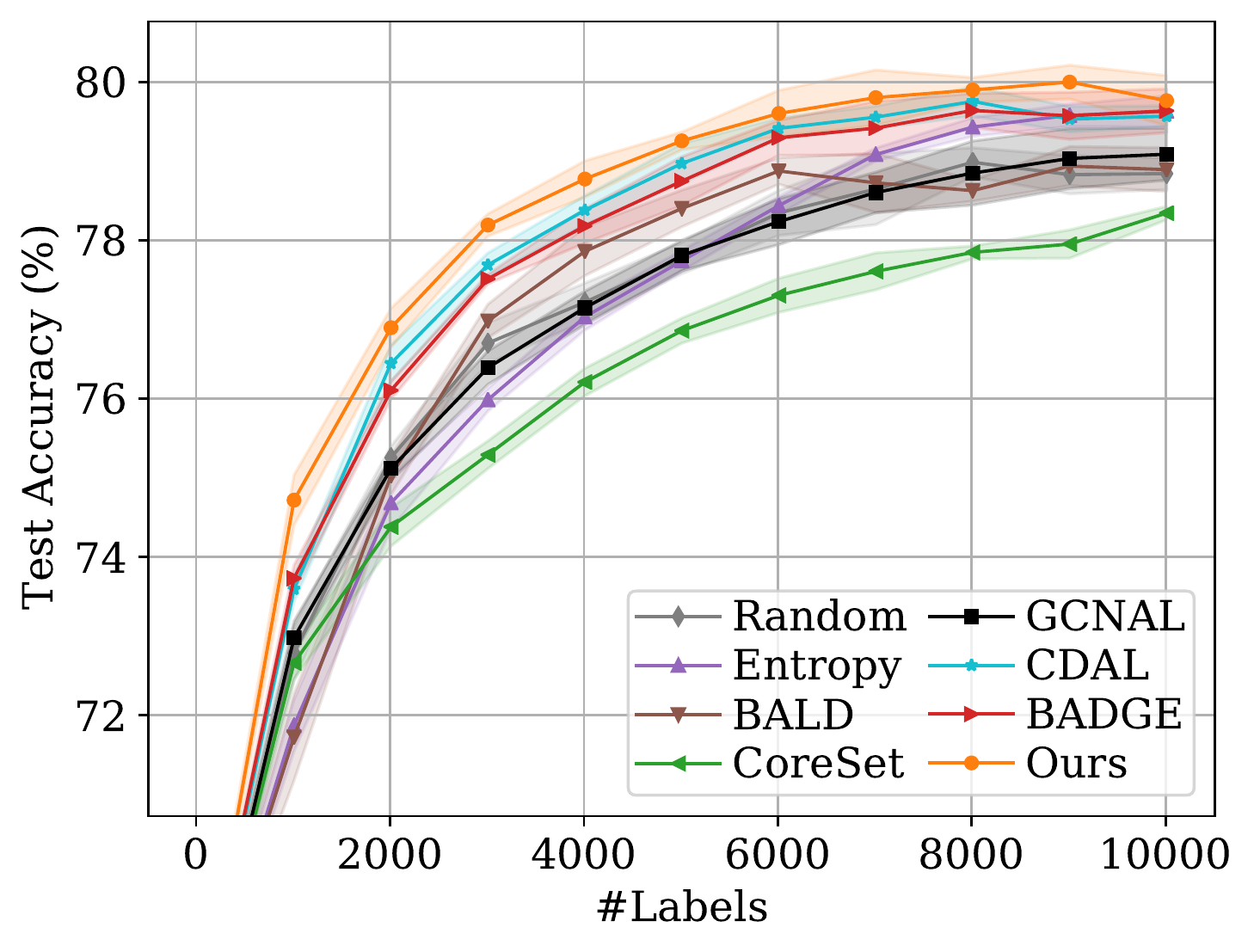}
		\vspace{-8mm}
		\caption{Small Budget, ViT-Small, Mini-ImageNet}
		\vspace{-2mm}
	\end{figure}
	\begin{figure}[t]
		\centering
		\includegraphics[width=1.\linewidth]{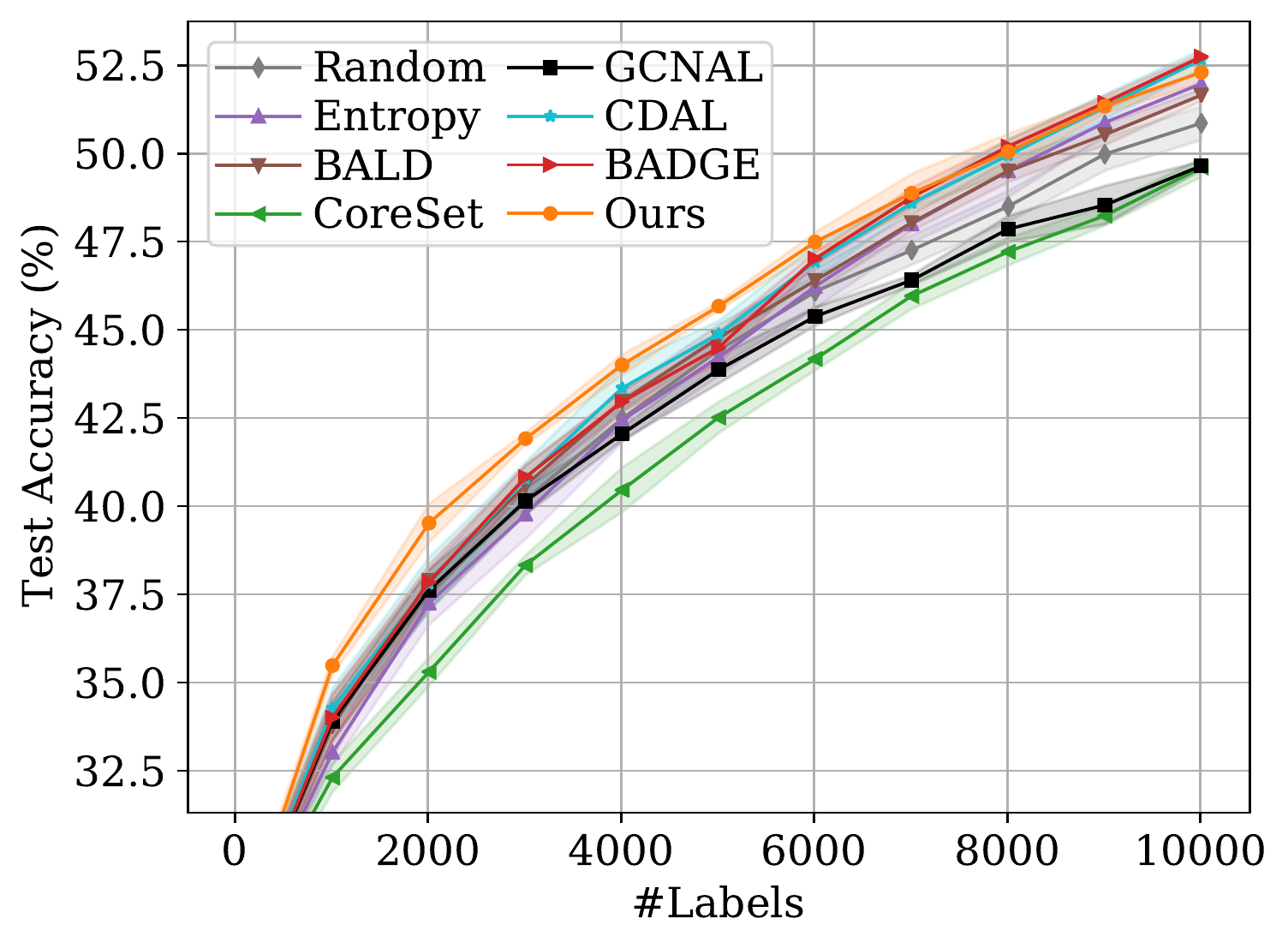}
		\vspace{-8mm}
		\caption{Small Budget, ViT-Small, CIFAR100}
		\vspace{-2mm}
	\end{figure}

	\begin{figure}[t]
		\centering
		\includegraphics[width=1.\linewidth]{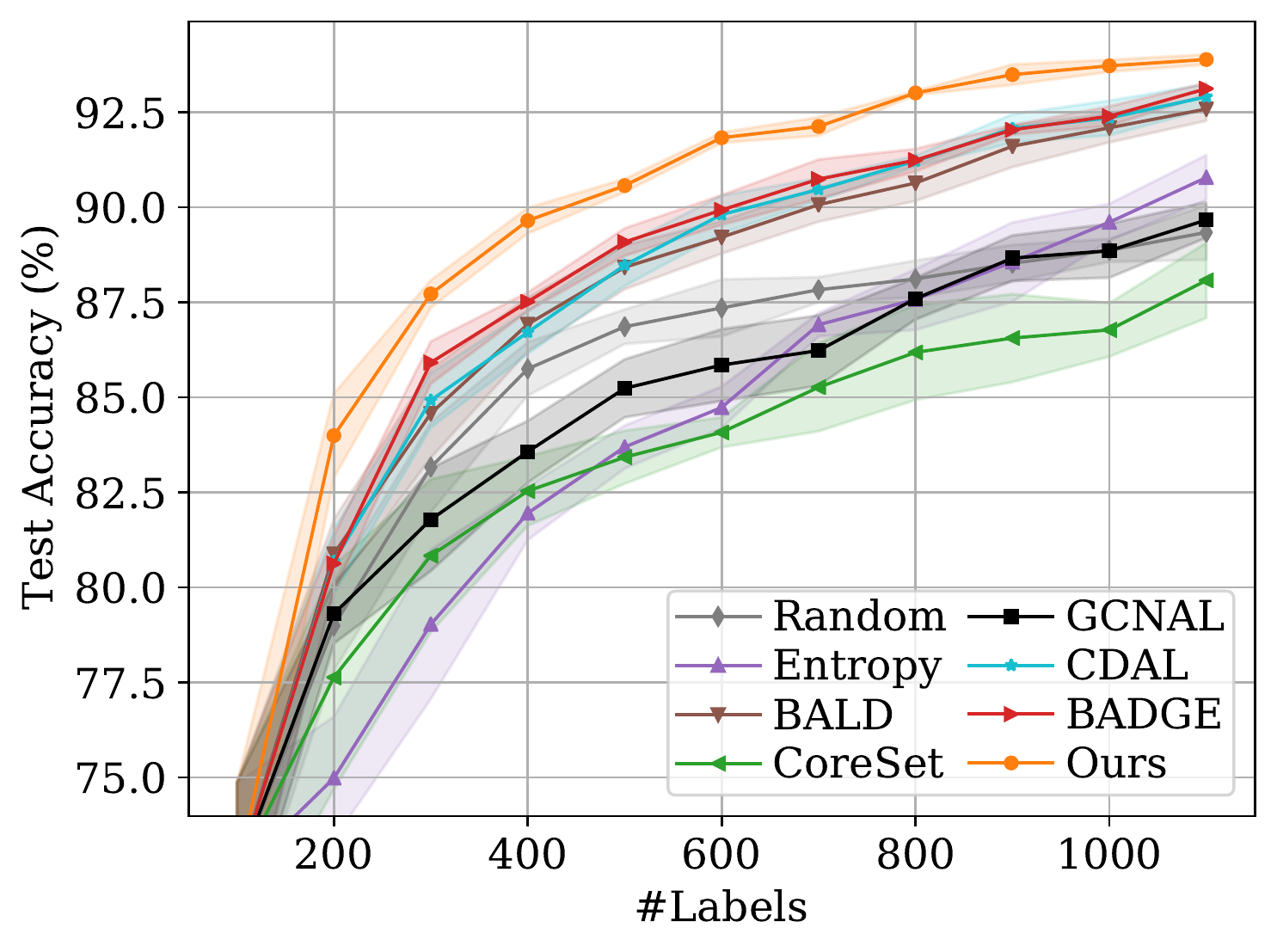}
		\vspace{-8mm}
		\caption{Small Budget, MLP, MNIST}
		\vspace{-2mm}
	\end{figure}
	\begin{figure}[t]
		\centering
		\includegraphics[width=1.\linewidth]{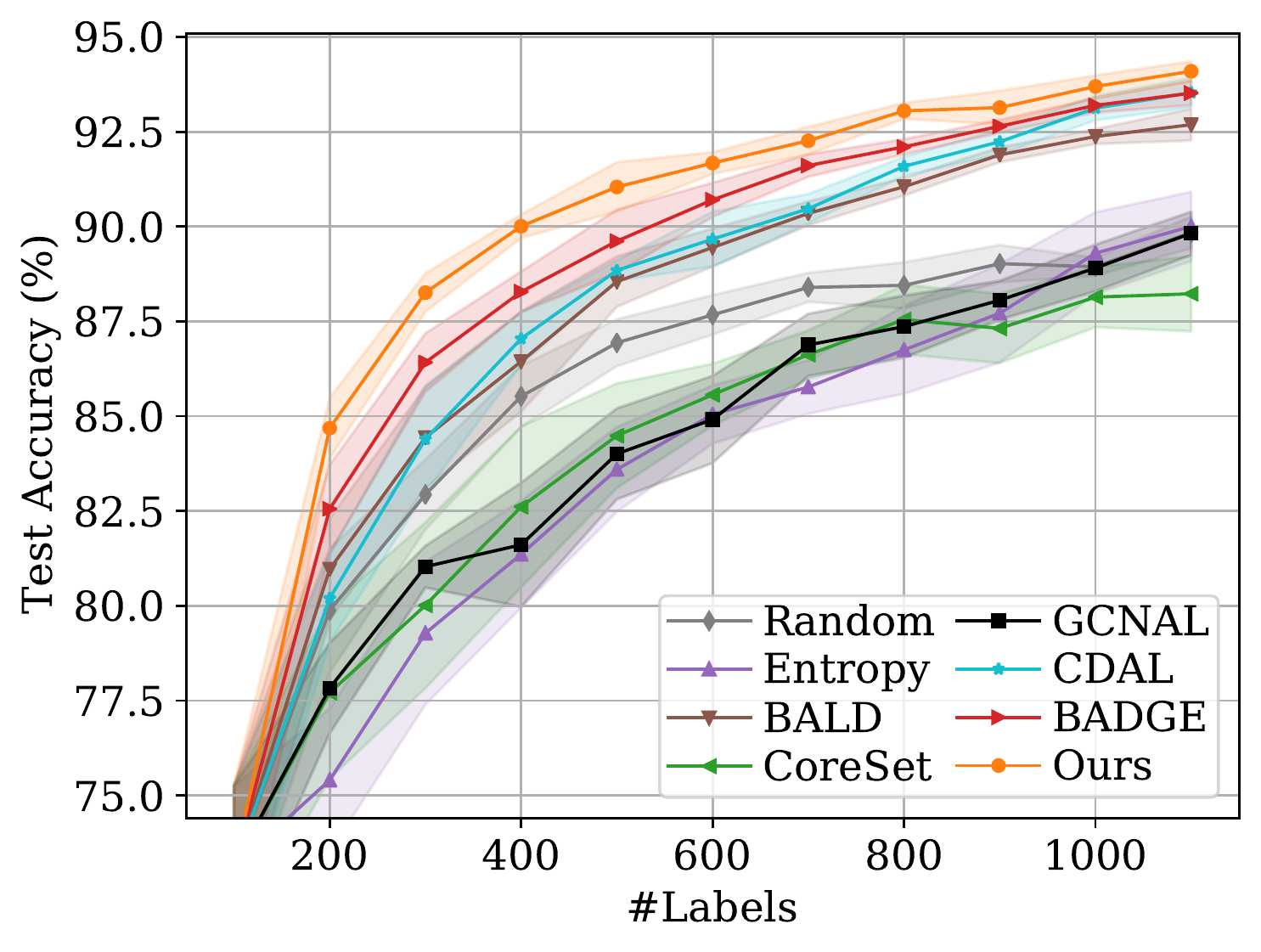}
		\vspace{-8mm}
		\caption{Small Budget, MLP, MNIST, Continue}
		\vspace{-2mm}
	\end{figure}
	\begin{figure}[t]
		\centering
		\includegraphics[width=1.\linewidth]{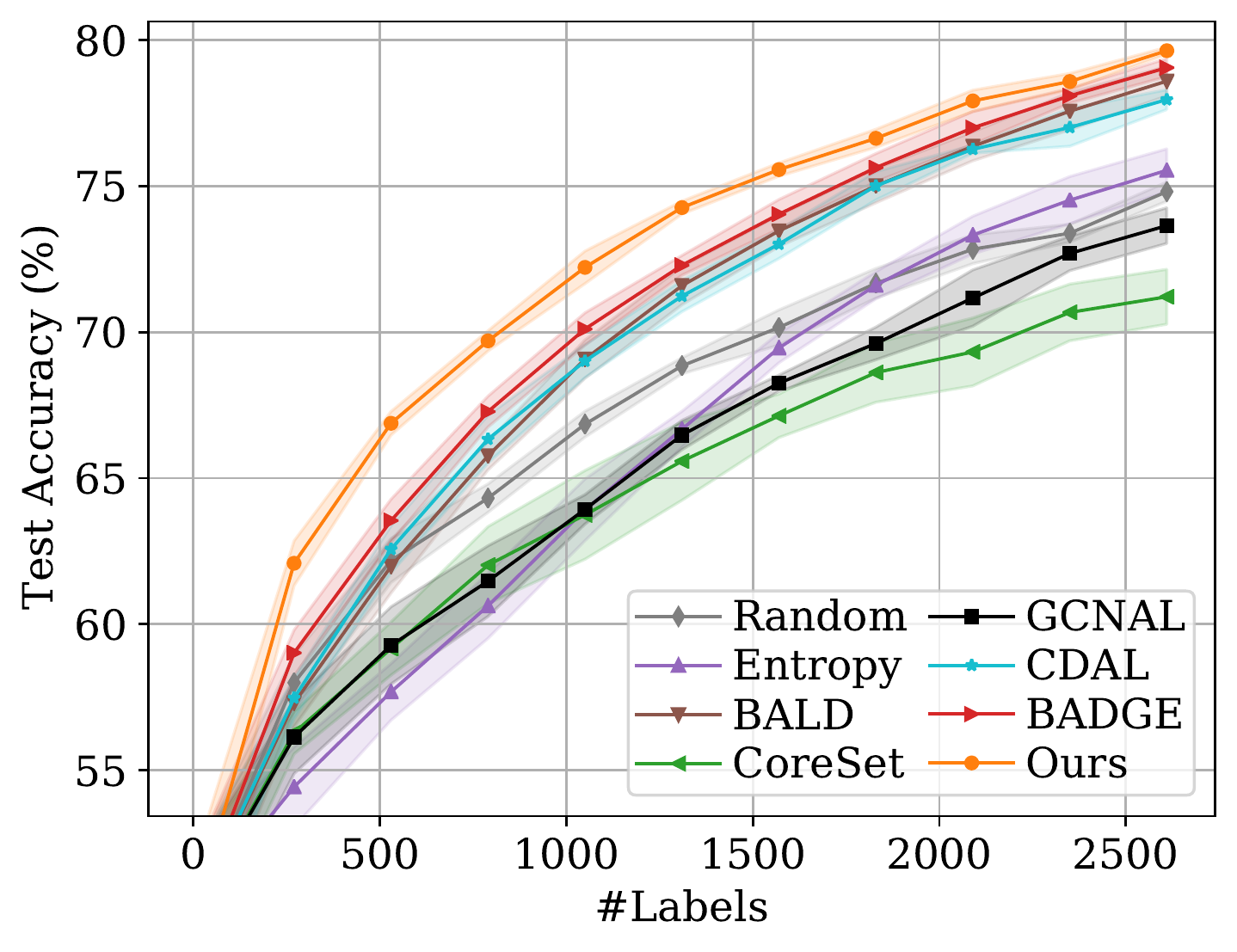}
		\vspace{-8mm}
		\caption{Small Budget, MLP, EMNIST}
		\vspace{-2mm}
	\end{figure}
	\begin{figure}[t]
		\centering
		\includegraphics[width=1.\linewidth]{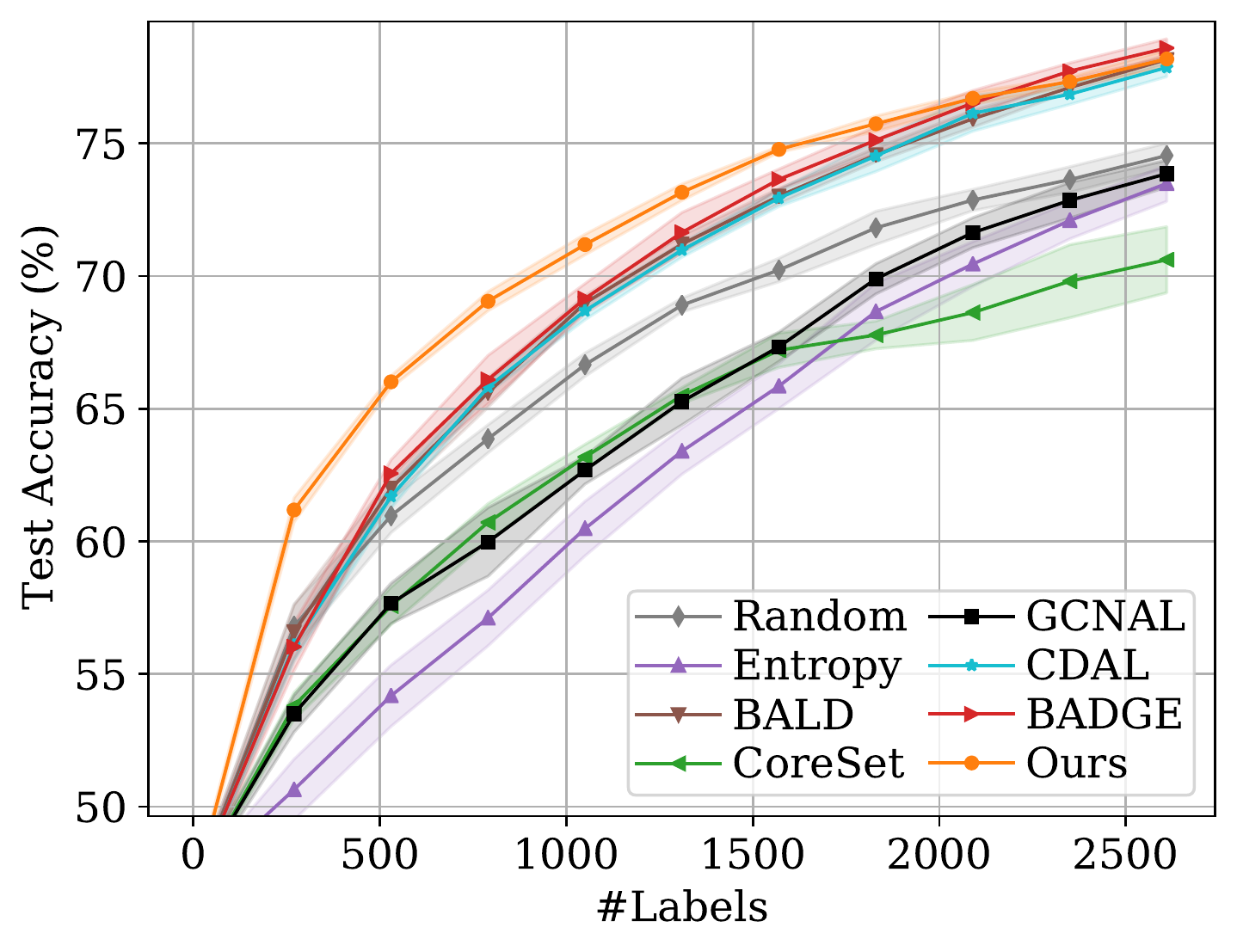}
		\vspace{-8mm}
		\caption{Small Budget, MLP, EMNIST, Continue}
		\vspace{-2mm}
	\end{figure}

	\begin{figure}[t]
		\centering
		\includegraphics[width=1.\linewidth]{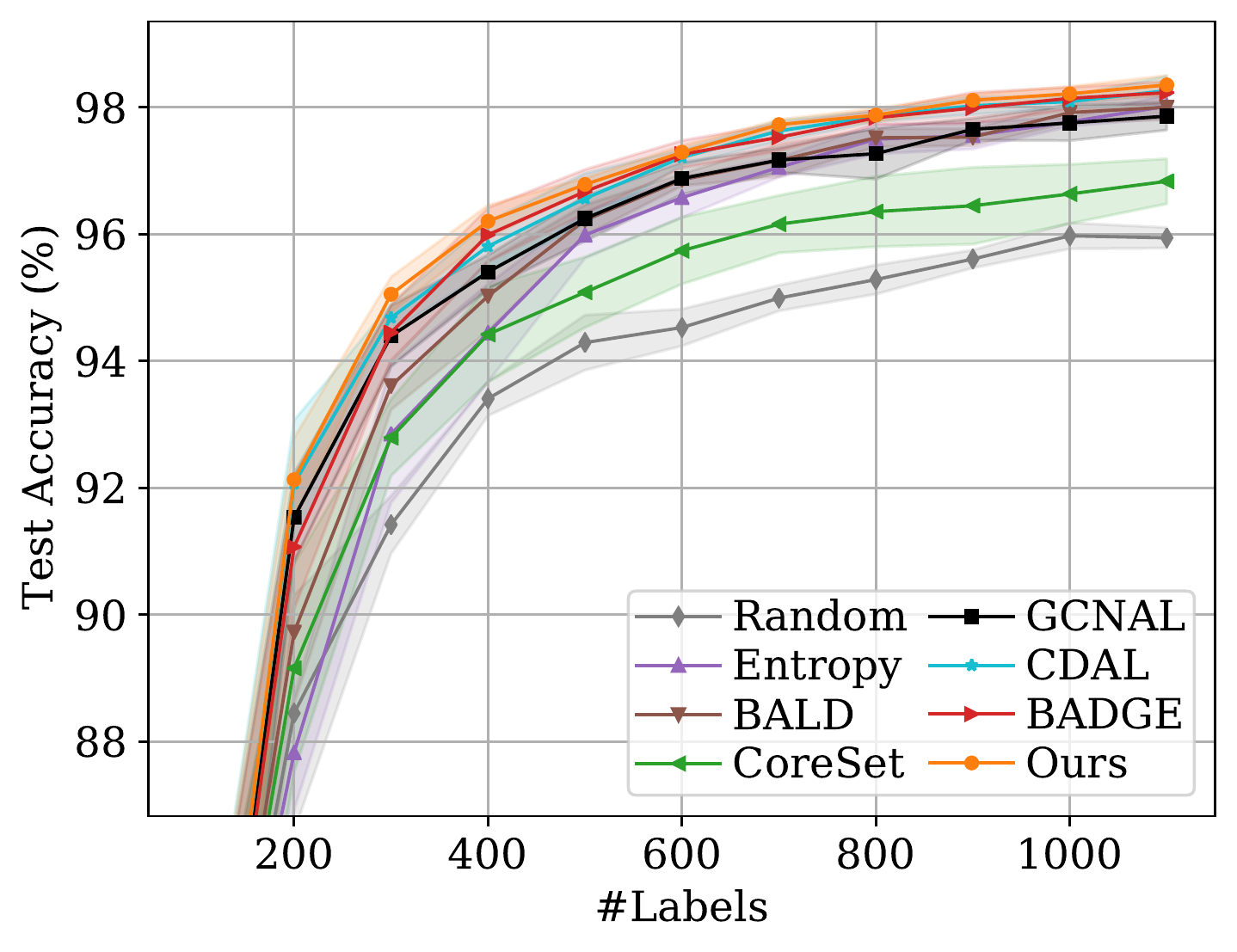}
		\vspace{-8mm}
		\caption{Small Budget, LeNet-5, MNIST}
		\vspace{-2mm}
	\end{figure}
	\begin{figure}[t]
		\centering
		\includegraphics[width=1.\linewidth]{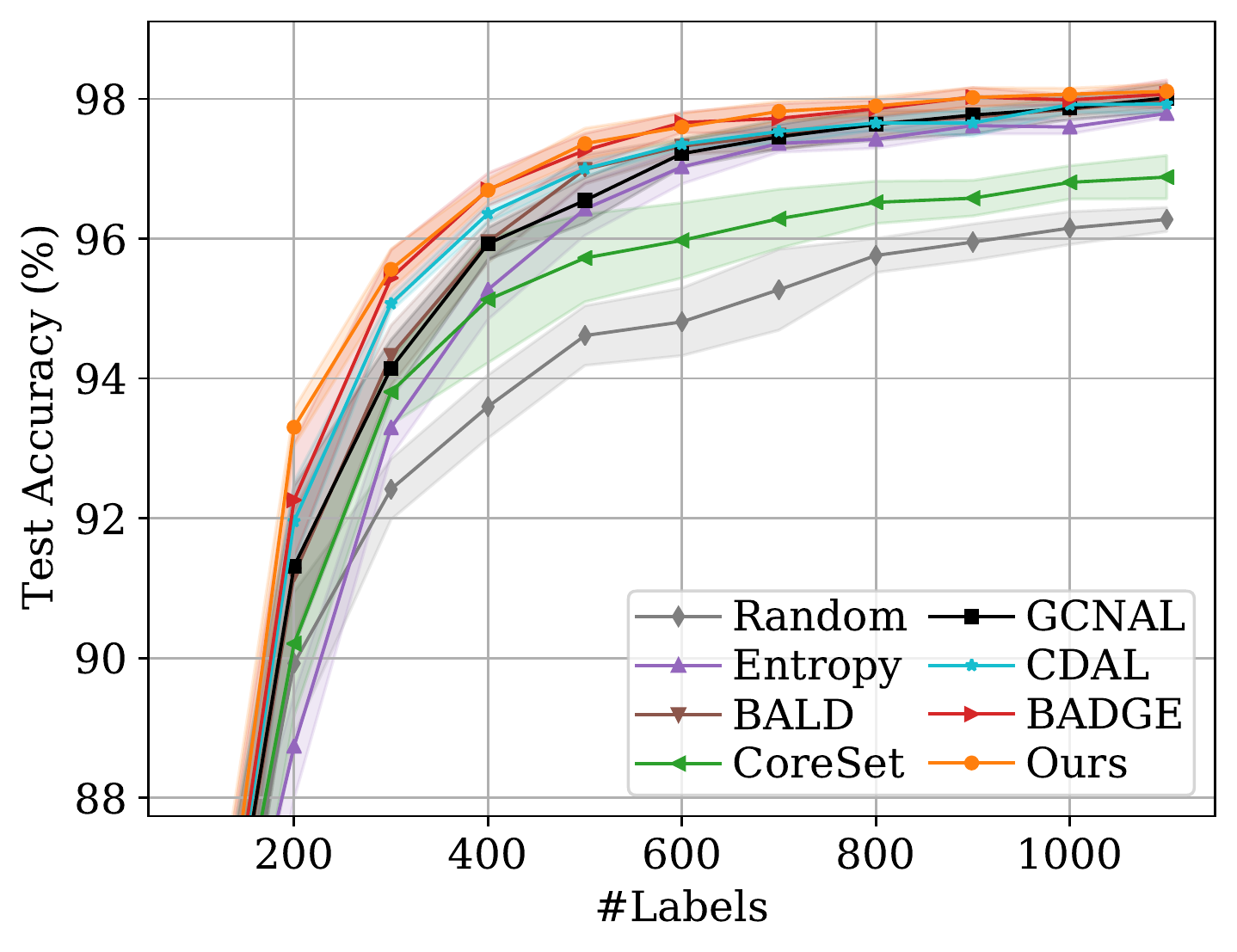}
		\vspace{-8mm}
		\caption{Small Budget, LeNet-5, MNIST, Continue}
		\vspace{-2mm}
	\end{figure}	
	\begin{figure}[t]
		\centering
		\includegraphics[width=1.\linewidth]{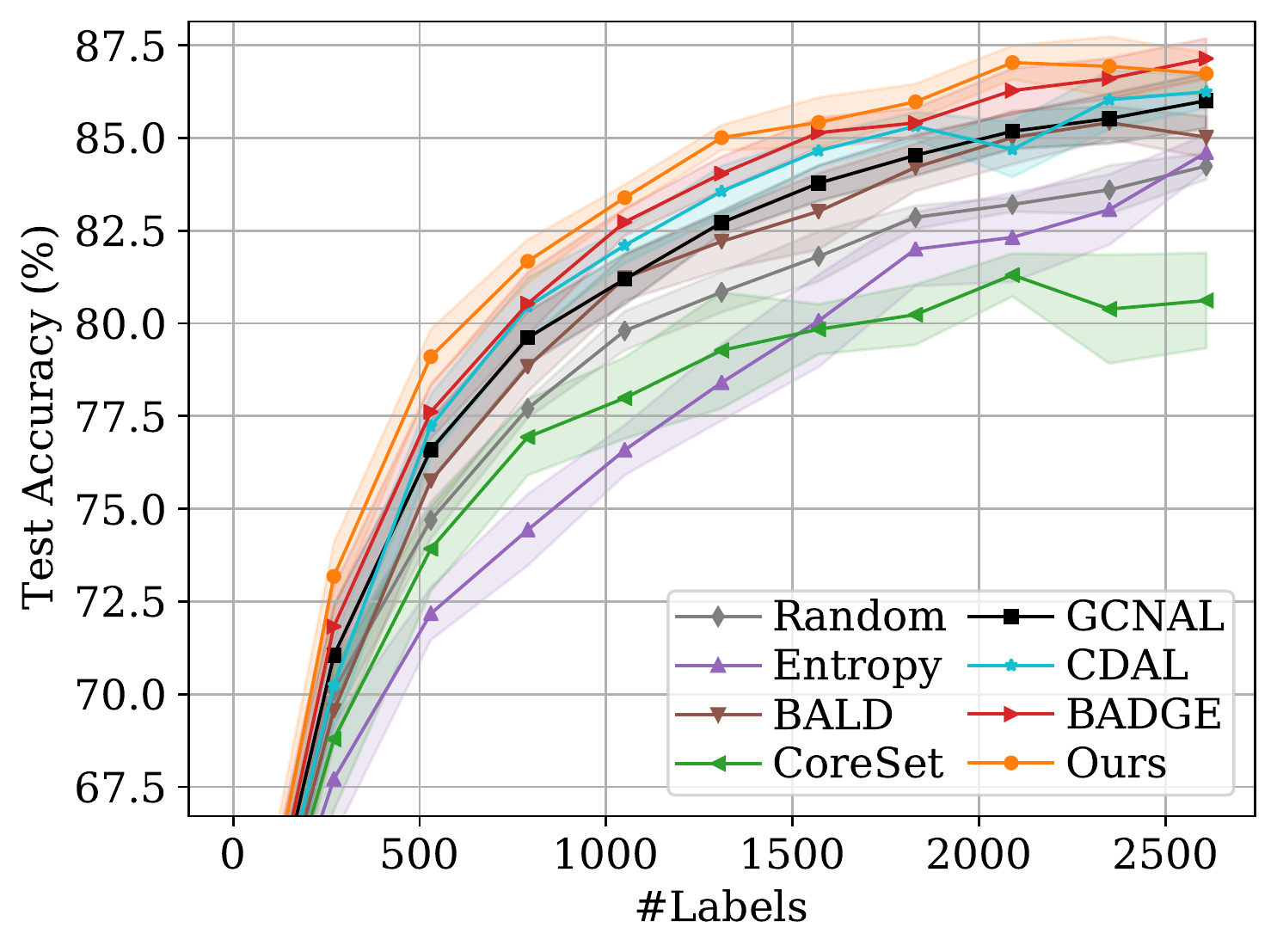}
		\vspace{-8mm}
		\caption{Small Budget, LeNet-5, EMNIST}
		\vspace{-2mm}
	\end{figure}
	
	\begin{figure}[t]
		\centering
		\includegraphics[width=1.\linewidth]{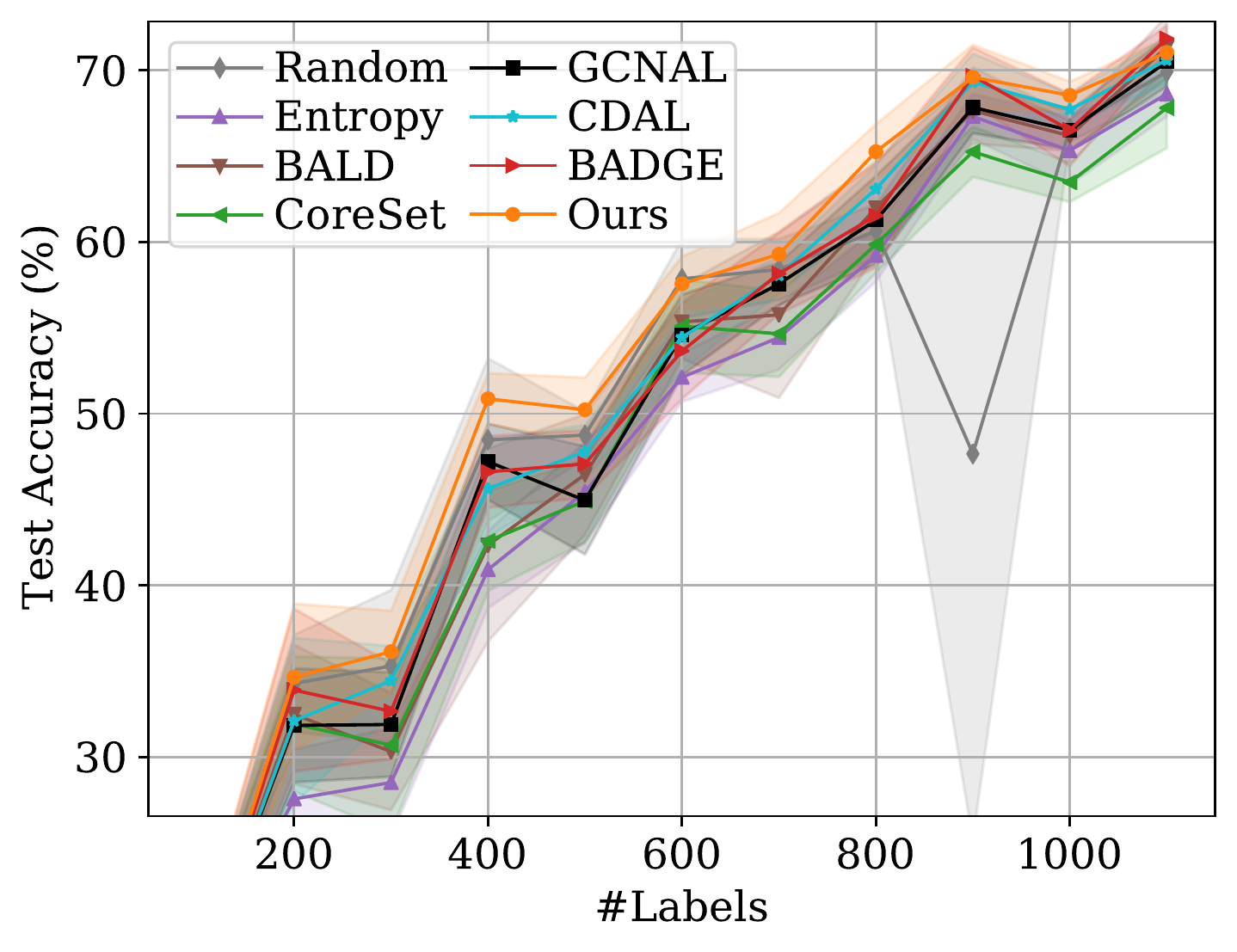}
		\vspace{-8mm}
		\caption{Small Budget-ResNet-18, SVHN}
		\vspace{-2mm}
		\label{fig:plot_domain_net_densenet}
	\end{figure}
	\begin{figure}[t]
		\centering
		\includegraphics[width=1.\linewidth]{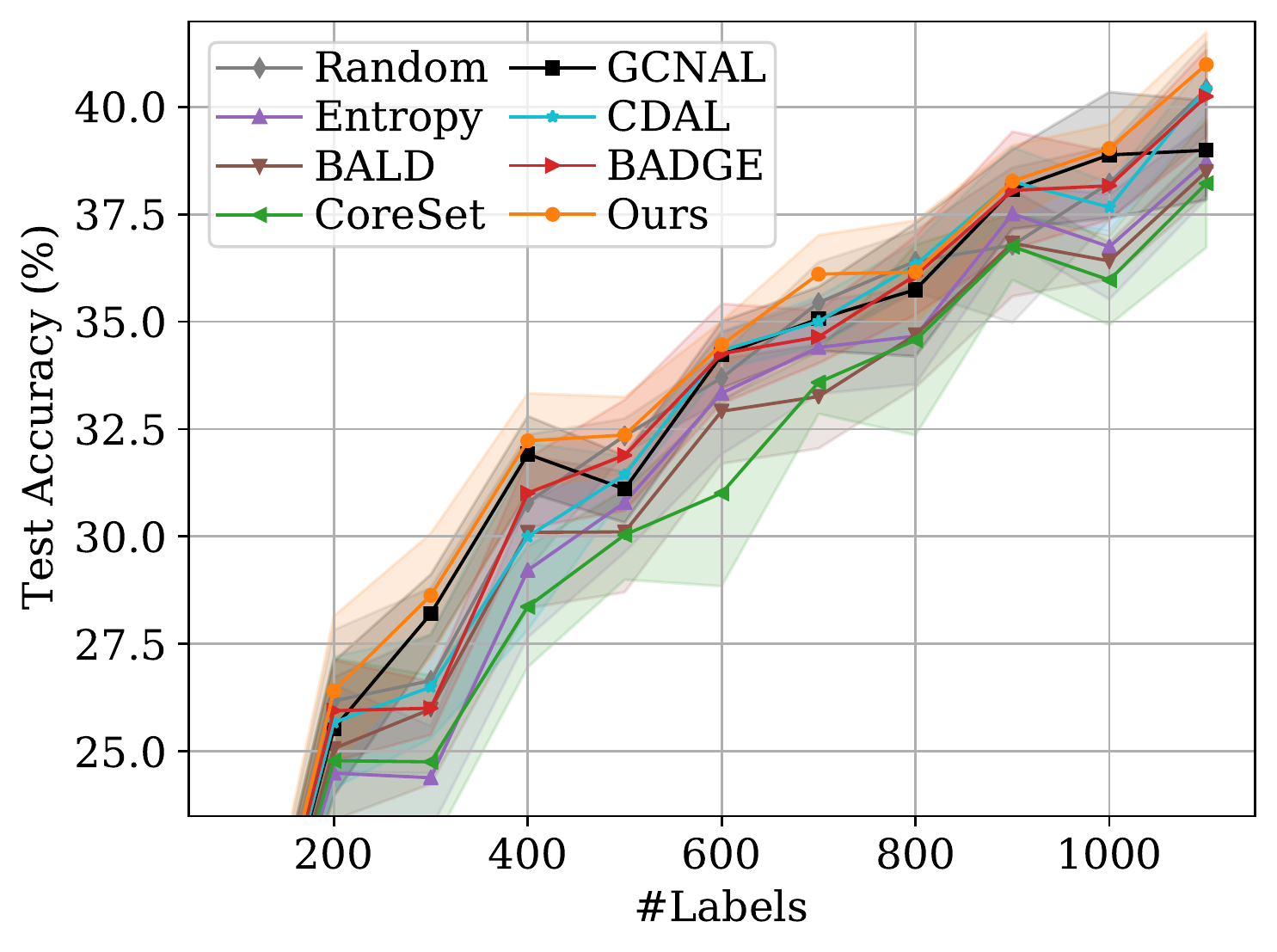}
		\vspace{-8mm}
		\caption{Small Budget, ResNet-18, CIFAR10}
		\vspace{-2mm}
	\end{figure}
	
	\begin{figure}[t]
		\centering
		\includegraphics[width=1.\linewidth]{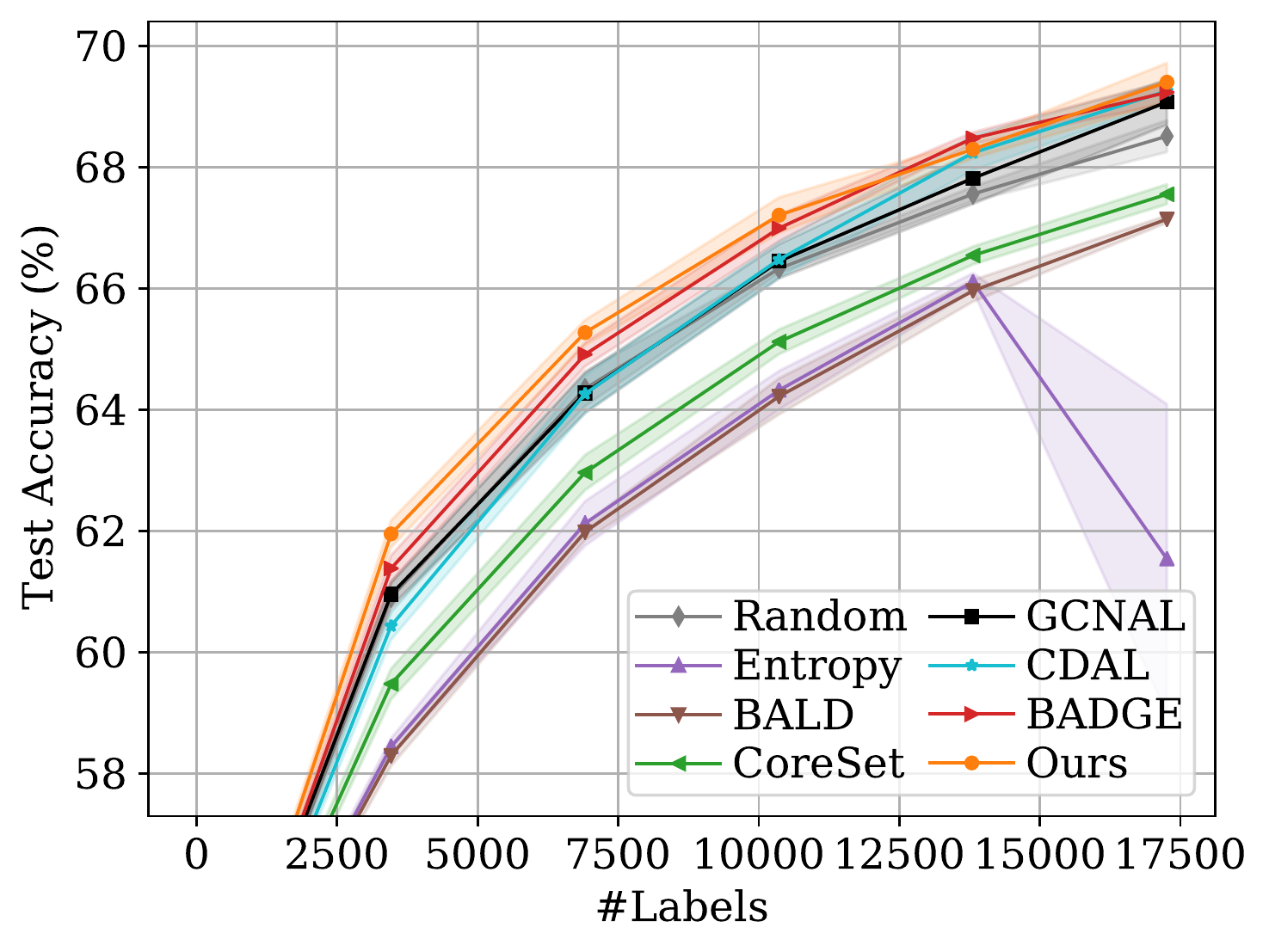}
		\vspace{-5mm}
		\caption{Small Budget, ResNet-18, DomainNet-Real}
		\vspace{-2mm}
	\end{figure}
	\begin{figure}[t]
		\centering
		\includegraphics[width=1.\linewidth]{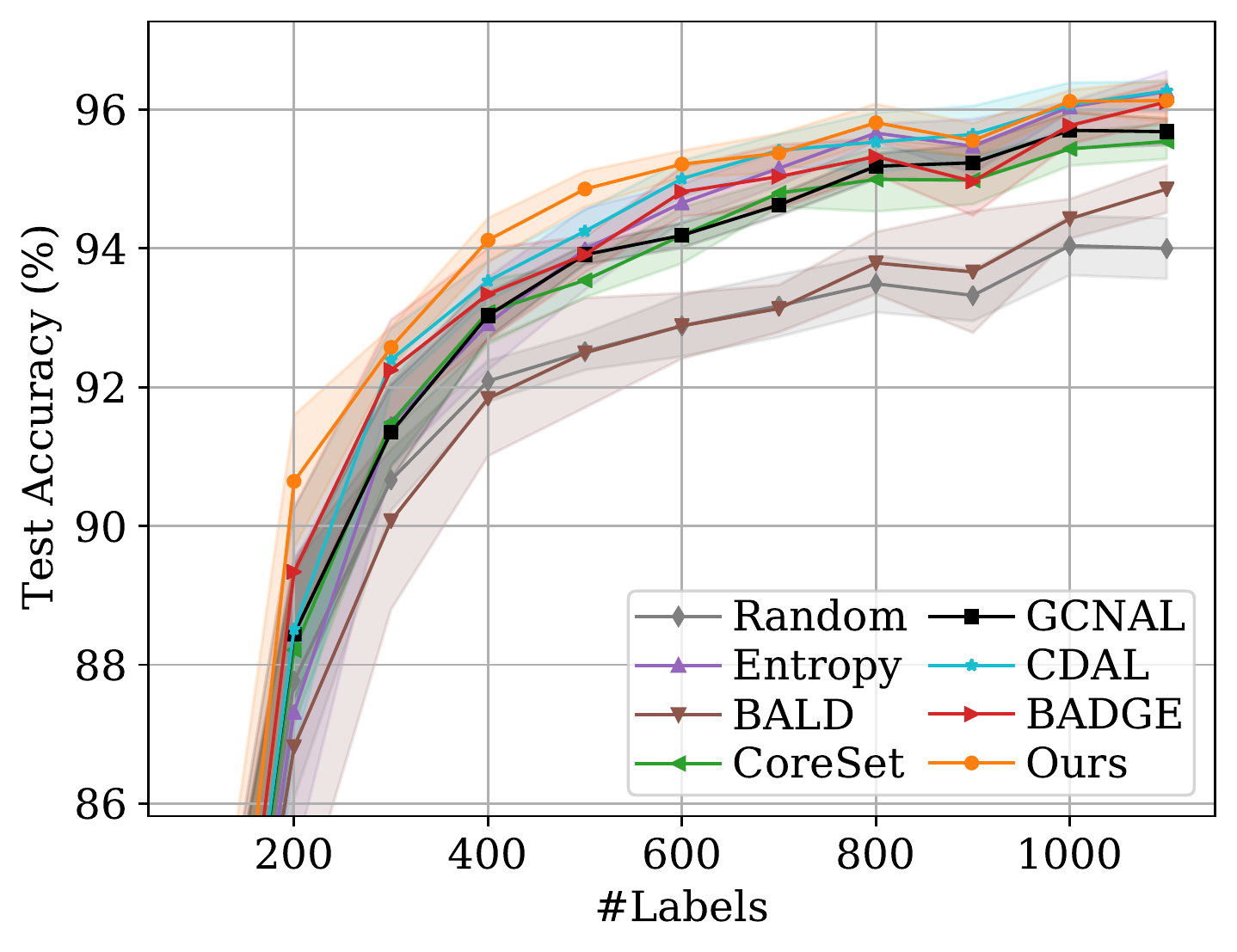}
		\vspace{-8mm}
		\caption{Small Budget, ResNet-18, DomainNet-Real-10}
		\vspace{-2mm}
	\end{figure}
	\begin{figure}[t]
		\centering
		\includegraphics[width=1.\linewidth]{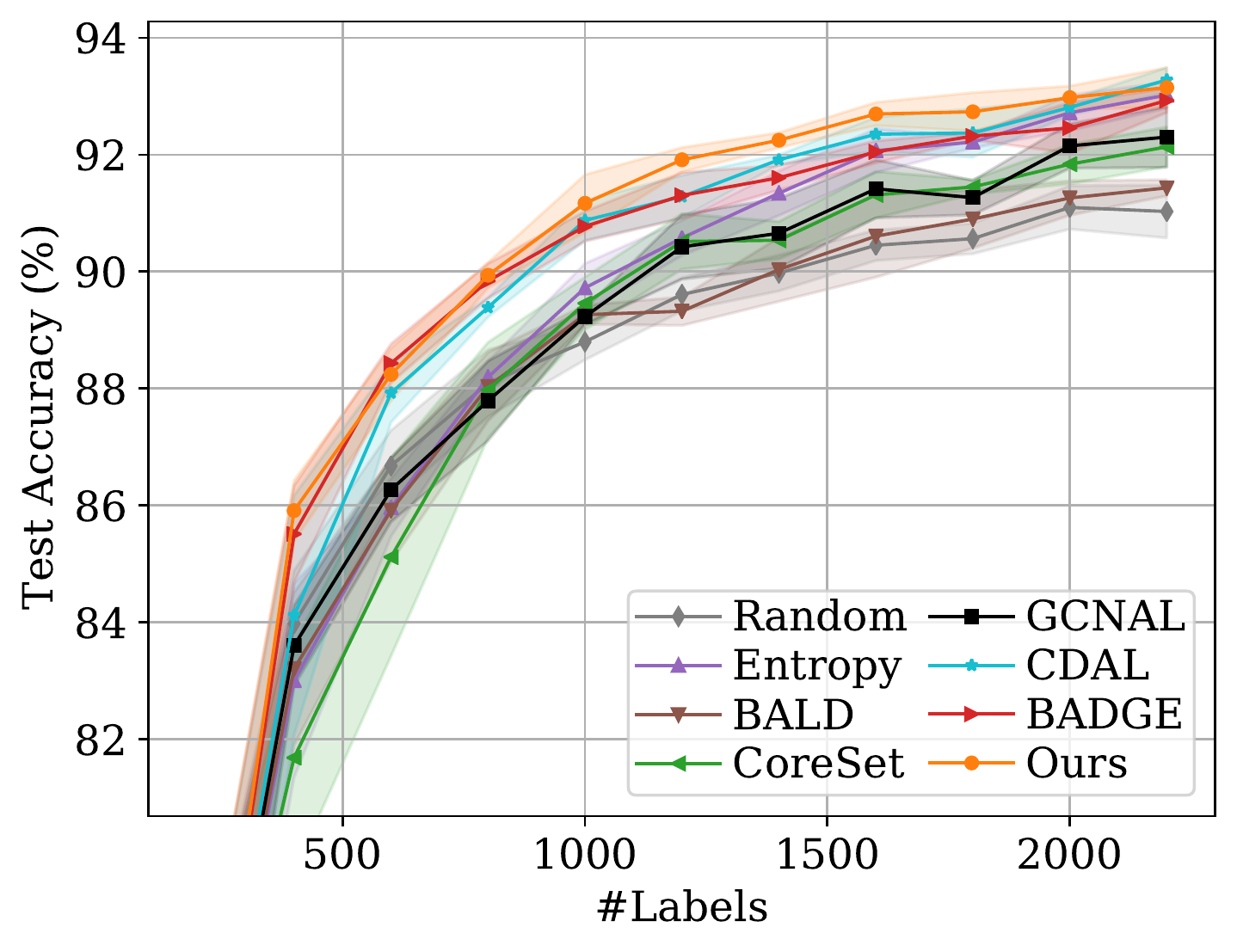}
		\vspace{-8mm}
		\caption{Small Budget, ResNet-18, DomainNet-Real-20}
		\vspace{-2mm}
	\end{figure}

	\begin{figure}[t]
		\centering
		\includegraphics[width=1.\linewidth]{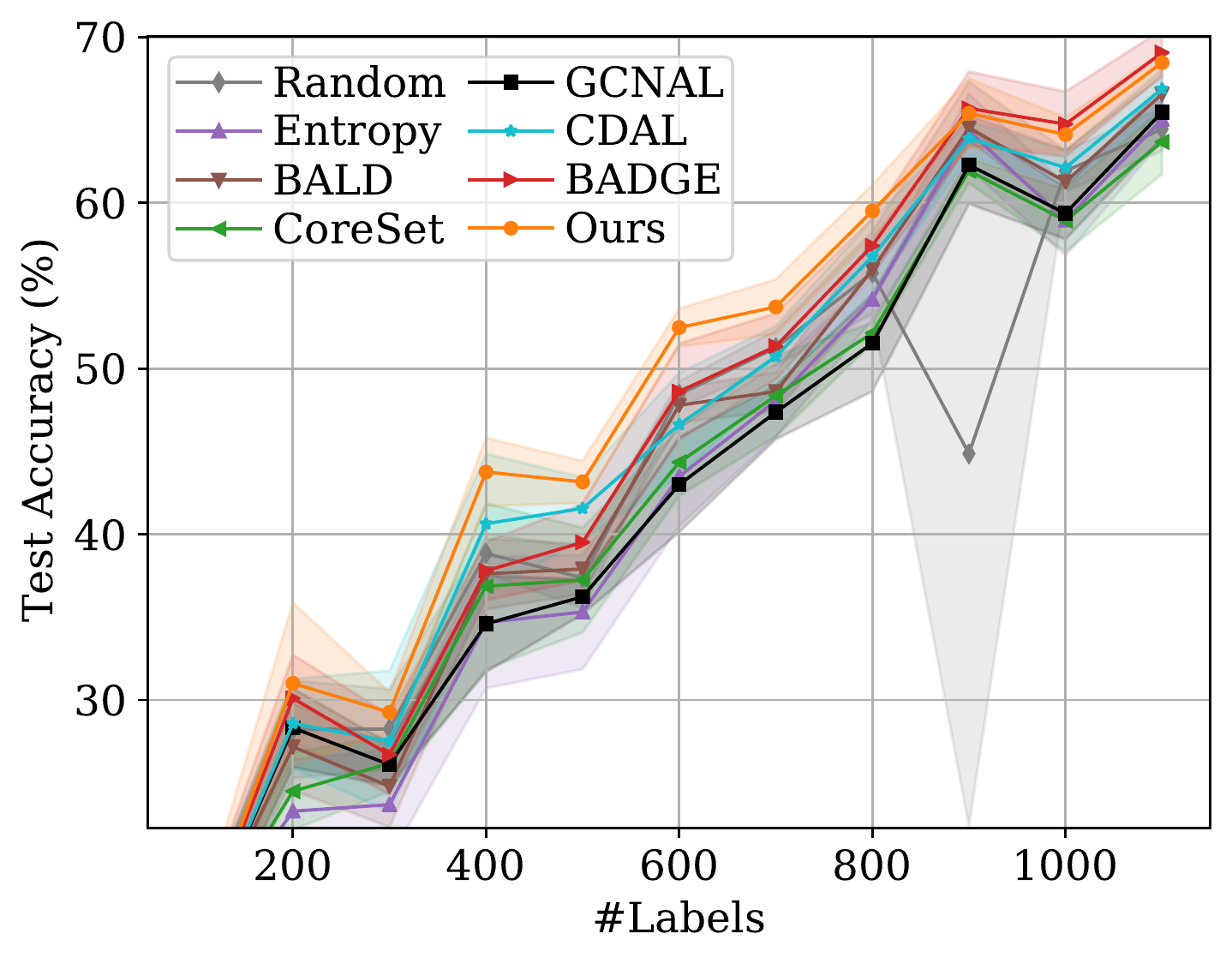}
		\vspace{-8mm}
		\caption{Small Budget, DenseNet-121, SVHN}
		\vspace{-2mm}
	\end{figure}
	\begin{figure}[t]
		\centering
		\includegraphics[width=1.\linewidth]{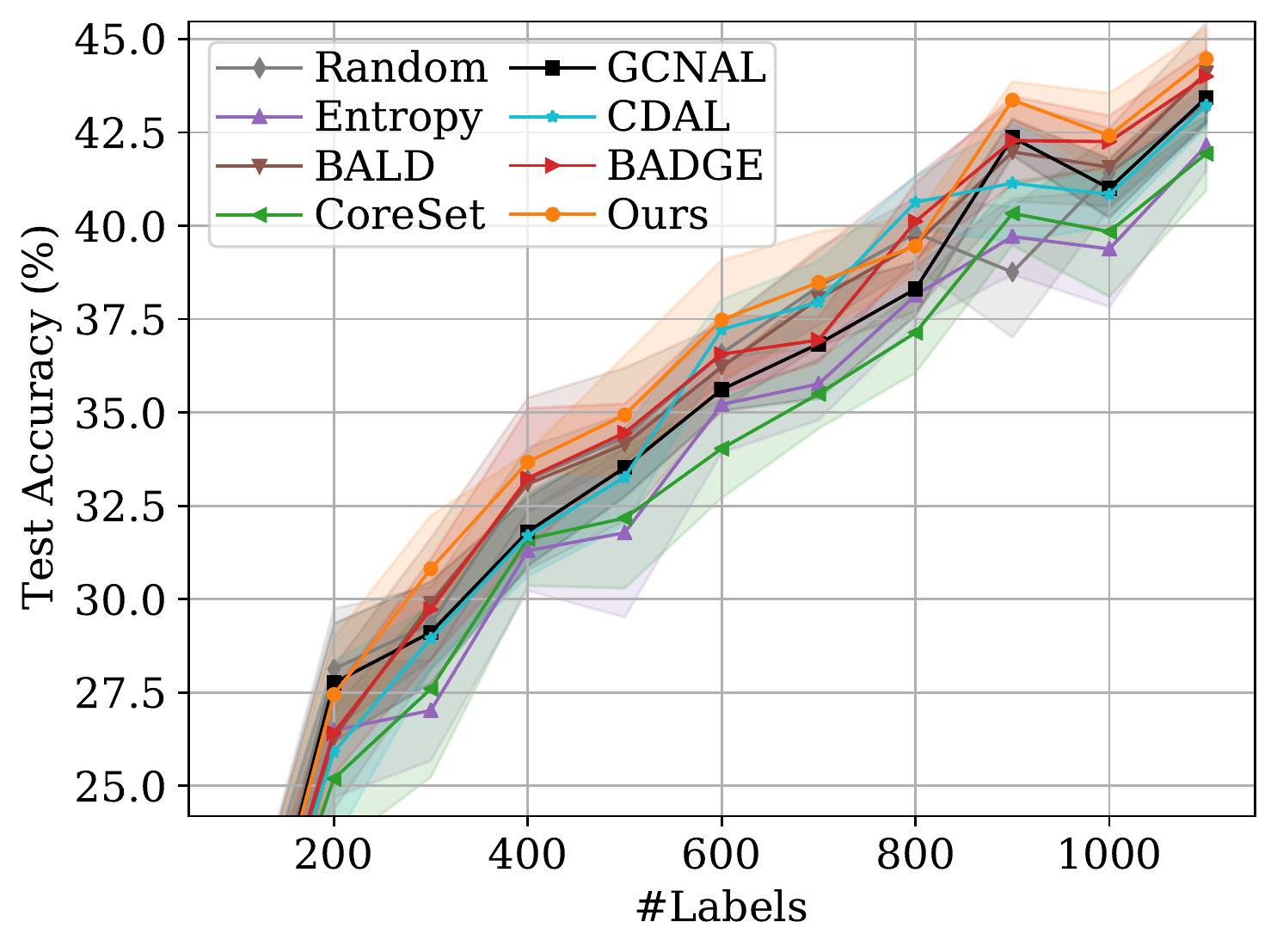}
		\vspace{-8mm}
		\caption{Small Budget, DenseNet-121, CIFAR10}
		\vspace{-2mm}
	\end{figure}
	\begin{figure}[t]
		\centering
		\includegraphics[width=1.\linewidth]{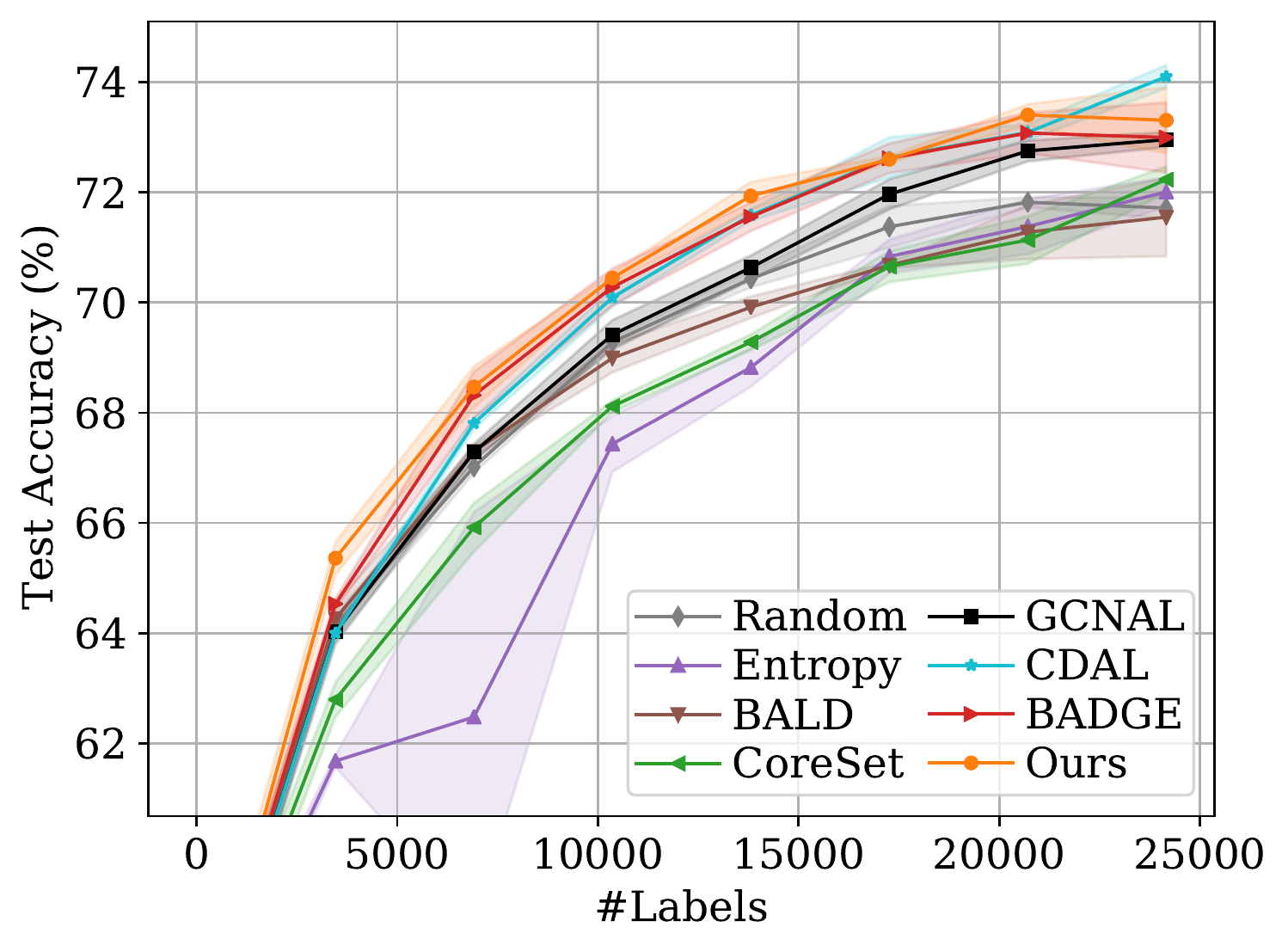}
		\vspace{-8mm}
		\caption{Small Budget, DenseNet-121, DomainNet-Real}
		\vspace{-2mm}
	\end{figure}
	\begin{figure}[t]
		\centering
		\includegraphics[width=1.\linewidth]{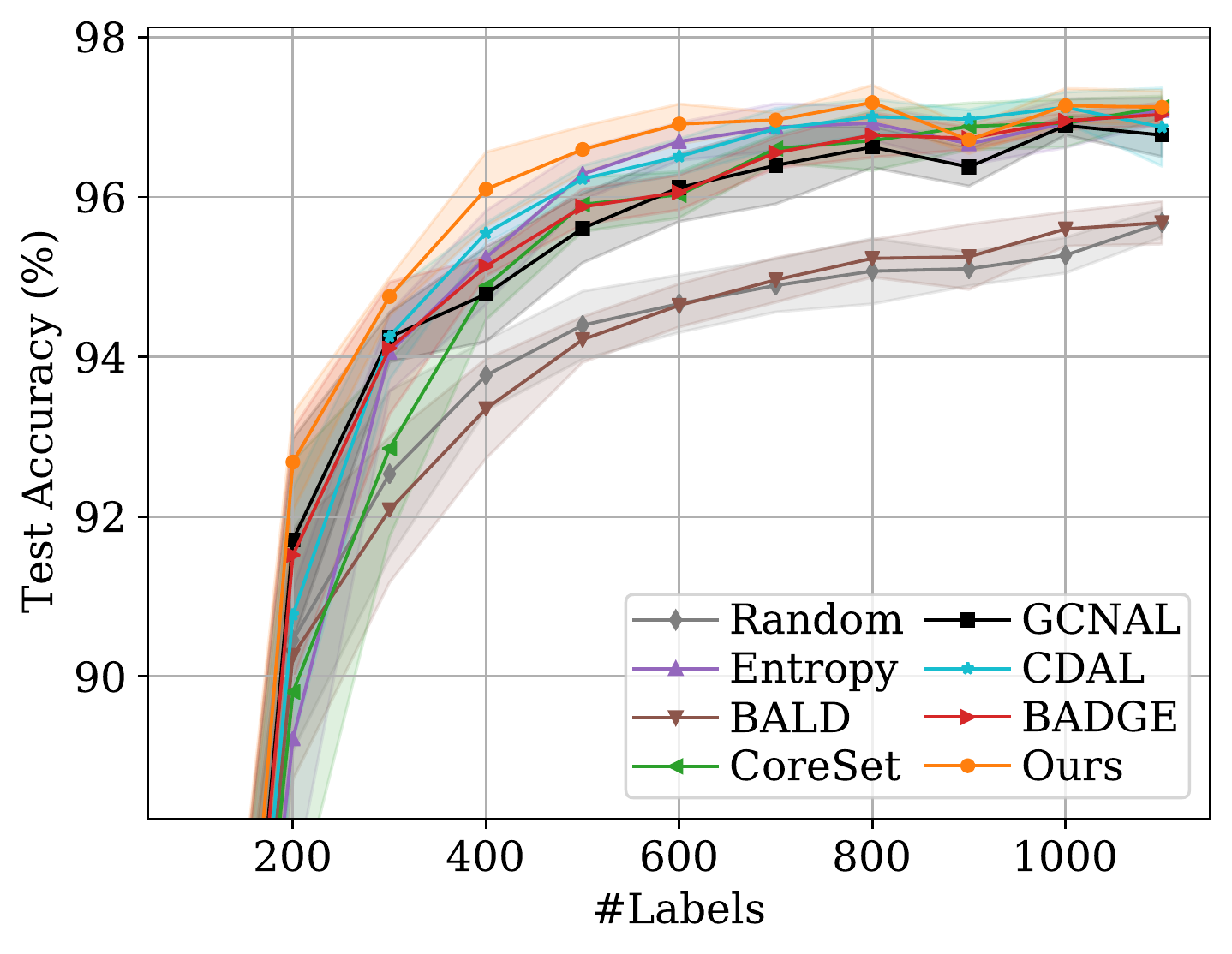}
		\vspace{-8mm}
		\caption{Small Budget, DenseNet-121, DomainNet-Real-10}
		\vspace{-2mm}
	\end{figure}
	\begin{figure}[t]
		\centering
		\includegraphics[width=1.\linewidth]{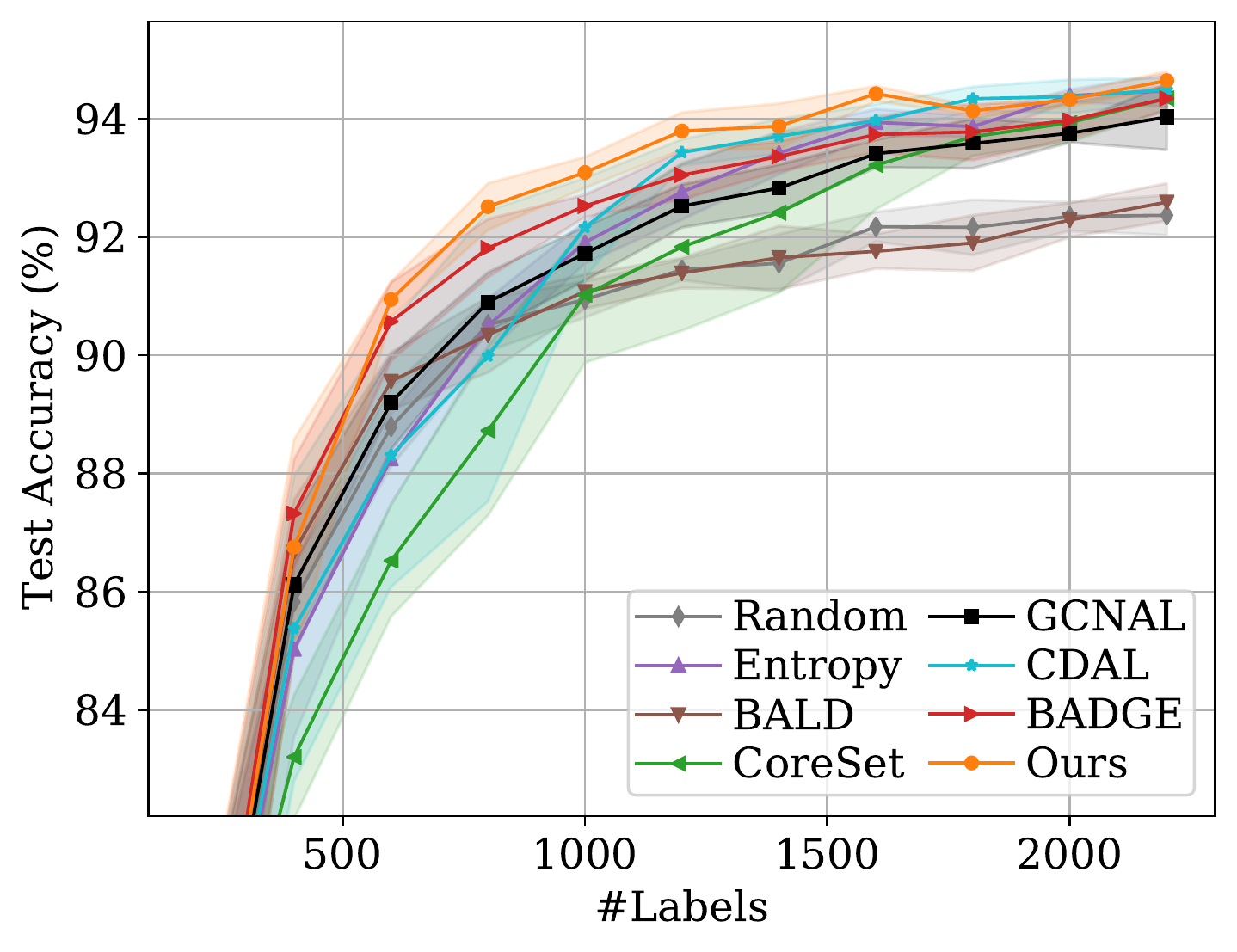}
		\vspace{-8mm}
		\caption{Small Budget, DenseNet-121, DomainNet-Real-20}
		\vspace{-2mm}
	\end{figure}
	
	\begin{figure}[t]
		\centering
		\includegraphics[width=1.\linewidth]{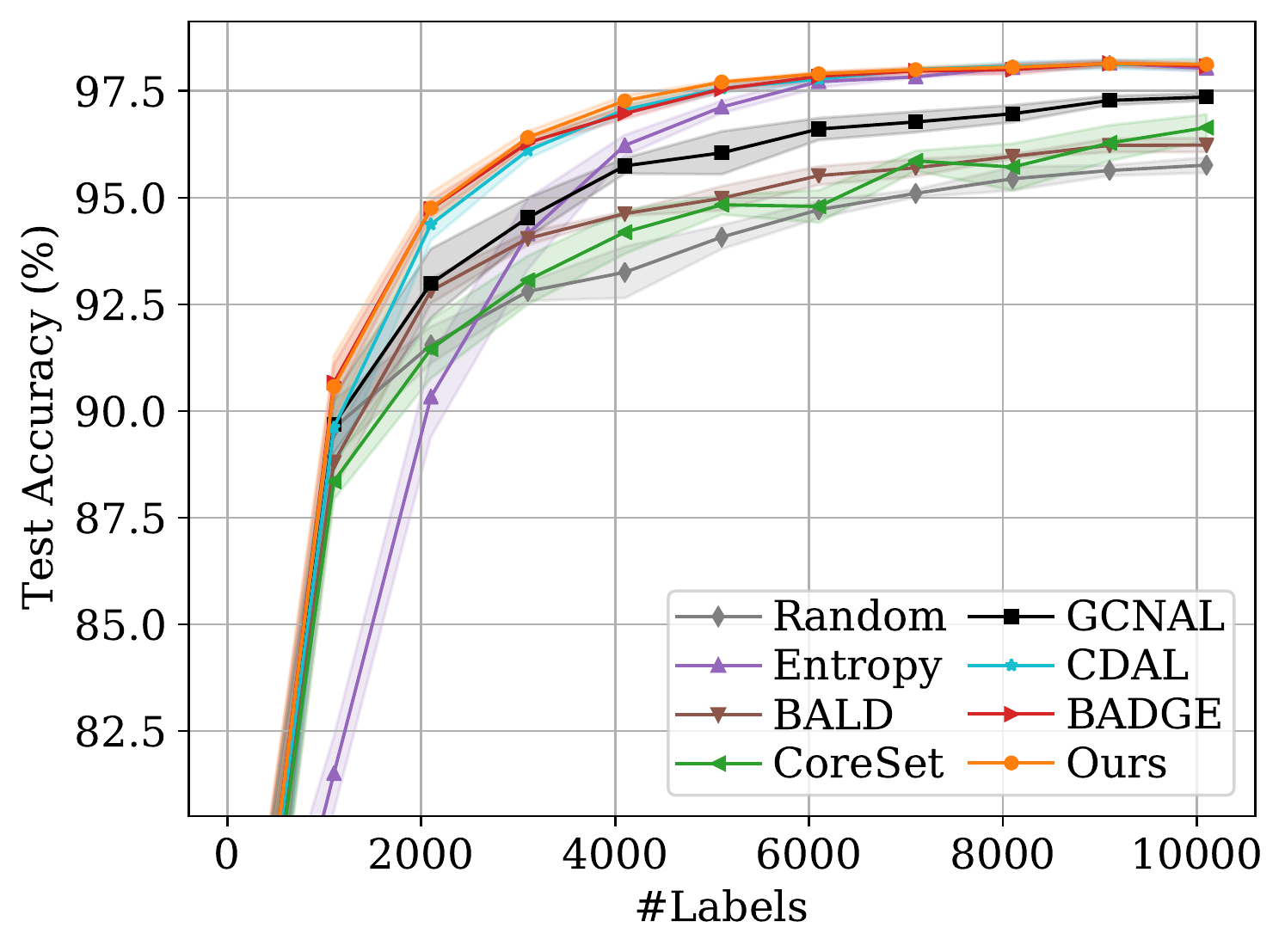}
		\vspace{-8mm}
		\caption{Large Budget, MLP, MNIST}
		\vspace{-2mm}
	\end{figure}
	\begin{figure}[t]
		\centering
		\includegraphics[width=1.\linewidth]{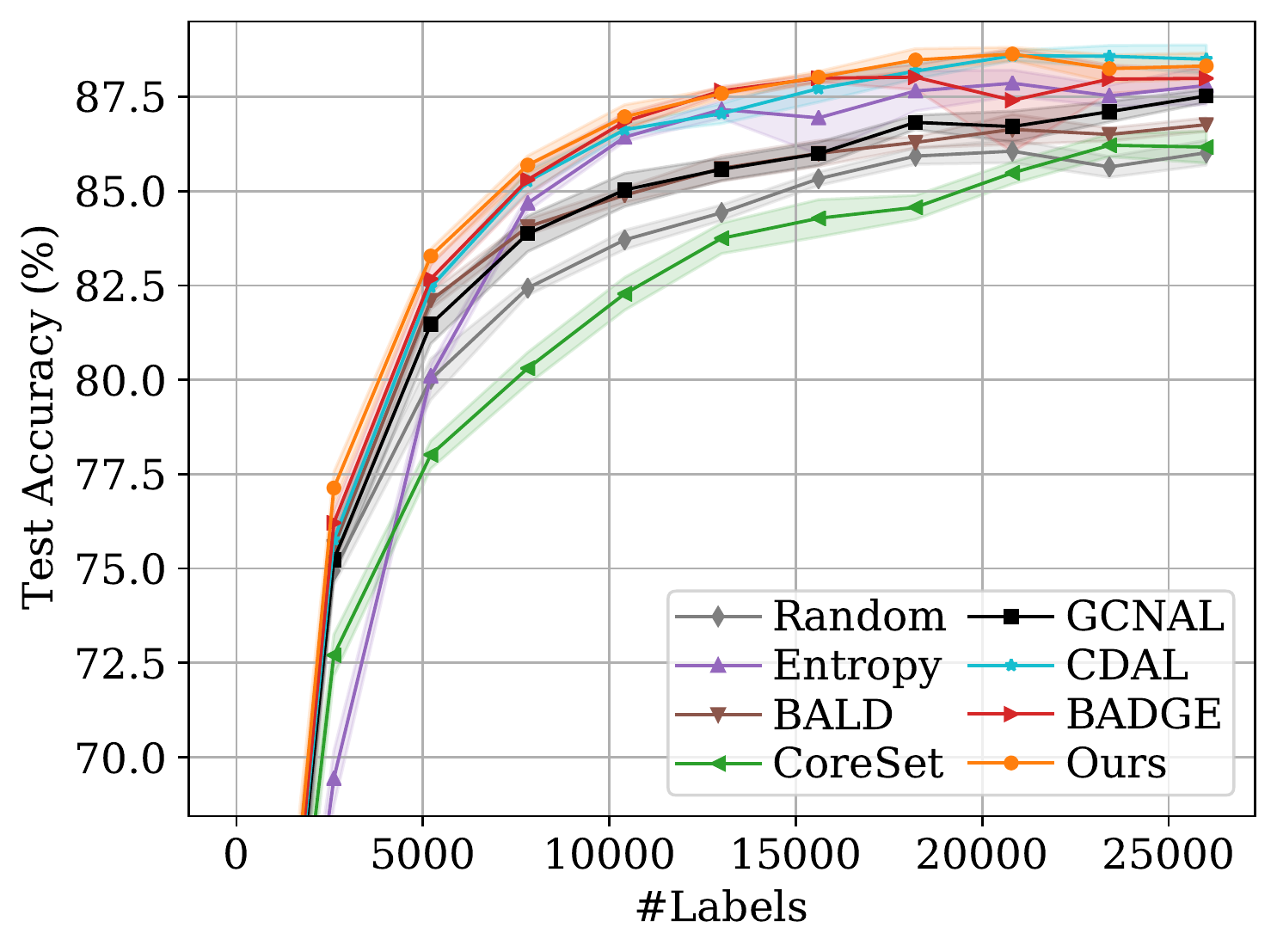}
		\vspace{-8mm}
		\caption{Large Budget,MLP, EMNIST}
		\vspace{-2mm}
	\end{figure}
	
	\begin{figure}[t]
		\centering
		\includegraphics[width=1.\linewidth]{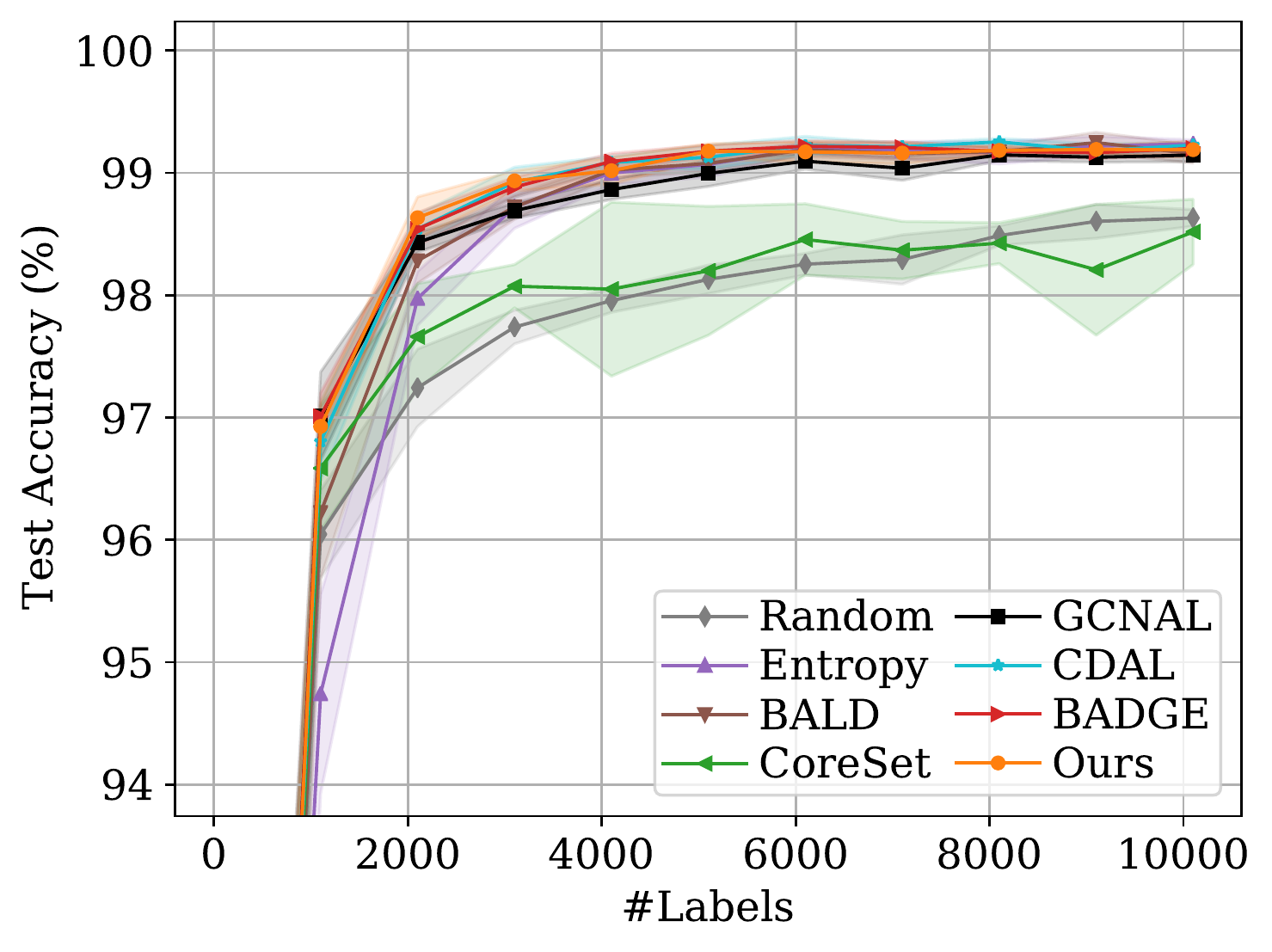}
		\vspace{-8mm}
		\caption{Large Budget, LeNet-5, MNIST}
		\vspace{-2mm}
	\end{figure}
	\begin{figure}[t]
		\centering
		\includegraphics[width=1.\linewidth]{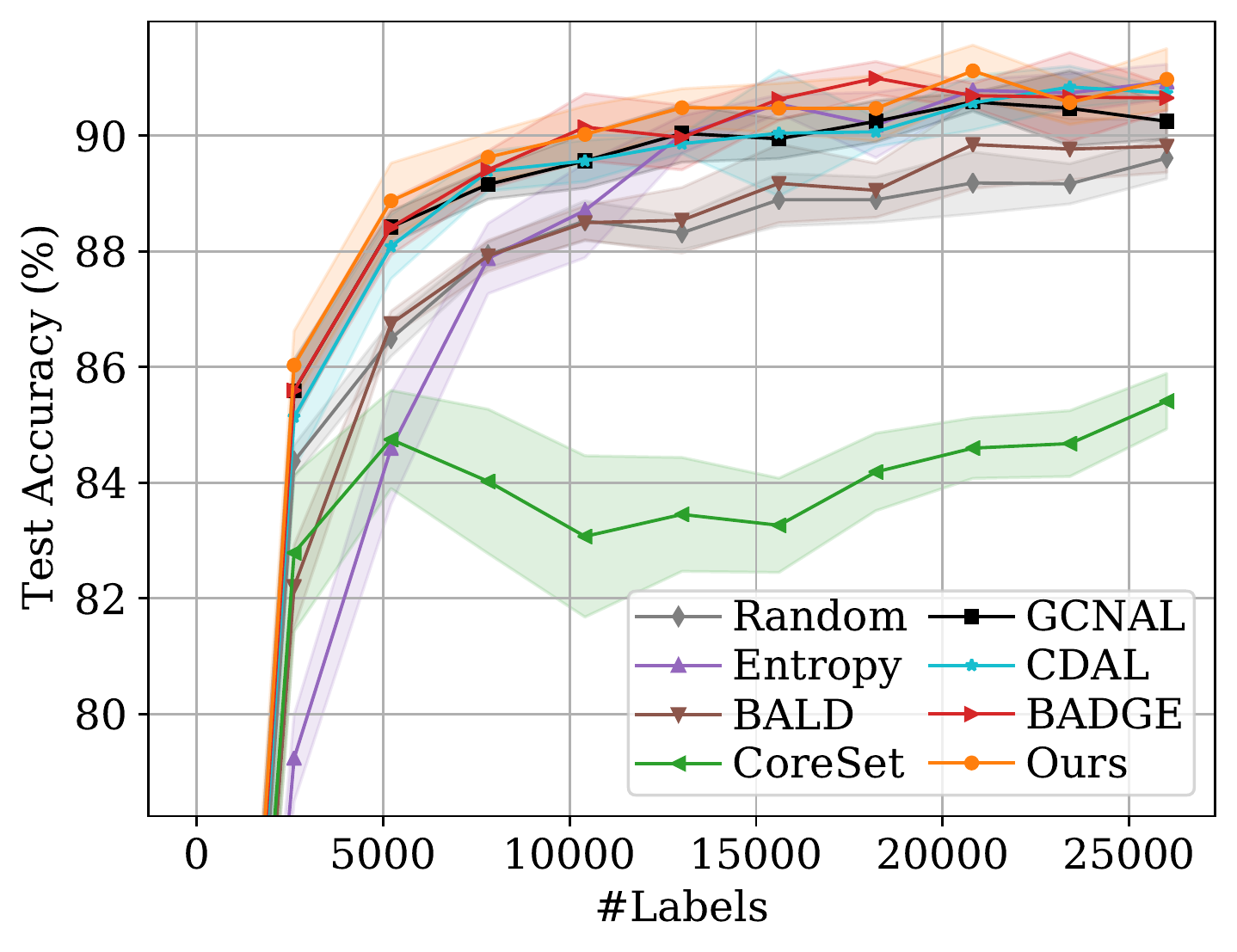}
		\vspace{-8mm}
		\caption{Large Budget, LeNet-5, EMNIST}
		\vspace{-2mm}
	\end{figure}
	
	\begin{figure}[t]
		\centering
		\includegraphics[width=1.\linewidth]{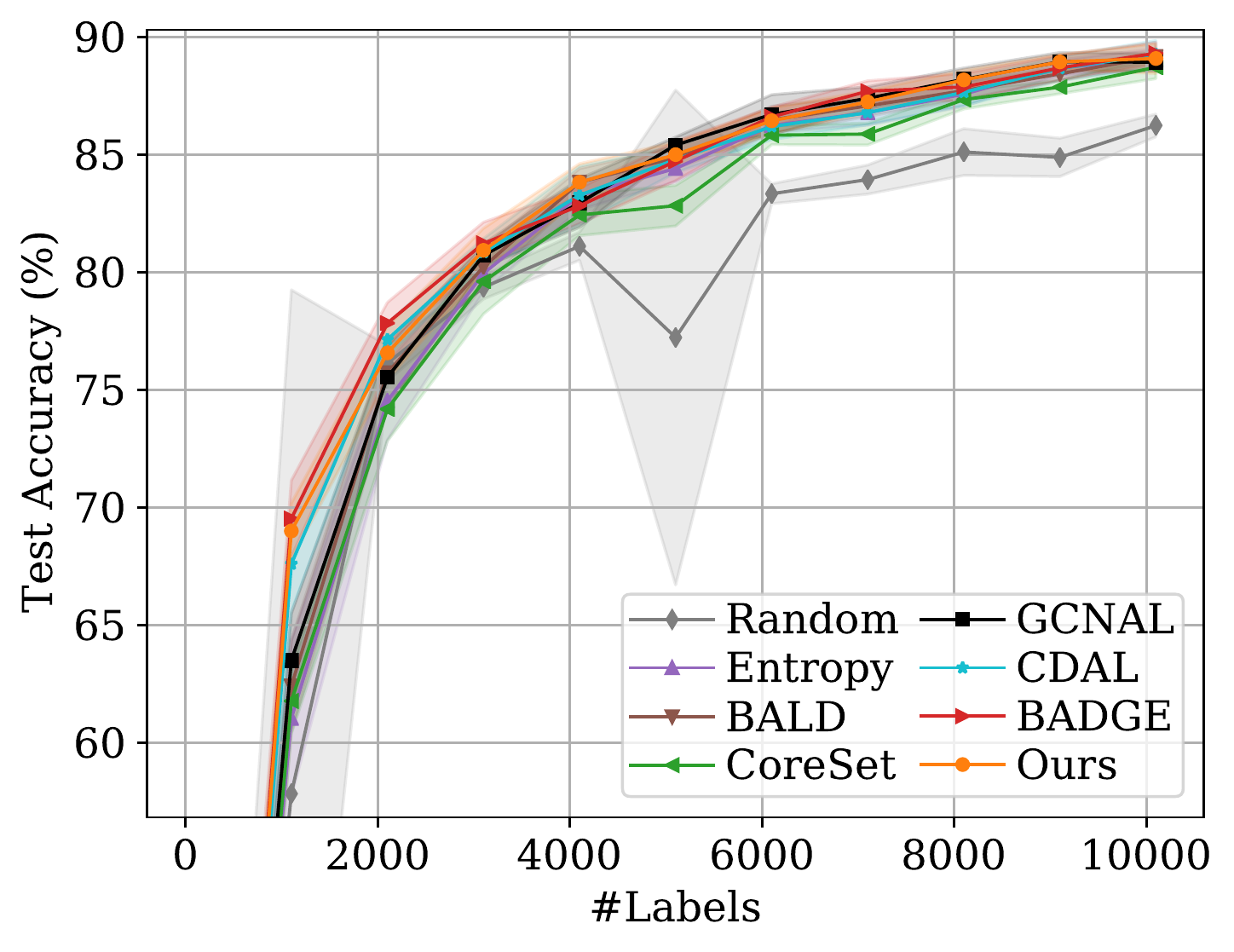}
		\vspace{-8mm}
		\caption{Large Budget, ResNet-18, SVHN}
		\vspace{-2mm}
	\end{figure}
	\begin{figure}[t]
		\centering
		\includegraphics[width=1.\linewidth]{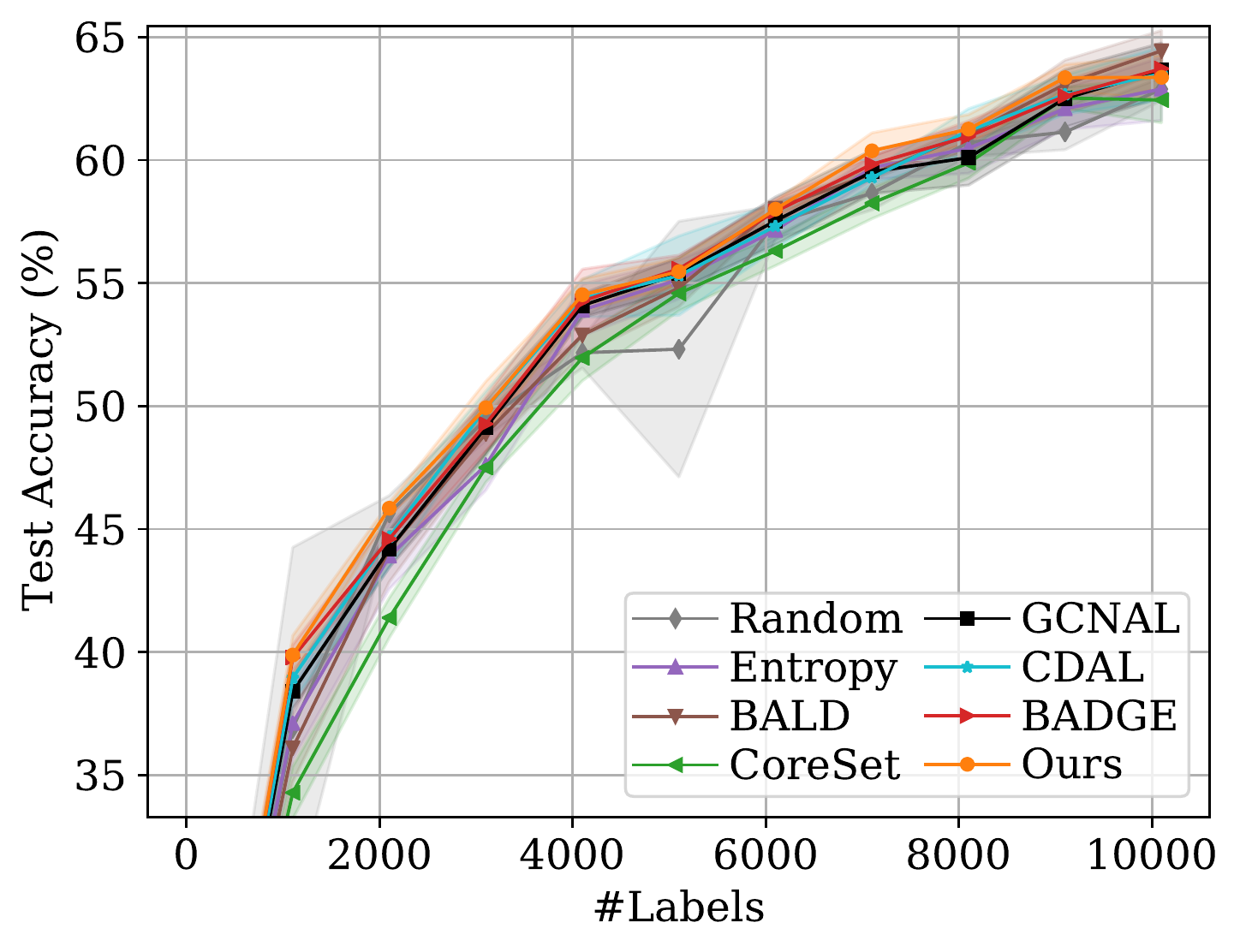}
		\vspace{-8mm}
		\caption{Large Budget, ResNet-18, CIFAR10}
		\vspace{-2mm}
	\end{figure}
	
	\begin{figure}[h]
		\centering
		\includegraphics[width=1.\linewidth]{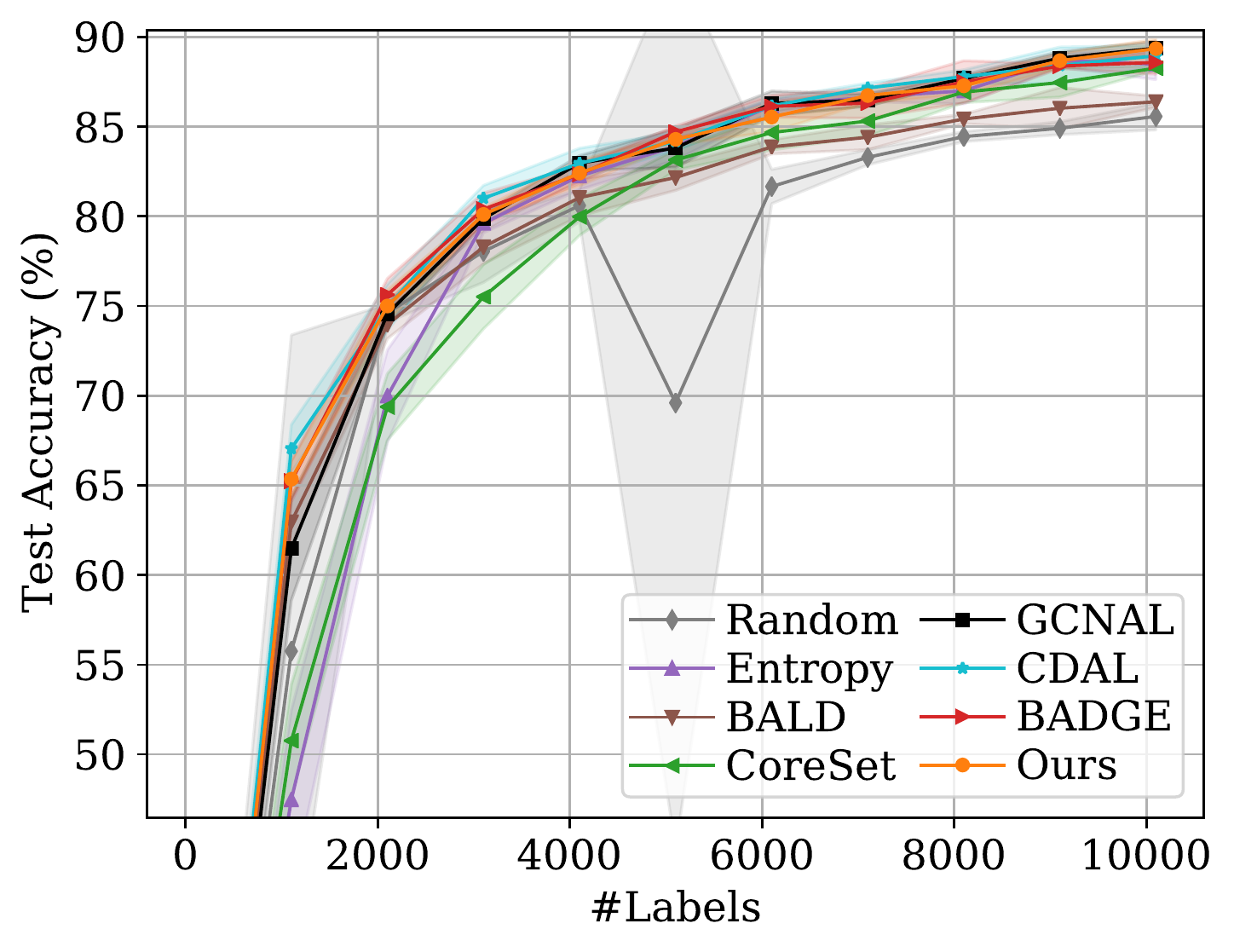}
		\vspace{-2mm}
		\caption{Large Budget, DenseNet-121, SVHN}
		\vspace{-2mm}
	\end{figure}
	\begin{figure}[h]
		\centering
		\includegraphics[width=1.\linewidth]{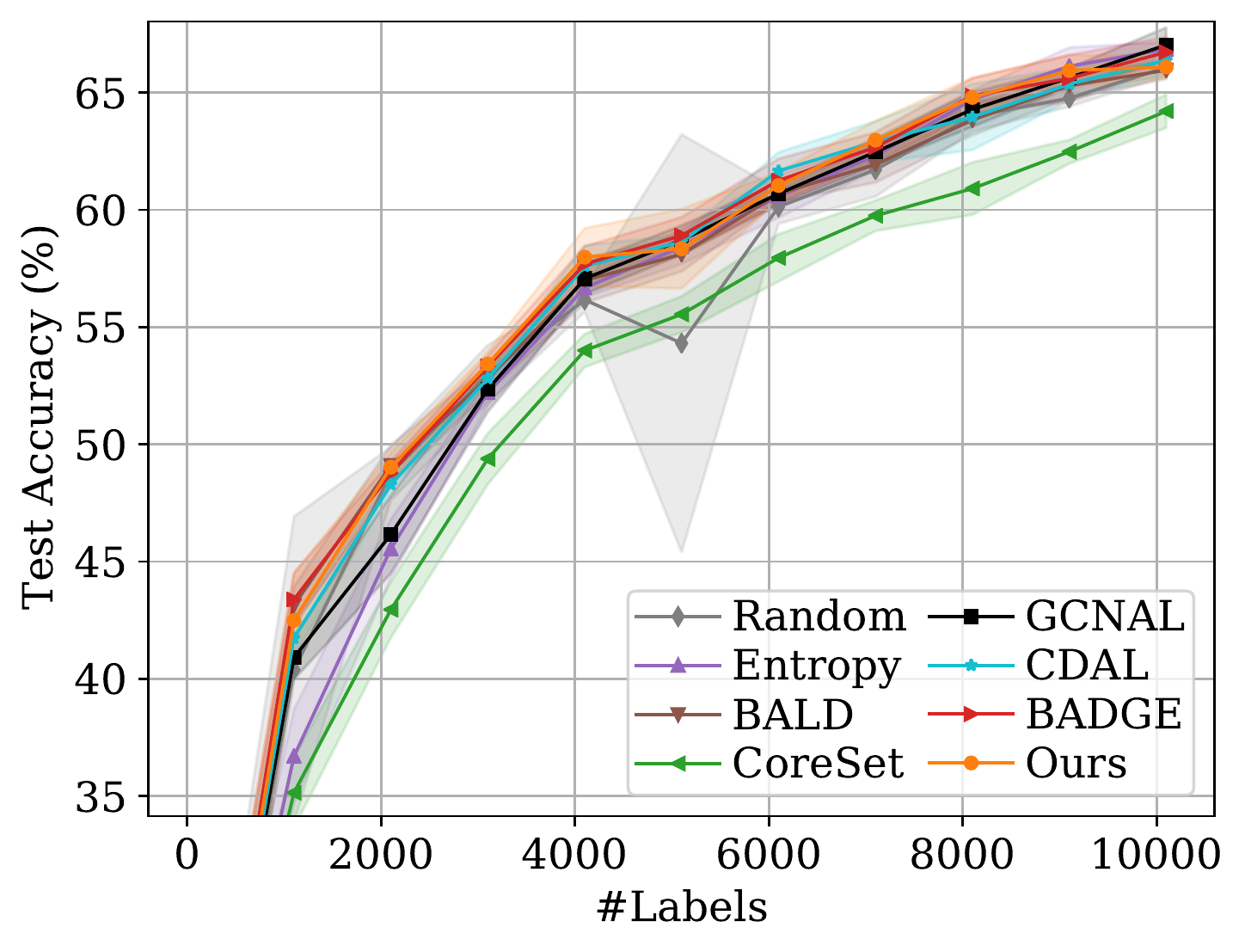}
		\vspace{-8mm}
		\caption{Large Budget, DenseNet-121, CIFAR10}
		\vspace{-2mm}
	\end{figure}
	
	\begin{figure}[h]
		\centering
		\includegraphics[width=1.\linewidth]{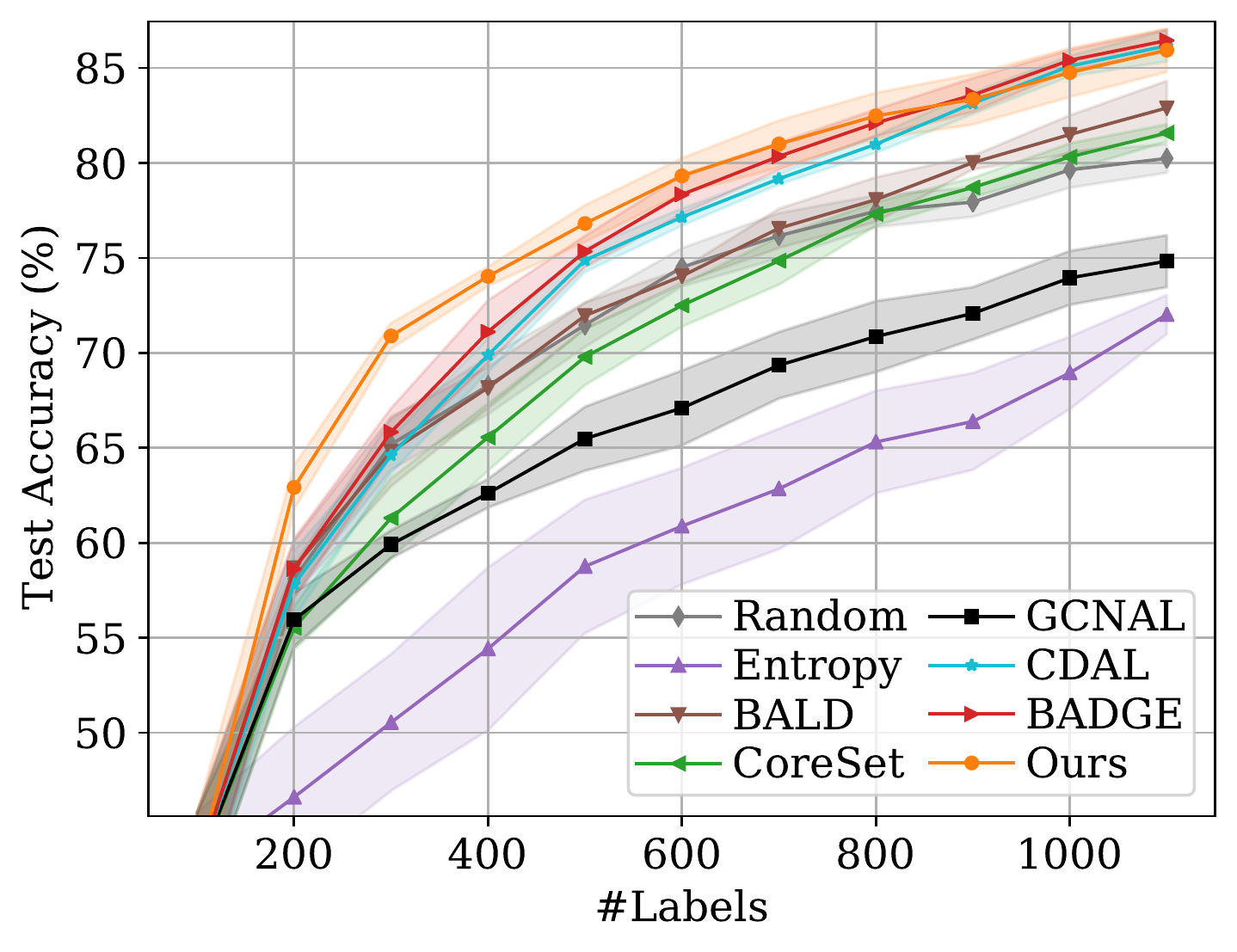}
		\vspace{-8mm}
		\caption{Small Budget, MLP, OpenML-6}
		\vspace{-2mm}
	\end{figure}
	\begin{figure}[h]
		\centering
		\includegraphics[width=1.\linewidth]{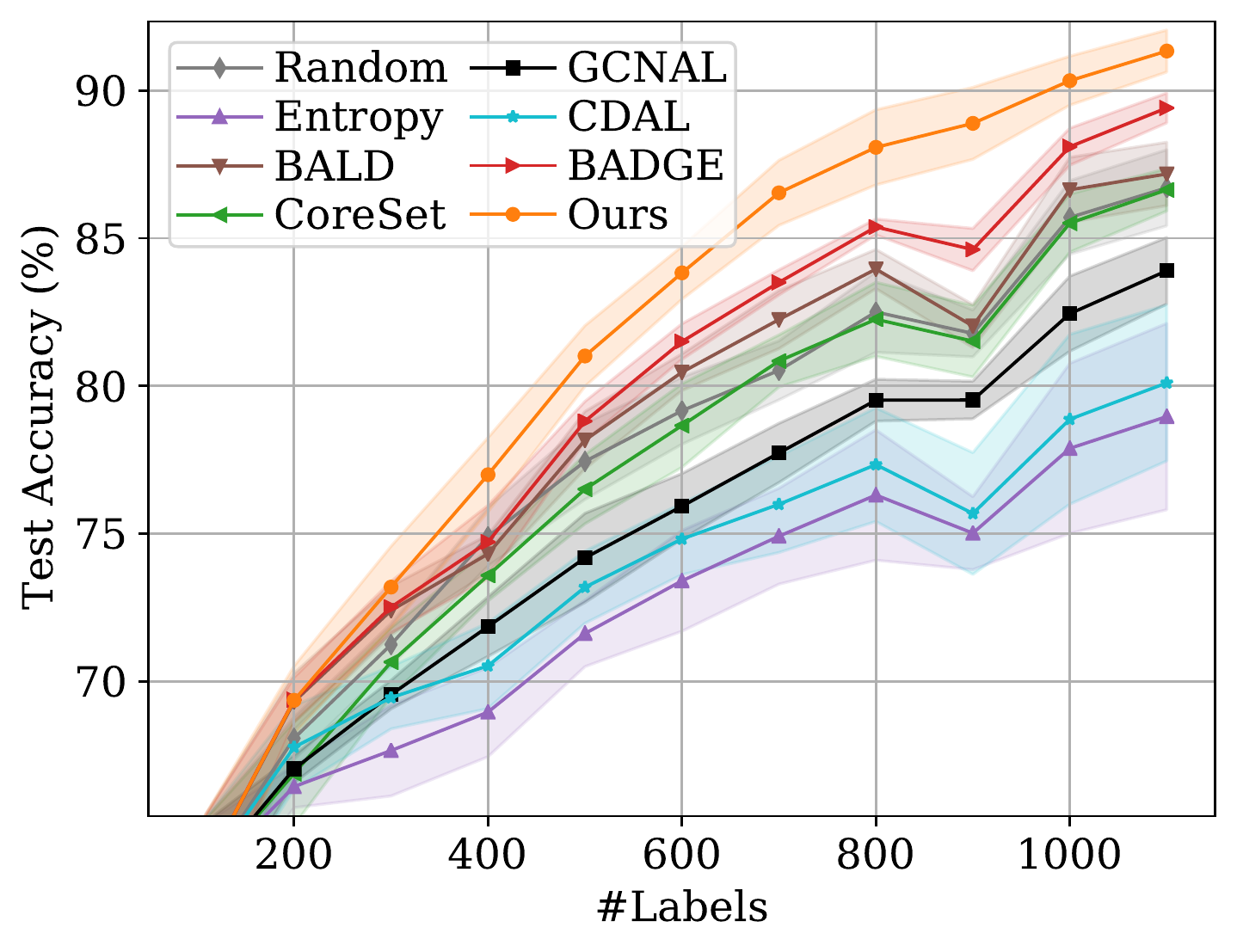}
		\vspace{-8mm}
		\caption{Small Budget, MLP, OpenML-155}
		\vspace{-2mm}
	\end{figure}

\end{document}